%% file: main.tex
\PassOptionsToPackage{dvipsnames}{xcolor}

\documentclass{article}

\usepackage{microtype}
\usepackage{graphicx}
\usepackage{booktabs} 

\usepackage{subcaption}

\usepackage{hyperref}

\usepackage[arxiv]{icml2025}

\usepackage{amsmath}
\usepackage{amssymb}
\usepackage{mathtools}
\usepackage{amsthm}

\usepackage{hyperref}
\usepackage{url}
\usepackage{booktabs} 
\usepackage{adjustbox}
\usepackage{graphicx}
\usepackage{amsmath}
\usepackage{amssymb}
\usepackage{pifont}
\usepackage{url}
\usepackage{tikz}
\usepackage{comment}
\usepackage{amsfonts}       
\usepackage{color}
\usepackage{bbm, dsfont}
\usepackage{mathtools}
\usepackage{amsthm}
\usepackage[normalem]{ulem}
\usepackage{colortbl}
\usepackage[dvipsnames]{xcolor}
\usepackage{multirow}
\usepackage{cutwin}
\usepackage{arydshln}
\usepackage{caption}
\usepackage{enumitem}
\usepackage{wrapfig}
\usepackage{microtype}
\usepackage{graphicx}
\usepackage{bm}
\usepackage{eqparbox}
\usepackage{makecell}
\usepackage{threeparttable}
\usepackage{wrapfig,lipsum,booktabs}

\definecolor{long_context_avg_color}{gray}{0.85}
\newcolumntype{m}{>{\columncolor{long_context_avg_color}}c}

\usepackage[capitalize,noabbrev]{cleveref}

\renewcommand{\Cref}[1]{\textcolor{RawSienna}{\cref{#1}}}  
\crefname{equation}{Eq.}{Eqs.}  
\crefname{section}{Sec.}{Secs.}  
\crefname{figure}{Fig.}{Figs.}  
\crefname{table}{Tab.}{Tabs.}  

\theoremstyle{plain}

\theoremstyle{definition}

\theoremstyle{remark}

\newcommand{\MethodName}{\textsc{RSQ}}
\newcommand{\MethodFullName}{Rotate, Scale, then Quantize}

\newcommand{\norm}[1]{\lVert#1\rVert}

\usepackage[textsize=tiny]{todonotes}

\icmltitlerunning{\MethodName{}: Learning from Important Tokens Leads to Better Quantized LLMs}

\begin{document}

\twocolumn[
\icmltitle{\MethodName{}: Learning from Important Tokens Leads to Better Quantized LLMs}



\icmlsetsymbol{equal}{*}

\begin{icmlauthorlist}
\icmlauthor{Yi-Lin Sung}{unc}
\icmlauthor{Prateek Yadav}{unc}
\icmlauthor{Jialu Li}{unc}
\icmlauthor{Jaehong Yoon}{unc}
\icmlauthor{Mohit Bansal}{unc}
\end{icmlauthorlist}

\icmlaffiliation{unc}{Department of Computer Science, UNC at Chapel Hill}

\icmlcorrespondingauthor{Yi-Lin Sung}{ylsung@cs.unc.edu}

\icmlkeywords{Quantization, Large Language Model, Compression, Layer-wise Quantization}

\vskip 0.3in
]



\printAffiliationsAndNotice{}  

\begin{abstract}

Layer-wise quantization is a key technique for efficiently compressing large models without expensive retraining. 
Previous methods typically quantize the weights of each layer by ``uniformly'' optimizing the layer reconstruction loss across all output tokens.
However, in this paper, we demonstrate that better-quantized models can be obtained by prioritizing learning from important tokens (\textit{e.g.} which have large attention scores).
Building on this finding, we propose \MethodName{} (\MethodFullName{}), which (1) applies rotations (orthogonal transformation) to the model to mitigate outliers (those with exceptionally large magnitude),
(2) scales the token feature based on its importance, and (3) quantizes the model using the GPTQ framework with the second-order statistics computed by scaled tokens. 
To compute token importance, we explore both heuristic and dynamic strategies. Based on a thorough analysis of all approaches, we adopt attention concentration, which uses attention scores of each token as its importance, as the best approach.
We demonstrate that \MethodName{} consistently outperforms baseline methods across multiple downstream tasks and three model families: LLaMA3, Mistral, and Qwen2.5. Additionally, models quantized with \MethodName{} achieve superior performance on long-context tasks, further highlighting its effectiveness.
Lastly, \MethodName{} demonstrates generalizability across various setups, including different model sizes, calibration datasets, bit precisions, and quantization methods. Our code is available at \url{https://github.com/ylsung/rsq}.

\end{abstract}

\input{sections/introduction}

\input{sections/related_work}

\input{sections/background}

\input{sections/method}

\input{sections/experiments}

\input{sections/conclusion}

\input{sections/impact_statement}

\input{sections/acknowledgement}

\bibliography{bibtex}
\bibliographystyle{icml2025}

\input{sections/appendix}

\end{document}

%% file: sections/introduction.tex
\section{Introduction} \label{sec:intro}

Large language models (LLMs)~\cite{team2024gemini,achiam2023gpt4} have recently achieved great success and have transformed the landscape of artificial intelligence. However, the substantial computational demands associated with these models pose significant challenges to their usage and deployment, especially in resource-constrained scenarios. 

Weight quantization~\cite{Han2015DeepCompression,Wu2015QuantizedCN} is a widely used technique for reducing the computational costs of LLMs by representing weight values with fewer bits. Among various approaches, post-training quantization (PTQ)~\cite{frantar2022obc,Liu2021ptq_in_vision} is particularly favored by practitioners, as it enables the quantization of pre-trained LLMs using only a small calibration dataset, eliminating the need for expensive retraining.

We focus on the layer-wise post-training quantization scheme~\cite{Hubara2020Adaquant,li2021brecq,Frantar2022GPTQ} that has been demonstrated to be both effective and efficient for quantizing large models. 
Layer-wise quantization methods quantize an LLM's weights one layer at a time by minimizing the token-level feature distance between the outputs of the original and quantized weights (i.e., the layer reconstruction loss, $\norm{\mathbf{W} \mathbf{X} - \tilde{\mathbf{W}} \mathbf{X}}^2_2$). 
Several advancements have been made to improve layer-wise quantization techniques in the past few years. For example, GPTQ~\cite{Frantar2022GPTQ} improves the efficiency and stability to compute second-order statistics and their inverses. QuIP\#~\cite{tseng2024quip_sharp} and AQLM~\cite{Egiazarian2024aqlm} represent quantized weights with vectors rather than fixed scalars. 
Additionally, QuIP~\cite{Chee2023QuIP} and QuaRot~\cite{ashkboos2024quarot} demonstrate through empirical studies that weight outliers--parameters with unusually large magnitudes--can be effectively mitigated by applying orthogonal transformations.

Previous methods commonly perform layer-wise weight quantization by optimizing the layer reconstruction loss across all input tokens uniformly. 
However, research has shown that LLMs do not treat all tokens equally: (1) StreamingLLM~\cite{Xiao2023attention_sink} shows that initial tokens often have strong attention scores, (2) H$_{2}$O~\cite{Zhang2023H2O} reveals that some tokens in KV cache contribute most of the attention values while decoding, and (3) RHO-1~\cite{Lin2024Rho1} demonstrates not all tokens are equal in training LLMs. 
Since quantized models inherently lose information due to the reduced capacity of lower bit representations, we argue that it should be particularly crucial for them to focus on learning and preserving the most critical information during the quantization process to maximize their performance. Inspired by these insights, we reconsider the conventional approach in quantization methods by optimizing the layer reconstruction loss over only a subset of important input tokens (\textit{i.e.}, using only the first 1/4 of the tokens). Our findings reveal that this strategy improves the quantized model's accuracy across ten downstream tasks by up to 2.2\%.

Building on our findings and previous approaches, we propose \MethodName{} to quantize the model in three steps: (1) \textit{rotate} (orthogonally transform) the model to mitigate weight outliers, 
(2) \textit{scale} the token feature based on its importance, and (3) \textit{quantize} the weights using the GPTQ mechanism while leveraging token importance. 
We note that the token importance integrates seamlessly into the GPTQ framework in the third step, ensuring both compatibility and efficiency. \Cref{fig:main_figure} illustrates the three steps in \MethodName{}.

In this paper, we explore two categories of approaches for obtaining token importance: (1) heuristic approaches and (2) dynamic approaches. Within the heuristic category, we investigate methods such as First-N and First\&Last-N, which prioritize initial tokens and a combination of initial and final tokens for quantization, respectively. 
These approaches outperform the conventional quantization method of optimizing across all tokens, achieving peak performance when N is roughly 5–10\% of the total tokens. 
It is important to note that the initial or final tokens do not inherently contain more meaningful semantic information. Instead, their importance likely stems from their positional characteristics and their tendency to receive stronger attention scores~\cite{Xiao2023attention_sink,sun2024massive_attention}. 

In this aforementioned approach, token importance was determined solely based on heuristics (e.g., token positions). To further improve performance beyond heuristic methods, we also explore several dynamic approaches for computing token importance individually based on each input. Specifically, we investigate TokenFreq, where \textit{less frequent} tokens are considered more important; ActNorm, which prioritizes tokens with \textit{larger norms}; ActDiff, where tokens whose \textit{features change less} after one layer are assigned higher importance; TokenSim, which gives greater weight to tokens that are \textit{less similar} to others; and AttnCon, where tokens receiving \textit{higher attention} scores are considered more important.
Among these, AttnCon performs the best as it more explicitly models each token's impact on the other tokens. Therefore, we adopt it as our final strategy.
In these dynamic approaches, we also observe the presence of positional biases, similar to those seen in heuristic methods. To ensure that tokens in ``less important'' positions are not wasted, we introduce a data augmentation strategy. This involves expanding each sample by shifting it forward by several positions and adding the shifted samples to the calibration dataset. This augmentation ensures a more comprehensive utilization of these tokens.

We evaluate \MethodName{} on WikiText-2 and a diverse set of tasks, including LAMBADA, WinoGrande, ARC, HellaSwag, PIQA, MMLU, GSM8k, and TruthfulQA, using three models: LLaMA3-3B-Instruct, Mistral-Nemo-12B, and Qwen2.5-7B. Our results demonstrate that \MethodName{} achieves absolute improvements of 1.6\%, 0.9\%, and 0.4\% in average accuracy across these tasks for the three models, respectively.
To facilitate a comprehensive evaluation, we compare \MethodName{} against baseline methods on subsets of various long-context benchmarks, including LongEval, L-Eval, LongICLBench, and LongCodeArena. Our results show that \MethodName{} achieves improvements of 3.0\%, 2.5\%, 0.6\%, and 0.025, respectively, over QuaRot on these benchmarks. Lastly, we demonstrate the generalizability of \MethodName{} across various steps, such as different model sizes, calibration datasets, bit precisions, and quantization methods. 
Notably, the performance improvement is more pronounced at lower bit precisions, suggesting that ``learning from important token'' can be a critical component in pursuing effective extreme compression.

%% file: sections/related_work.tex
\section{Related Work} \label{sec:related work}

Post-training quantization (PTQ) has gained significant attention for its ability to quantize pre-trained models without requiring expensive retraining.
Methods such as ZeroQuant~\cite{Yao2022ZeroQuantEA} and QLoRA~\cite{Dettmers2023QLoRAEF} apply round-to-nearest techniques to the weights, even without utilizing calibration datasets. While these approaches are extremely efficient, they often yield suboptimal performance due to their lack of information about the data distribution.
To address this limitation, several PTQ methods leverage calibration datasets for improved quantization. These data-dependent PTQ methods are also often referred to as layer-wise quantization methods, as they typically quantize models one layer at a time for efficiency.
Specifically, GPTQ~\cite{Frantar2022GPTQ} and OBC~\cite{frantar2022obc} quantize weights and adjust the remaining weights with data-derived Hessian matrices (second-order statistics) accordingly. 
AWQ~\cite{Lin2023AWQAW} minimizes quantization error by rescaling weights based on the activation distribution.
QuIP\#~\cite{tseng2024quip_sharp} and AQLM~\cite{Egiazarian2024aqlm} represent groups of quantized weights with vectors instead of scalar values. 

One challenge in weight quantization is the presence of outliers in the weights, as LLMs often contain a subset of weights with exceptionally large magnitudes~\cite{dettmers2024spqr,Kim2023SqueezeLLMDQ,Yu2024super_weight}. During quantization, these outliers increase the range of the quantized weights. As a result, most weights are forced into a narrow range to fit the outliers within the quantized representation, which ultimately leads to suboptimal performance. 
One way to address this challenge is through mixed-precision quantization~\cite{dettmers2024spqr,Dettmers2022LLMint88M,Kim2023SqueezeLLMDQ}; however, current hardware lacks efficient support for inference using this technique. 
Recent studies, such as QuIP~\cite{Chee2023QuIP} and QuaRot~\cite{ashkboos2024quarot}, have empirically shown that outliers in weights can be effectively mitigated by applying orthogonal transformations (rotations)~\cite{ashkboos2024slicegpt}, which reduce their impact and minimize the need for mixed-precision quantization. 

Different from previous advancements in layer-wise quantization, \MethodName{} focuses on improving the learning objective of layer-wise quantization methods (detailed in \Cref{ssec:rsq}) by prioritizing important tokens. 
Our approach also builds on QuaRot by applying rotations to mitigate weight outliers and leveraging GPTQ for quantization. This integration makes \MethodName{} a comprehensive and holistic quantization strategy.

%% file: sections/background.tex
\section{Background} \label{sec:background}

\subsection{Transformer Architecture} \label{ssec:transformer}

The core components of a transformer are the attention layers and feed-forward network (FFN). An attention layer consists of four linear modules: $\mathbf{W}_q$, $\mathbf{W}_k$, $\mathbf{W}_v$ and $\mathbf{W}_o$. Similarly, an FFN contains three linear modules: $\mathbf{W}_{up}$, $\mathbf{W}_{gate}$, $\mathbf{W}_{down}$. Attention layers are responsible for capturing token dependency and global information while FFNs, positioned after each attention block, perform token-wise feature transformations. 

\subsection{Removing Outliers with Rotation} \label{ssec:rotation}

Transformer architecture exhibits computational invariance~\cite{ashkboos2024slicegpt}, allowing an orthogonal transformation (\textit{a.k.a.} rotation) to be applied to one layer and its transpose to be applied to the subsequent layer without altering the outputs. Specifically, given a two-layer module with its output defined as $\mathbf{Y} =\mathbf{W}_2\mathbf{W}_1\mathbf{X}$,  it follows that:
\begin{equation}
    \mathbf{Y} =(\mathbf{W}_2\mathbf{Q}^{\top})(\mathbf{Q}\mathbf{W}_1)\mathbf{X}
\end{equation}
where both weight matrices, $\mathbf{W}_1$ and $\mathbf{W}_2$ are orthogonally transformed by orthogonal matrix $\mathbf{Q}$ ($\mathbf{Q}^\top \mathbf{Q} = \mathbf{Q} \mathbf{Q}^\top = \mathbf{I}$ by definition). This property remains valid even when an RMSNorm~\cite{Zhang2019RMSNorm} layer is placed between $\mathbf{W}_1$ and $\mathbf{W}_2$.

Prior studies have empirically shown that applying orthogonal transformations to modern pre-trained LLMs effectively reduces weight outliers~\cite{Chee2023QuIP, ashkboos2024quarot}. Moreover, due to the computational invariance property, these transformations do not alter the model's output if they are properly inserted into the model.

Concretely, assume we initialize an orthogonal transformation matrix $\mathbf{Q}$, which can be a random orthogonal matrix or a randomized Hadamard matrix.
We transform the following weight matrices from $\mathbf{W}$ to $\mathbf{W}\mathbf{Q}^{\top}$: $\mathbf{W}_q$, $\mathbf{W}_k$, $\mathbf{W}_v$ in attention layers, $\mathbf{W}_{up}$, $\mathbf{W}_{gate}$ in FFN layers, and the \text{lm\_head} layer. Similarly, we transform the following weight matrices from $\mathbf{W}$ to $\mathbf{Q}\mathbf{W}$: $\mathbf{W}_o$ in attention layers, $\mathbf{W}_{down}$ in FFN layers, and the embedding layer.
For more details on the rotation and orthogonal transformation, please refer to SliceGPT~\cite{ashkboos2024slicegpt}. 

\subsection{Layer-wise Quantization} \label{ssec:layer-wise quantization}

Layer-wise quantization offers a more efficient alternative to full-model quantization by processing each layer individually. In this approach, the quantized weights for each layer are optimized by minimizing the layer-wise reconstruction loss, which measures the feature distance between the outputs produced by the original weight matrix $\mathbf{W}$ and the quantized weight matrix $\tilde{\mathbf{W}}$: $\norm{\mathbf{W} \mathbf{X} - \tilde{\mathbf{W}} \mathbf{X}}^2_2 = \sum_{i=0}^{T} \norm{\mathbf{W} \mathbf{x}_i - \tilde{\mathbf{W}} \mathbf{x}_i}_2^2$, where $\mathbf{X}$ is input tokens' features.
Note that the reconstruction loss for each token is weighed \textbf{uniformly} across a sequence of token features $(\mathbf{x}_1, \mathbf{x}_2, ..., \mathbf{x}_T)$.
The OBC framework~\cite{frantar2022obc} provides an explicit formula to optimally quantize a column of weight based on the layer-reconstruction loss, as well as the optimal update of the remaining weights which compensates for the quantization. 
Specifically, for the $q$-th column of weight $\mathbf{W}_{:,q}$ to be quantized, OBC first quantizes the weight in the round-to-nearest manner ($\tilde{\mathbf{W}}_{:, q} = \text{quant} (\mathbf{W}_{:,q})$), and adjust the remaining weights according to the following formula:

\begin{equation} \label{eq:gptq}
    \bm{\delta} = - \frac{\mathbf{W}_{:,q} - \text{quant} (\mathbf{W}_{:,q})}{\mathbf{H}^{-1}_{qq}} \cdot \mathbf{H}^{-1}_{q, :}
\end{equation}

where $\mathbf{H} = 2 \mathbf{X}\mathbf{X}^{\top}$ is the Hessian matrix. After quantizing each layer, we compute its output using the quantized weights, which are then used as input to the next layer. This process is repeated iteratively until all layers are quantized.

%% file: sections/method.tex
\section{Methodology} \label{sec:method}

In this section, we first present the key observation that motivates our approach to paying different attention to each token (\Cref{ssec:motivation}). Next, we provide a detailed outline of each step in \MethodName{} (\Cref{ssec:rsq}). Finally, we describe our strategy for determining the importance of different tokens (\Cref{ssec:strategy_importance}).

\begin{figure*}[t]
    \centering
    \includegraphics[width=\linewidth]{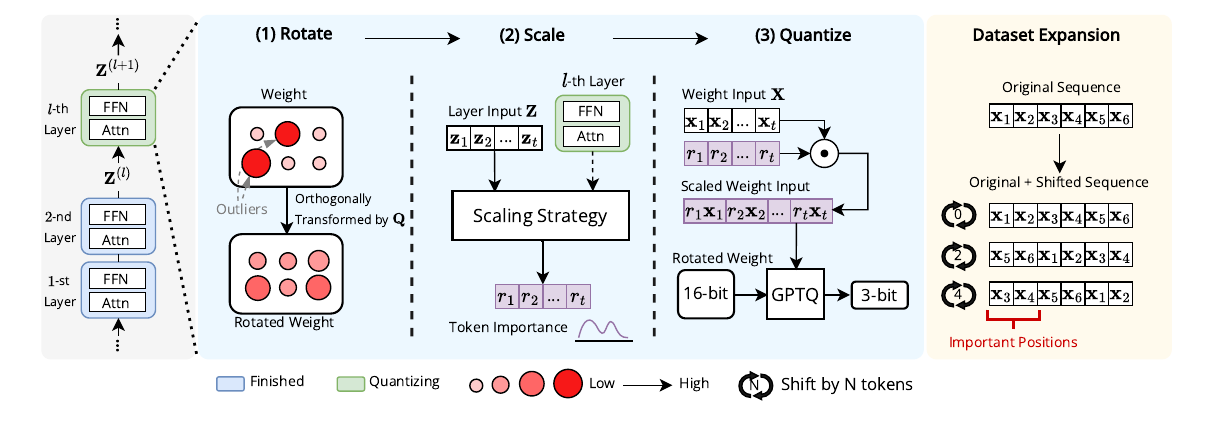}
    \caption{Illustration of layer-wise quantization (left), three-step process of \MethodName{} (middle) and the dataset expansion (right). On the middle, circle size and red color intensity represent weight magnitude, with larger circles and deeper red colors indicating greater magnitudes.
    }
    \label{fig:main_figure}
\end{figure*}

\subsection{Observation: Less Tokens yet Better Performance} \label{ssec:motivation}

As described in the \Cref{ssec:layer-wise quantization}, layer-wise quantization methods, such as GPTQ and QuaRot, uniformly minimize the reconstruction loss across all tokens. In the standard setup, using 256 data points from WikiText-2~\cite{Merity2016wiki2}, each containing 4096 tokens, to quantize LLaMA3-8B-Instruct~\cite{dubey2024llama3} to 3-bit precision, QuaRot achieves a perplexity of $9.51$ on WikiText-2 and an average accuracy of $63.8\%$ across 10 tasks (detailed in \Cref{ssec:compare to baselines}). 

Here, we present a surprising finding: \textbf{using only a subset of tokens can improve performance}. Specifically, we divide the input tokens into four non-overlapping chunks (token IDs: 1–1024, 1025–2048, 2049–3072, 3073–4096), with each chunk containing 1024 tokens. We then perform four separate quantization runs, applying the reconstruction loss to only one chunk at a time. 
It is important to note that while the loss is computed exclusively on the selected chunk, all tokens are used for the forward pass. Therefore, the earlier tokens, even when not selected, still indirectly contribute to the loss of the later tokens. 
However, due to the autoregressive nature of modern LLMs, tokens after the N-th chunk do not contribute to the loss if the reconstruction loss is applied solely to the N-th chunk.
Interestingly, as shown in \Cref{tab:motivation}, the performance when using the 2nd, 3rd, or 4th chunks is inferior to that of the 1st chunk, despite these chunks having access to more input information (as tokens in the 1st chunk are still used to produce features for subsequent tokens). We hypothesize that this effect arises because LLMs tend to heavily attend to ``the initial tokens'', making it crucial to preserve their features by directly minimizing their reconstruction loss. This hypothesis is further supported by broad observations from prior research~~\cite{sun2024massive_attention,Xiao2023attention_sink}.
Moreover, we demonstrate that the quantized LLMs \textbf{using only the 1st chunk even outperforms the result obtained by using all tokens}. We believe this is because the quantized model has to allocate its limited capacity to learn from all tokens, including those that may be less important than the first chunk.

Based on this observation, we argue that the current approach of uniformly applying reconstruction loss to all tokens is suboptimal, as it fails to prioritize and preserve the most critical information in the model after quantization. 
To address this, we modify the layer-wise quantization objective to account for token importance and propose \MethodName{}, which we detail in the next section.

\begin{table}
    \centering
    \caption{Evaluation results of quantizing LLaMA-3-8B-Instruct with WikiText-2 using different subsets of tokens.}
    \label{tab:motivation}
    \resizebox{0.98\linewidth}{!}{
    \begin{tabular}{lccccc}
    \toprule
    \makecell{Used \\Token IDs} & \makecell{All: 1 - \\ 4096}& \makecell{1 - \\1024} & \makecell{1025 - \\2048} & \makecell{2049 - \\3072} & \makecell{3073 - \\4096} \\ 
    \midrule
    Wiki PPL $\downarrow$ & 9.51$_{.11}$ & \textbf{9.27}$_{.05}$ & 10.26$_{.12}$ & 10.16$_{.08}$ & 10.25$_{.01}$ \\
    Avg Acc (\%) $\uparrow$ & 63.8$_{0.5}$ & \textbf{64.5}$_{0.4}$ & 61.7$_{0.7}$ & 61.4$_{0.4}$ & 61.3$_{0.5}$ \\ 
    \bottomrule
    \end{tabular}
    }
\end{table}

\subsection{\MethodName{} (\MethodFullName{})} \label{ssec:rsq}

In the previous section, we demonstrated that a simple approach—selecting only the first chunk—already leads to improved results. Building on this insight, we further explore more advanced strategies for assigning importance to different tokens, which we detail in \Cref{ssec:strategy_importance}.

Before delving into these strategies, let us first formally introduce the algorithm of \MethodName{}, which quantizes the model in three steps: (1) \textit{rotate} to reduce outliers in the weights, (2) \textit{scale} to reweight the input tokens, and (3) \textit{quantize} via GPTQ mechanism with the scaled tokens. Below, we provide a detailed explanation of each step.

\paragraph{Rotate.} 
Before quantization, we first mitigate the outliers by rotating the model.
As mentioned in \Cref{ssec:rotation}, computational invariance holds when RMSNorm is used in transformers. However, existing LLaMA-like LLMs, such as LLaMA3, Mistral, and Qwen2.5, use LayerNorm~\cite{Ba2016LayerNorm} instead. Fortunately, following ~\citet{ashkboos2024slicegpt}, LayerNorm can be converted to RMSNorm by fusing its linear component into the subsequent linear layer.
After this conversion, we initialize 
$\mathbf{Q}$ as a randomized Hadamard matrix~\cite{halko2011finding} and apply the rotation matrix to the transformer's weights as described in \Cref{ssec:rotation}. 

\paragraph{Scale.}
As described in \Cref{ssec:layer-wise quantization}, previous methods treat the layer reconstruction loss of every token equally, that is $\norm{\mathbf{W} \mathbf{X} - \tilde{\mathbf{W}} \mathbf{X}}^2_2 = \sum_{i=0}^{T} \norm{\mathbf{W} \mathbf{x}_i - \tilde{\mathbf{W}} \mathbf{x}_i}_2^ 2$.
In contrast, \MethodName{} assigns different importance to different tokens and modifies the objective function accordingly:
\begin{align} \label{eq:reconstruction_rsq}
    \norm{(\mathbf{W} \mathbf{X} - \tilde{\mathbf{W}} \mathbf{X})\mathbf{R}}^2_2 = \sum_{i=0}^{T} \norm{r_i(\mathbf{W} \mathbf{x}_i - \tilde{\mathbf{W}} \mathbf{x}_i)}_2^ 2
\end{align}
where $\mathbf{R}$ is a diagonal matrix with diagonal entries $(r_1, r_2, ..., r_T)$ that scale the token representations $(\mathbf{x}_1, \mathbf{x}_2, ..., \mathbf{x}_T)$, respectively. The specific methods for assigning values to the importance matrix $\mathbf{R}$ are detailed in the next section.

\input{tables_figures/main_table}

\paragraph{Quantize.}
Given \MethodName{}'s proposed objective function (\Cref{eq:reconstruction_rsq}), as in \Cref{ssec:layer-wise quantization}, we follow the GPTQ framework to solve the optimal quantized weight while minimizing the loss. The resulting formulation remains mostly the same, except for a modified Hessian matrix  $\mathbf{H}_{\MethodName{}} = 2\mathbf{X}\mathbf{R}^2\mathbf{X}^{\top}$, which essentially represents the outer product of the scaled token features (\textit{i.e.} $\mathbf{X}\mathbf{R}$). 
Next, we quantize the rotated weight by applying the modified Hessian matrix to the weight update formula presented in \Cref{eq:gptq}. \Cref{fig:main_figure} displays the illustration of the three steps of \MethodName{}.

\subsection{Strategies to Compute Token Importance} \label{ssec:strategy_importance}

To align with the nature of layer-wise quantization, we compute token importance per layer independently. Furthermore, we avoid using any global information, such as model gradients, as it would violate the layer-wise assumption, where only one layer is accessed at a time.
During the quantization of the $l$-th layer, let $\mathbf{Z}^{(l)} = \{\mathbf{z}_1^{(l)}, \mathbf{z}_2^{(l)}, ..., \mathbf{z}_T^{(l)}\}$ ($\mathbf{z} \in \mathbb{R}^d$) represent the $d$-dimensional input features of the current layer (note that $\mathbf{Z}^{(l+1)} = \texttt{Layer}^{(l)}(\mathbf{Z}^{(l)})$). We compute the token importance $\mathbf{R}^{(l)} = \{r^{(l)}_1, r^{(l)}_2, ..., r^{(l)}_T\}$ ($r \in \mathbb{R}$) to reweight the input feature of the weight in this layer to $\{r^{(l)}_1\mathbf{x}^{(l)}_1, r^{(l)}_2\mathbf{x}^{(l)}_2, ..., r^{(l)}_T\mathbf{x}^{(l)}_T\}$ before applying GPTQ for weight quantization. Note that we use $\mathbf{Z}$ to represent the input features of a ``layer'', distinguishing it from $\mathbf{X}$, which denotes the input features associated with a ``weight.''  The token importance is kept consistent across all weights within a layer, as we observe this yielding better performance. For clarity and simplicity, we omit the superscript for the layer index $l$ in the remaining of this section.

Next, we present several methods for assigning importance to tokens to complete the second step of \MethodName{}. We begin by introducing two \textit{heuristic} approaches that prioritize training on specific positions within the sequence.

\textbf{First-N.} Building on our observation in \Cref{ssec:motivation} that using fewer tokens (specifically, tokens from the first chunk) leads to better performance, we further divide the inputs into smaller chunks, each containing fewer tokens. We then evaluate the quantization performance using only the first chunk. Formally, we define $r_i = 1$ when $i \leq N$, and $r_i = 0$ for the rest.

\textbf{First\&Last-N.} This approach extends \textit{First-N} method. While First-N can be viewed as using the first N/2 tokens and the second N/2 tokens, we explore an alternative approach that uses the first-N/2 and last-N/2 tokens for quantization. We hypothesize that replacing the second chunk with the last chunk may better capture long-term dependencies. Formally, we assign $r_i = 1$ when $i \leq N \text{ or } i > T - N$, and $r_i = 0$ for the rest.

The aforementioned approaches assign token importance based on positional heuristics, ignoring variations across different samples and layers. To address this limitation, we explore several \textit{dynamic} approaches where token importance is determined adaptively based on the layer inputs $\mathbf{Z}$ and model characteristics. We rigorously compare these approaches and adopt the most effective one at the end.

\textbf{Token Frequency (TokenFreq).} This approach assumes that a token's importance is related to its frequency, and we observe that assigning greater weight to \textit{less} frequent tokens yields better results. We compute token frequency based on the calibration dataset used for quantization, denoting the occurrence count of token t as $C(\text{t}) \in \mathbb{R}$. Given the input token sequence $\{\text{t}_1, \text{t}_2, ..., \text{t}_T\}$, we define token importance $\mathbf{R} \in \mathbb{R}^T$ as $\{-C(\text{t}_i) : 1 \leq i \leq T\}$. Finally, we linearly transform the importance values ($\forall r \in \mathbf{R}$) into a bounded range $[r_{min}, r_{max}]$:

\begin{equation} \label{eq: map to range}
    r = r_{min} + \frac{r - \min(\mathbf{R})}{\max(\mathbf{R}) - \min(\mathbf{R})} \cdot (r_{max} - r_{min})
\end{equation}

Here, $r_{min}$ and $r_{max}$ are hyperparameters, and we always set $r_{max}$ to 1 and adjust $r_{min}$ to vary the emphasis on less important tokens. Note that we apply \Cref{eq: map to range} to normalize the scores into a bounded range for \textit{all} dynamic approaches.

\textbf{Activation Norm (ActNorm).} 
Previous studies have shown that inputs with larger norms have a greater impact on the layer's outputs~\cite{virmaux2018lipschitz}. \citet{sun2024massive_attention} also show that attention in LLMs tends to concentrate on tokens with larger norms. 
Based on these, we design an approach to assigning importance scores to tokens based on the norm of their input activations, that is $\mathbf{R} = \{\norm{\mathbf{z}_i} : 1 \leq i \leq T\}$.

\textbf{Activation Difference (ActDiff).} Our next approach defines token importance based on the feature changes between inputs and outputs~\cite{sajjad2023droppinglayers}. We observe that assigning greater weight to tokens with \textit{smaller} changes yields better results compared to assigning greater weight to those with larger changes. This suggests that these ``steady tokens'' play a more crucial role in the model. Specifically, the scores are calculated as $\mathbf{R} = \{- \norm{\texttt{Layer}(\mathbf{z}_i) - \mathbf{z}_i} : 1 \leq i \leq T\}$.

\textbf{Token Similarity (TokenSim).} This approach assigns token importance based on the pairwise similarity between each token and all other tokens. Our assumption is that tokens that are \textit{less} similar (has larger distance) to others are more important, as their information is \textit{rarer} within the sequence. Let $\mathbf{S} \in \mathbb{R}^{T \times T}$, where $\mathbf{S}_{ij} = \norm{\mathbf{z}_i - \mathbf{z}_j}$, denote the $l2$ distance between $i$-th and $j$-th token features. 
Formally, the scores are calculated as $\mathbf{R} = \{\sum_{j}\mathbf{S}_{ij} : 1 \leq i \leq T\}$.

\textbf{Attention Concentration (AttnCon).} 
Moreover, several works have shown that some tokens contribute most of the values in attention maps~\cite{Zhang2023H2O}. Based on this insight, we compute attention concentration to determine token importance. Specifically, consider a multi-head ($M$ heads) attention example in a given layer. Let $\mathbf{A} \in \mathbb{R}^{M \times T \times T}$ represent the attention probability map, where $\mathbf{A}_{mij}$ denotes the proportion of attention $j$-th token receives from the $i$-th token in the $m$-th head of the attention. Due to the autoregressive nature of LLMs, $\mathbf{A}_{mij} = 0$ for $j > i$. To calculate the attention concentration of the $j$-th token, we sum over the second dimension of $\mathbf{A}$, and further sum the scores of every head together. Specifically, the importance score is calculated as $\mathbf{R} = \{\sum_{m,i}\mathbf{A}_{mij} : 1 \leq j \leq T\}$.

Note that the computed scores can be seamlessly integrated into the GPTQ framework as described in \Cref{ssec:rsq} to preserve the efficiency of the algorithm. 
We select AttnCon as our final strategy due to its superior performance and present the comparison and analysis of all methods in \Cref{ssec:evaluating scaling strategies}.

\subsection{Dataset Expansion} \label{ssec:data_expansion}
In our exploration, we find that important tokens tend to be biased toward specific positions. For example, our heuristic methods inherently select tokens from predefined positions. Moreover,
AttnCon consistently assigns higher importance to the initial and final tokens, despite not explicitly enforcing this behavior (\Cref{fig:attncon_visualization}). This bias may lead to inefficiency, as tokens in other positions are significantly overlooked.

To address this, we propose data expansion, a data augmentation technique designed to "shift" tokens within a sequence, ensuring every token can occupy important positions. Specifically, given a token sequence of length $T$ and an expansion factor of $M$ ($=8$ in this paper), we generate shifted versions of the sequence by offsetting it by $T/M$, $2T/M$, $3T/M$, ..., $(M-1)T/M$.
The excessive tokens are then inserted at the beginning of the sequence.
This process effectively distributes token importance more evenly, mitigating positional biases and improving overall token utilization. We illustrate this approach in \Cref{fig:main_figure}.

%% file: tables_figures/main_table.tex
\begin{table*}
    \centering
    \caption{Comparison of recent layer-wise quantization approaches with \MethodName{} on multiple downstream tasks. We report perplexity for WikiText and accuracy for all other tasks. The model is quantized to 3-bit. The best-performing method across all quantization approaches is highlighted in bold. We denote the standard deviation across three runs as a subscript.}
    \label{tab:main result}
    \resizebox{\textwidth}{!}{
    \begin{tabular}{lcccccccccccm}
    \toprule
    Method & Wiki & $\text{LAMBADA}_{openai}$	 & $\text{LAMBADA}_{std}$ & WinoGrande & ArcC & ArcE &
    HellaSwag & PIQA & MMLU & GSM8k &  TruthfulQA & Avg \\
    \midrule
    \multicolumn{13}{c}{LLaMA3-8B-Instruct} \\
    \midrule
    Full Model & 8.311 & 71.9 & 65.0 & 71.7 & 56.7 & 79.6 & 75.8 & 78.5 & 65.6 & 79.9 & 51.7 & 69.7 \\
    GPTQ & 10.682$_{.04}$ & 50.7$_{2.0}$ & 44.6$_{1.0}$ & 65.7$_{0.7}$ & 35.8$_{1.4}$& 56.9$_{2.3}$ & 65.8$_{0.2}$ & 70.4$_{1.7}$ & 50.8$_{0.4}$ & 27.5$_{1.5}$ & 44.8$_{1.0}$ & 51.3$_{0.7}$ \\
    QuaRot & 9.517$_{.11}$ & 68.8$_{1.2}$ & 59.7$_{1.0}$ & 70.2$_{1.2}$ & 49.6$_{1.5}$ & 74.9$_{0.9}$ & 71.3$_{0.0}$ & 76.8$_{0.7}$ & 57.6$_{0.5}$ & 61.2$_{1.7}$ & 48.0$_{0.9}$ & 63.8$_{0.5}$ \\
    \MethodName{} & \textbf{9.046}$_{.01}$ & \textbf{70.8}$_{0.3}$ & \textbf{62.3}$_{0.4}$ & \textbf{70.6}$_{0.7}$ & \textbf{50.3}$_{1.1}$ & \textbf{76.5}$_{1.3}$ & \textbf{72.1}$_{0.2}$ & \textbf{77.1}$_{0.5}$ & \textbf{60.0}$_{0.3}$ & \textbf{63.4}$_{2.4}$ & \textbf{50.6}$_{1.7}$ & \textbf{65.4}$_{0.1}$ \\
    \midrule
    \multicolumn{13}{c}{Mistral-NeMo-12B} \\
    \midrule
    Full Model & 6.095 & 75.8 & 68.3 & 75.1 & 59.1 & 80.1 & 82.2 & 82.1 & 68.2 & 80.8 & 54.8 & 72.7 \\
    GPTQ & 9.537$_{.01}$ & 44.6$_{0.6}$ & 29.8$_{4.3}$ & 57.3$_{1.1}$ & 38.8$_{0.2}$ & 58.0$_{1.6}$ & 60.5$_{2.0}$ & 71.2$_{0.6}$ & 48.0$_{1.1}$ & 37.6$_{1.5}$ & 47.0$_{1.5}$ & 49.3$_{0.4}$ \\
    QuaRot & 6.782$_{.00}$ & \textbf{75.6}$_{0.3}$ & 60.9$_{5.8}$ & 72.8$_{0.7}$ & \textbf{55.8}$_{1.5}$ & 76.9$_{1.1}$ & 78.1$_{0.3}$ & 79.7$_{0.4}$ & 63.5$_{0.0}$ & \textbf{71.1}$_{1.0}$ & 52.5$_{0.9}$ & 68.7$_{1.1}$ \\
    \MethodName{} & \textbf{6.673}$_{.01}$ & 75.4$_{0.3}$ & \textbf{66.5}$_{0.6}$ & \textbf{73.5}$_{0.4}$ & 55.7$_{0.5}$ & \textbf{77.2}$_{1.1}$ & \textbf{78.5}$_{0.1}$ & \textbf{80.7}$_{0.2}$ & \textbf{64.3}$_{0.3}$ & \textbf{71.1}$_{0.4}$ & \textbf{52.9}$_{0.5}$ & \textbf{69.6}$_{0.2}$ \\
    \midrule
    \multicolumn{13}{c}{Qwen-2.5-7B-Instruct} \\
    \midrule
    Full Model & 5.335 & 75.2 & 68.6 & 72.7 & 59.0 & 76.4 & 85.2 & 81.0 & 83.3 & 84.4 & 65.5 & 75.1 \\
    GPTQ & 9.577$_{.03}$ & 53.6$_{1.0}$ & 48.6$_{1.8}$ & 62.1$_{0.9}$ & 47.1$_{1.8}$ & 66.9$_{2.1}$ & 72.8$_{0.0}$ & 73.7$_{0.5}$ & 60.2$_{0.5}$ & 40.9$_{10.5}$ & 53.7$_{1.5}$ & 58.0$_{1.8}$ \\
    QuaRot & 8.053$_{.02}$ & 67.9$_{0.3}$ & 62.1$_{0.4}$ & 68.3$_{1.2}$ & \textbf{54.0}$_{1.6}$ & \textbf{78.7}$_{2.9}$ & 76.7$_{0.4}$ & \textbf{78.8}$_{0.8}$ & 68.7$_{0.6}$ & 76.4$_{0.7}$ & 60.0$_{1.4}$ & 69.2$_{0.5}$ \\
    \MethodName{} & \textbf{8.051}$_{.01}$ & \textbf{68.7}$_{0.2}$ & \textbf{63.5}$_{0.5}$ & \textbf{68.5}$_{0.6}$ & 53.7$_{1.7}$ & 78.2$_{2.0}$ & \textbf{77.1}$_{0.3}$ & \textbf{78.8}$_{1.0}$ & \textbf{69.0}$_{0.4}$ & \textbf{77.1}$_{0.8}$ & \textbf{61.2}$_{0.9}$ & \textbf{69.6}$_{0.3}$ \\
    \bottomrule
    \end{tabular}
    }
\end{table*}

%% file: sections/experiments.tex
\section{Experiments} \label{sec:experiments}

\input{tables_figures/ablation_on_importance_measure}

We first compare \MethodName{} with recent layer-wise quantization methods on eleven tasks across different model families (\Cref{ssec:compare to baselines}). Next, we evaluate several design choices in \MethodName{}, such as various scaling strategies and data expansion (\Cref{ssec:evaluating scaling strategies}). We then study the effect of \MethodName{} on long-context tasks (\Cref{ssec:long-context tasks evaluation}). Lastly, we assess the generalizability of \MethodName{} (\Cref{ssec:generalizability of rsq}) across various model sizes, calibration datasets, bit precisions, and quantization methods. Note that we quantize models to 3-bit if not further specified and \textbf{our experiments are conducted using three different seeds}.

\subsection{Comparison of \MethodName{} Against Baselines}\label{ssec:compare to baselines}

\textbf{Setup.} 
We evaluate \MethodName{} and other baselines across three model families: LLaMA3-8B-Instruct~\cite{dubey2024llama3}, Mistral-NeMo-12B~\cite{Jiang2023Mistral7}, and Qwen-2.5-7B-Instruct~\cite{Yang2024Qwen25TR}. We quantize each model to 3-bit on WikiText-2 with 256 data samples, each with 4096 tokens, which is the setup adopted in recent studies~\cite{Egiazarian2024aqlm,tseng2024quip_sharp}.
We use AttnCon as the scaling strategy and set the data expansion factor $M$ to 8.

The quantized models are tested on a diverse set of tasks, including LAMBADA (with two splits: the original paper's version and OpenAI's version)~\cite{Paperno2016LAMBADA}, WinoGrande~\cite{Sakaguchi2019WinoGrande} and ARC (Challenge and Easy splits)~\cite{Clark2018Arc}, HellaSwag~\cite{Zellers2019HellaSwagCA}, PIQA~\cite{Bisk2019PIQARA}, MMLU~\cite{Hendrycks2020MMLU}, GSM8k~\cite{Cobbe2021GSM8k}, and TruthfulQA~\cite{Lin2021TruthfulQAMH}. Accuracy is used as the evaluation metric.

We use GPTQ and QuaRot as the baselines in this study. GPTQ performs layer-wise quantization as described in \Cref{ssec:layer-wise quantization}. QuaRot, on the other hand, mitigates outliers by applying rotations (detailed in \Cref{ssec:rotation}), followed by applying GPTQ to quantize the rotated model.

\textbf{Results.} 
\Cref{tab:main result} presents the evaluation results for the 16-bit and 3-bit quantized models using GPTQ, QuaRot, and \MethodName{}. GPTQ demonstrates a notable performance gap compared to QuaRot and \MethodName{}, primarily due to the negative impact of outliers in the model, which degrade the quantization quality. When comparing \MethodName{} to QuaRot, our approach achieves superior results in Wiki Perplexity and most evaluation tasks across the three models, where \MethodName{} achieves absolute improvements of 1.6\%, 0.9\%, and 0.4\% in average accuracy against QuaRot, respectively. This indicates that incorporating token importance into the quantization process is an effective strategy, even when using the same number of total tokens.

\input{tables_figures/long_context}

\subsection{Evaluating Design Choices in \MethodName{}}\label{ssec:evaluating scaling strategies}

\textbf{Setup.} We follow the same setup described in \Cref{ssec:motivation} to quantize LLaMA3-8B-Instruct to 3-bit precision using 256 samples of 4096 tokens each from WikiText-2. \textbf{To avoid overfitting, we only use the perplexity on WikiText-2 as the evaluation metric} in this part.

\textbf{Results.} We first compare the performance of two heuristic approaches, First-N and First\&Last-N, across different numbers of activated tokens. As shown in \Cref{fig:first_last_n}, we observe that perplexity decreases steadily as the number of tokens is reduced from 4096 to around 512 or 256, but increases when using fewer tokens for both approaches. This suggests that \textbf{using the fewest tokens does not necessarily yield the best results}. We hypothesize that the model requires a certain number of tokens to effectively capture token interactions that are brought by the attention mechanism. 

We also observe that First\&Last-N often outperforms First-N when using the same number of tokens, achieving their optimal perplexity values of 9.15 and 9.18, respectively. This suggests that \textbf{learning is more effective when incorporating the last chunk of tokens into the first chunk, rather than using the first and middle chunks}. 

In \Cref{fig:adaptive_method}, we compare five dynamic approaches, TokenFreq, ActNorm, ActDiff, TokenSim and AttnCon, with varying hyperparameter $r_{min}$ across $\{0.005, 0.01, 
0.02, 0.05, 0.1\}$. We observe that TokenFreq and ActDiff perform less competitively than the other approaches, suggesting that token frequency and feature changes after a layer are not strong indicators of token importance for quantizing LLMs.
Among the other three approaches, we find AttnCon reaches its optimal perplexity (9.028) at $r_{min} = 0.01$ while ActNorm and TokenSim reach their optimal perplexity (9.075, 9.047, respectively) at $r_{min} = 0.005$. These relatively small $r_{min}$ indicate that \textbf{placing lower emphasis on less important tokens and focusing more on the most important tokens is beneficial}. We adopt AttnCon as our final scaling strategy, as it achieves the best perplexity performance.

We also assess the effectiveness of data expansion ($M=8$) by incorporating it into each approach using its optimal hyperparameter. As shown in \Cref{fig:expanding}, most scaling strategies benefit from data expansion in terms of perplexity.

\input{tables_figures/all_ablation_study_table}

\subsection{Evaluation on Long-Context Tasks}\label{ssec:long-context tasks evaluation}
Previous approaches have primarily focused on tasks with relatively short inputs. However, recent large language models (LLMs) have demonstrated impressive long-context capabilities, greatly expanding their potential for user-facing applications such as chatbots, search engines, and collaborative code-writing systems. Neglecting the evaluation of long-context benchmarks may fail to capture the full impact and effectiveness of LLM quantization, particularly in these emerging use cases. To foster a complete assessment, in this section, we thus evaluate \MethodName{} and QuaRot against several long-context benchmarks.

\textbf{Setup.} 
We use LLaMA3 as the backbone model for this experiment. Since this model has a context length limit of 8k tokens, the dataset samples used for evaluation do not exceed this length. When evaluating long-context tasks, a natural question arises: \textit{does using longer input sequences in the calibration dataset improve long-context performance?} To explore this, we test three different configurations of the calibration dataset. Specifically, we set the number of samples to 256, 512, and 1024, with corresponding sequence lengths of 4096, 2048, and 1024 tokens, respectively. Note that we adjust the number of samples to ensure the total number of tokens remained consistent across configurations. 

Lost in the Middle (LITM)~\cite{liu2024litm} is a retrieval task designed to assess positional biases of the answer placed in the input documents. We set $P = 1, 15, 30$ to indicate that the answer appears in the $P$-th document out of a total of 30 documents (avg length=4.5k).
Next, we adopt some closed-ended tasks from L-Eval \cite{An2024LEval} for evaluating the long-context understanding, such as TOEFL (avg length=3.6k) \cite{Tseng2016TOEFL, chung2018TOEFL}, QuALITY (6.2k) \cite{pang2022quality}, Coursera (6.8k), SFiction (7.2k), GSM (4.8k), CodeU (7.4k), and TopicRet (7.6k) \cite{li2023longeval}.
LongICLBench \cite{Li2024LongICLBench} focuses on evaluating the long-context in-context learning capabilities of LLMs. From this benchmark, we sample two datasets: Banking77 (avg length=7.7k) \cite{Casanueva2020Banking77} and TecRED (6.6k) \cite{zhang2017TecRED}. LongCodeArena \cite{Bogomolov2024LongCodeArena} is a benchmark for code processing tasks that require project-wide context, and we sample the library-based code generation task to evaluate the model's ability to utilize the given library.

Most of the datasets employ accuracy as evaluation metrics, except we use the F1 score for TecRED and ChrF \cite{popovic2015chrf} for library code generation in LongCodeArena.

\textbf{Results.} 
\Cref{tab:long context} displays our evaluation of long-context tasks using three different calibration dataset configurations. \MethodName{} consistently outperforms QuaRot across nearly all benchmarks in all three configurations. This demonstrates that the strategy of prioritizing important tokens produces quantized models that perform effectively on both short- and long-context tasks. Furthermore, the results indicate that \textbf{focusing on a subset of tokens is sufficient to capture the long-term dependencies across tokens}.

We do not observe a clear trend indicating that using a calibration dataset with longer sequences results in better or worse performance on long-context tasks. \textbf{This suggests that simply matching the length distribution between the calibration dataset and downstream tasks is insufficient}. A more advanced strategy, whether through improving the data or the method sides, is needed to further enhance performance in long-context scenarios. 

In the LITM evaluation, we observe that LLaMA3-8B-Instruct generally performs better when the answer appears in the early documents ($P=1$). However, its performance on later documents does not consistently surpass that on middle documents, deviating from the findings reported by \citet{liu2024litm} for other models.
Interestingly, we also find that the quantized models outperform the 16-bit model by 2–4\% in average accuracy on LongICLBench, similar to the findings from \citet{hong2024DecodingCompressedTrust}. This shows that quantization does not necessarily hurt performance in every aspect and can yield improvements in certain scenarios. One possible explanation is that quantization might reduce weight noise, potentially enhancing the model’s robustness for some tasks.

\subsection{Generalizability of \MethodName{}} \label{ssec:generalizability of rsq}

In this section, we evaluate the generalizability of \MethodName{} under various setups, including its performance across different model sizes, calibration datasets, bit precisions, and quantization methods.

\textbf{Different Model Sizes.} We choose three models from the mistral family: Mistral-7B-Instruct-v0.3, Mistral-NeMo-12B and Mistral-Small-Instruct-2409, whose sizes are 7B, 12B, and 22B, respectively. We quantize each model to 3-bit on WikiText-2 with 256 data samples, each with 4096 tokens. The results (\Cref{fig:model scaling}) show that \MethodName{} consistently outperforms QuaRot across all three models, with the performance gap being slightly larger in the 22B model 

\textbf{Different Calibration Datasets.} In addition to using WikiText-2 as the calibration dataset~\cite{bandari2024c4,ji2024beware}, we also evaluate \MethodName{}'s performance with using RedPajama~\cite{weber2024redpajama}, C4~\cite{Raffel2019c4andt5}, and PTB~\cite{marcus1993ptb}. For this experiment, we quantize LLaMA3 to 3-bit using 512 data samples, each with 2048 tokens. The results, presented in \Cref{tab:calibration dataset}, demonstrate that our approach consistently outperforms QuaRot across varying calibration datasets, demonstrating the robustness of the approach.

\textbf{Different Bit Precisions.} In previous experiments, we consistently quantized the model to 3-bit. Here, we extend the analysis by exploring the effects of \MethodName{} when quantizing to 2-bit and 4-bit. When quantizing LLaMA3 with WikiText-2 (512 data samples, each containing 2048 tokens), our results in \Cref{tab:bits} demonstrate that \MethodName{} outperforms QuaRot across different bit precisions. Notably, the performance gap is larger at lower bit precisions, suggesting that the idea of "learning from important tokens" can be a crucial factor for achieving effective extreme compression.

\begin{wraptable}[5]{r}{0.5\linewidth}
    \centering
    \vspace{-0.18in}
    \vspace{-0.1in}
    \caption{\MethodName{} + VQ.}
    \label{tab:vector quantization}
    \resizebox{\linewidth}{!}{
    \begin{tabular}{lcc}
    \toprule
    \multirow{2}[2]{*}{Method} & \multicolumn{2}{c}{Metrics} \\
    \cmidrule{2-3}
    & \makecell{Wiki PPL $\downarrow$} & \makecell{Avg Acc (\%) $\uparrow$}\\ 
    \midrule
    QuaRot & 24.69$_{0.6}$ & 42.5$_{1.0}$ \\
    \MethodName{} & \textbf{20.08}$_{0.3}$ & \textbf{44.3}$_{0.2}$ \\
    \bottomrule
    \end{tabular}
    }
\end{wraptable}
\textbf{\MethodName{} for vector quantization.} In previous experiments, we quantized model weights individually using scalar quantization. In this section, we extend \MethodName{} to vector quantization (VQ), which better captures the high-dimensional distribution of weights. Specifically, we replace the 2-bit integer grid in scalar quantization with the 2-bit comparable E8P codebook~\cite{tseng2024quip_sharp} and adapt the quantizer from GPTQ to LDLQ, following the original implementation, as the two are shown to be equivalent in the QuIP paper~\cite{Chee2023QuIP}. We then quantize LLaMA3 using WikiText-2 (512 data samples, each with 2048 tokens) under this new setup and present the results in \Cref{tab:vector quantization}. Our findings indicate that vector quantization improves the average accuracy for both QuaRot and \MethodName{} compared to scalar quantization (2-bit results in \Cref{tab:bits}), with our approach achieving the best overall performance.

%% file: tables_figures/ablation_on_importance_measure.tex
\begin{figure*}[t]
    \begin{minipage}{.34\linewidth}
        \centering
        \includegraphics[width=\linewidth]{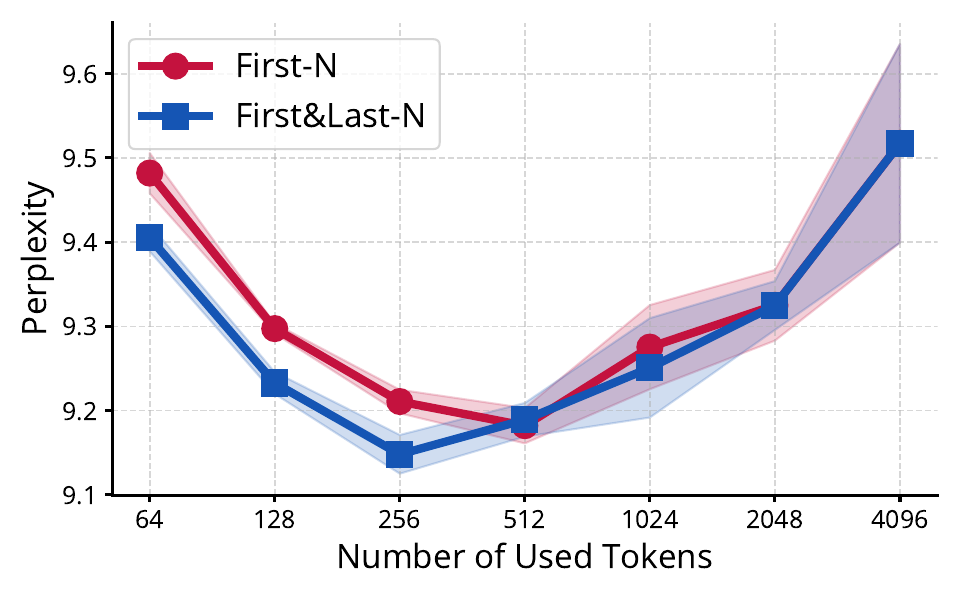}
        \vspace{-20pt}
        \caption{Evaluation of two heuristic approaches with varying numbers of used tokens.}
        \label{fig:first_last_n}
    \end{minipage}
    \hfill
    \begin{minipage}{.35\linewidth}
        \centering
        \includegraphics[width=\linewidth]{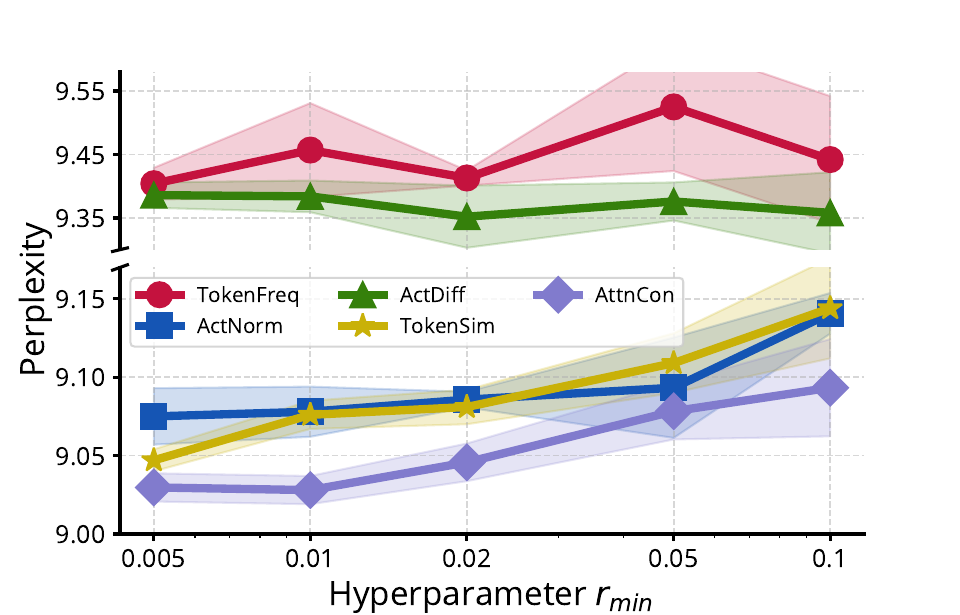}
        \vspace{-20pt}
        \caption{Evaluation of five dynamic approaches with varying $r_{min}$.}
        \label{fig:adaptive_method}
    \end{minipage}
    \hfill
    \begin{minipage}{.26\linewidth}
        \centering
        \vspace{22pt}
        \includegraphics[width=\linewidth]{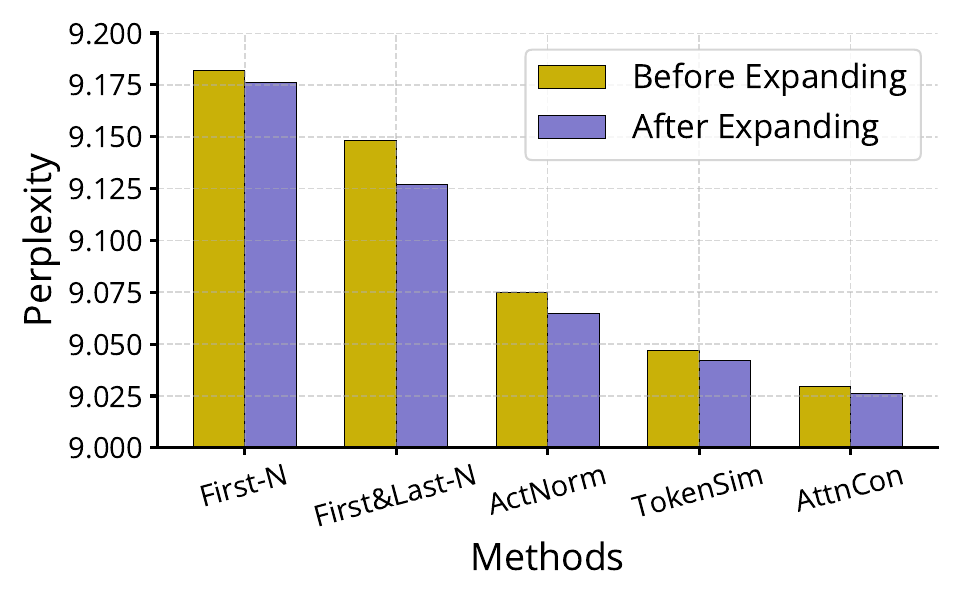}
        \vspace{-5pt}
        \vspace{-10pt}
        \caption{The effect of expanding the dataset on different methods.}
        \label{fig:expanding}
    \end{minipage}
    \vspace{-7pt}
\end{figure*}

%% file: tables_figures/long_context.tex
\begin{table*}
    \centering
    \caption{Comparison of \MethodName{} and QuaRot across multiple long-context benchmarks using three different calibration dataset configurations. The model is quantized to 3-bit. The best-performing method among all quantization approaches is highlighted in bold.}
    \label{tab:long context}
    \resizebox{\textwidth}{!}{
    \begin{tabular}{lcccmcccccccmccmc}
    \toprule
    \multirow{2}[2]{*}{Method} & \multicolumn{4}{c}{LITM} & \multicolumn{8}{c}{L-Eval} & \multicolumn{3}{c}{LongICLBench} & \multicolumn{1}{c}{LongCodeArena} \\
    \cmidrule{2-17}
     & $P$=1 & $P$=15 & $P$=30 & Avg & TOEFL & QuALITY & Coursera & SFiction & GSM & CodeU & TopicRet & Avg & Banking77 & TecRED & Avg & CodeGen \\
    \midrule
    Full Model & 53.67 & 45.61 & 46.33 & 48.54 & 81.04 & 60.40 & 52.62 & 71.88 & 81.00 & 4.44 & 64.67 & 59.43 & 59.40 & 41.65 & 50.52 & 0.298 \\
    \midrule
    \multicolumn{17}{c}{number of samples = 256, sequence length = 4096} \\
    \midrule
    QuaRot & 50.80$_{1.6}$ & 43.47$_{2.4}$ & 43.29$_{2.5}$ & 45.85$_{2.0}$ & 72.24$_{0.9}$ & 52.15$_{0.8}$ & 46.90$_{0.7}$ & \textbf{65.62}$_{3.1}$ & 63.33$_{6.6}$ & 2.22$_{0.9}$ & 45.11$_{3.5}$ & 49.65$_{1.4}$ & \textbf{62.40}$_{4.3}$ & 44.76$_{5.2}$ & 53.58$_{3.2}$ & 0.206$_{.013}$ \\
    \MethodName{} & \textbf{52.23}$_{0.5}$ & \textbf{46.03}$_{2.7}$ & \textbf{45.59}$_{3.5}$ & \textbf{47.95}$_{2.0}$ & \textbf{76.21}$_{2.1}$ & \textbf{54.46}$_{0.7}$ & \textbf{49.27}$_{1.3}$ & 63.80$_{4.0}$ & \textbf{66.33}$_{3.8}$ & \textbf{4.07}$_{0.5}$ & \textbf{50.89}$_{2.4}$ & \textbf{52.14}$_{0.3}$ & 58.60$_{7.7}$ & \textbf{49.83}$_{0.7}$ & \textbf{54.21}$_{3.5}$ & \textbf{0.231}$_{.008}$\\
    \midrule
    \multicolumn{17}{c}{number of samples = 512, sequence length = 2048} \\
    \midrule
    QuaRot & 49.17$_{1.2}$ & 44.32$_{1.0}$ & 43.51$_{0.5}$ & 45.67$_{0.2}$ & 71.13$_{2.3}$ & 51.16$_{1.4}$ & 46.95$_{0.7}$ & 58.33$_{0.9}$ & 65.67$_{4.9}$ & \textbf{4.81}$_{2.1}$ & 48.00$_{7.0}$ & 49.43$_{2.0}$ &  \textbf{65.40}$_{0.6}$ & 45.48$_{3.1}$ & \textbf{55.44}$_{1.8}$ & 0.205$_{.007}$ \\
    \MethodName{} & \textbf{51.39}$_{1.6}$ & \textbf{45.89}$_{0.8}$ & \textbf{45.87}$_{1.2}$ & \textbf{47.72}$_{0.8}$ & \textbf{73.23}$_{2.2}$ & \textbf{53.96}$_{1.7}$ & \textbf{48.79}$_{1.0}$ & \textbf{61.98}$_{2.6}$ & \textbf{70.00}$_{2.1}$ & 4.44$_{0.0}$ & \textbf{51.78}$_{7.0}$ & \textbf{52.02}$_{1.9}$ & 62.33$_{2.2}$ & \textbf{46.17}$_{5.5}$ & 54.25$_{3.8}$ & \textbf{0.227}$_{.002}$ \\
    \midrule
    \multicolumn{17}{c}{number of samples = 1024, sequence length = 1024} \\
    \midrule
    QuaRot & 48.66$_{2.6}$ & 47.20$_{0.9}$ & \textbf{48.80}$_{1.7}$ & 48.22$_{1.4}$ & 73.11$_{2.1}$ & 51.82$_{1.4}$ & 48.64$_{1.8}$ & 62.24$_{1.3}$ & 65.00$_{0.8}$ & 2.59$_{1.3}$ & 40.45$_{3.0}$ & 49.12$_{0.9}$ & 59.73$_{8.0}$ & 46.02$_{7.0}$ & 52.87$_{6.9}$  & 0.220$_{.001}$ \\
    \MethodName{} & \textbf{51.10}$_{0.2}$ & \textbf{48.01}$_{0.7}$ & 48.62$_{1.3}$ & \textbf{49.24}$_{0.2}$ & \textbf{74.72}$_{2.8}$ & \textbf{55.78}$_{2.0}$ & \textbf{50.58}$_{0.8}$ & \textbf{64.85}$_{3.3}$ & \textbf{69.33}$_{1.2}$ & \textbf{3.70}$_{0.5}$ & \textbf{47.55}$_{7.4}$ & \textbf{52.35}$_{2.0}$ & \textbf{61.20}$_{5.2}$ & \textbf{46.30}$_{1.9}$ & \textbf{53.75}$_{3.5}$ & \textbf{0.225}$_{.002}$ \\
    \bottomrule
    \end{tabular}
    }
\end{table*}

%% file: tables_figures/all_ablation_study_table.tex
\begin{table*}
    \begin{minipage}{0.25\linewidth}
        \centering
        \includegraphics[width=0.98\linewidth]{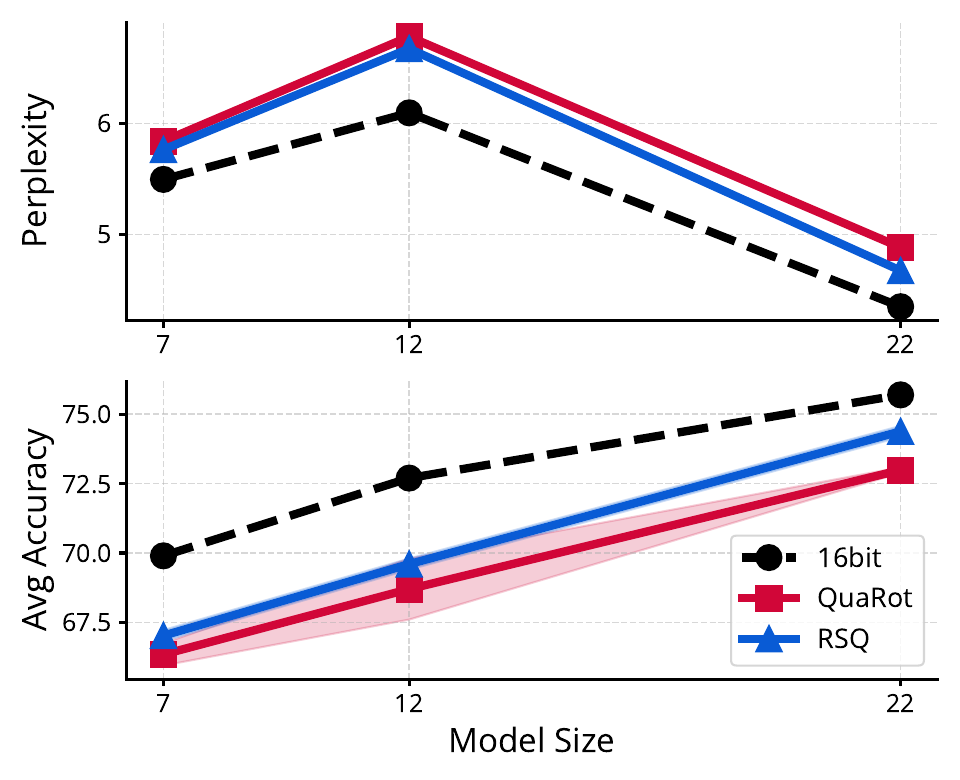}
        \vspace{-12pt}
        \captionof{figure}{Ablation on model sizes.}
        \label{fig:model scaling}
    \end{minipage}
    \begin{minipage}{0.41\linewidth}
        \centering
        \caption{Ablation on calibration dataset.}
        \label{tab:calibration dataset}
        \resizebox{0.98\linewidth}{!}{
        \begin{tabular}{lccccc}
        \toprule
        \multirow{2}[2]{*}{Metric} & \multirow{2}[2]{*}{Method} & \multicolumn{4}{c}{Calibration Dataset} \\
        \cmidrule{3-6}
        & & \makecell{Wiki} & \makecell{RedPajama} & \makecell{C4} & \makecell{PTB} \\ 
        \midrule
        \multirow{2}{*}{Wiki PPL $\downarrow$} & QuaRot & 9.34$_{.03}$ & 9.77$_{.09}$ & 9.90$_{.02}$ & 9.98$_{.06}$\\
        & \MethodName{} & \textbf{9.00}$_{.01}$ & \textbf{9.31}$_{.03}$ & \textbf{9.41}$_{.03}$  & \textbf{9.61}$_{.08}$ \\
        \midrule
        \multirow{2}{*}{Avg Acc (\%) $\uparrow$} & QuaRot & 64.1$_{0.2}$ & 64.2$_{0.3}$ & 64.0$_{0.2}$ & 63.7$_{0.2}$  \\ 
        & \MethodName{} & \textbf{65.1}$_{0.2}$ & \textbf{65.5}$_{0.3}$ & \textbf{65.6}$_{0.3}$ & \textbf{64.5}$_{0.1}$  \\  
        \bottomrule
        \end{tabular}
        }
    \end{minipage}
    \begin{minipage}{0.34\linewidth}
        \centering
        \caption{Ablation on number of bits}
        \label{tab:bits}
        \resizebox{0.98\linewidth}{!}{
        \begin{tabular}{lcccc}
        \toprule
        \multirow{2}[2]{*}{Metric}  & \multirow{2}[2]{*}{Method} & \multicolumn{3}{c}{Number of Bits} \\
        \cmidrule{3-5}
        & & \makecell{4} & \makecell{3} & \makecell{2}\\ 
        \midrule
        \multirow{2}{*}{Wiki PPL $\downarrow$} & QuaRot & 8.57$_{.06}$ & 9.34$_{.03}$ & 22.71$_{0.5}$ \\
        & \MethodName{} & \textbf{8.47}$_{.01}$ & \textbf{9.00}$_{.01}$ & \textbf{16.26}$_{.12}$ \\
        \midrule
        \multirow{2}{*}{Avg Acc (\%) $\uparrow$} & QuaRot & 68.1$_{0.0}$ &  64.1$_{0.2}$ & 35.3$_{0.8}$ \\ 
        & \MethodName{} & \textbf{68.3}$_{0.1}$ & \textbf{65.1}$_{0.2}$ & \textbf{40.6}$_{0.3}$ \\ 
        \bottomrule
        \end{tabular}
        }
    \end{minipage}
    \vspace{-5pt}
\end{table*}

%% file: sections/conclusion.tex
\section{Conclusion} \label{sec:conclusion}

This paper first presents an observation that quantizing models based on only the first 25\% of tokens leads to improved performance compared to using all tokens. This insight motivated us to modify the GPTQ objective and develop \MethodName{} (\MethodFullName{}), which demonstrates effectiveness across diverse benchmarks and configurations.

%% file: sections/impact_statement.tex
\section{Impact Statements} \label{sec:impact_statement}

This paper presents work whose goal is to advance the field of Machine Learning. There are many potential societal consequences of our work, none of which we feel must be specifically highlighted here.

%% file: sections/acknowledgement.tex
\section*{Acknowledgement}

This work was supported
by NSF-CAREER Award 1846185, DARPA ECOLE Program No. HR00112390060, and NSF-AI Engage Institute
DRL-2112635. Any opinions, findings, and conclusions or
recommendations in this work are those of the author(s) and
do not necessarily reflect the views of the sponsors.

%% file: sections/appendix.tex
\appendix

\newpage

\section{Extented Task Details} \label{sec:extended task details}

\subsection{Downstream Task Details}

The quantized models are tested on a diverse set of tasks, including LAMBADA (with two splits: the original paper's version and OpenAI's version)~\cite{Paperno2016LAMBADA} for word prediction, WinoGrande~\cite{Sakaguchi2019WinoGrande} and ARC (Challenge and Easy splits)~\cite{Clark2018Arc} for commonsense reasoning, HellaSwag~\cite{Zellers2019HellaSwagCA} for commonsense natural language inference, PIQA~\cite{Bisk2019PIQARA} for physical commonsense reasoning, MMLU~\cite{Hendrycks2020MMLU} as a comprehensive knowledge benchmark, GSM8k~\cite{Cobbe2021GSM8k} for grade school math, and TruthfulQA~\cite{Lin2021TruthfulQAMH} to assess the model's truthfulness in generation. All datasets use accuracy as the evaluation metric.

\subsection{Long-context Benchmark Details} \label{ssec: extended long-context benchmark details}

Lost in the Middle (LITM)~\cite{liu2024litm} is a retrieval task designed to assess positional biases of the answer placed in the input documents. We set $P = 1, 15, 30$ to indicate that the answer appears in the $P$-th document out of a total of 30 documents. LongEval dataset \cite{li2023longeval} includes a synthetic retrieval task, where each line consists of a key-value pair. Given an input sample of $L$ lines, the model is asked to extract the value corresponding to a specified key in the query. The hyperparameter $L$ controls both the input length and the task complexity. We set $L$ = 300, 460, and 620, where the corresponding input lengths are around 4k, 6k, and 8k, respectively. Next, we adopt a couple of closed-ended tasks from L-Eval \cite{An2024LEval} for evaluating the long-context understanding. TOEFL (avg length=3.6k) \cite{Tseng2016TOEFL, chung2018TOEFL}, QuALITY (6.2k) \cite{pang2022quality}, and Coursera (6.8k) are multiple-choice QA tasks, while SFiction (7.2k) is a True/False QA task. GSM (4.8k) evaluates the model's in-context learning ability, CodeU (7.4k) assesses its capability to deduce program outputs, and TopicRet (7.6k) \cite{li2023longeval} is designed as a retrieval task. 
LongICLBench \cite{Li2024LongICLBench} focuses on evaluating the long-context in-context learning capabilities of LLMs. From this benchmark, we sample two datasets: Banking77 (avg length=7.7k) \cite{Casanueva2020Banking77} and TecRED (6.6k) \cite{zhang2017TecRED}. These datasets primarily test the model's ability to learn and generalize a large number of concepts using only a few-shot examples. LongCodeArena \cite{Bogomolov2024LongCodeArena} is a benchmark for code processing tasks that go beyond a single file and require project-wide context. 
We sample the library-based code generation task to evaluate the ability of the model to solve tasks by using a given library.

\section{Extented Related Work} \label{sec:extended related work}

Recent state-of-the-art open-source LLMs~\cite{Yang2024Qwen25TR,dubey2024llama3,Jiang2024MixtralOE} typically exceed 50 billion parameters. While their released weights make them accessible,
their massive weight size poses significant challenges for practitioners and limits the feasibility of fine-tuning or even inference with these models.
Pruning and ~\cite{Frantar2023SparseGPTML,sun2023wanda,Sung2024ECoFLaP} and quantization are common strategies to compress these models' weight, and this paper focuses on quantization as it is often more effective~\cite{Kuzmin2023pruning_vs_quantization}. 

Post-training quantization (PTQ) has gained significant attention for its ability to quantize pre-trained models without requiring expensive retraining.
Methods such as ZeroQuant~\cite{Yao2022ZeroQuantEA} and QLoRA~\cite{Dettmers2023QLoRAEF} apply round-to-nearest techniques to the weights, even without utilizing calibration datasets. While these approaches are extremely efficient, they often yield suboptimal performance due to their lack of information about the data distribution.
To address this limitation, several PTQ methods leverage calibration datasets for improved quantization. These data-dependent PTQ methods are also often referred to as layer-wise quantization methods, as they typically quantize models one layer at a time for efficiency.
Specifically, GPTQ~\cite{Frantar2022GPTQ} and OBC~\cite{frantar2022obc} quantize weights and adjust the remaining weights with data-derived Hessian matrices (second-order statistics) accordingly. 
AWQ~\cite{Lin2023AWQAW} minimizes quantization error by rescaling weights based on the activation distribution.
QuIP\#~\cite{tseng2024quip_sharp},  AQLM~\cite{Egiazarian2024aqlm} and QTIP~\cite{tseng2025qtip} represent groups of quantized weights with vectors instead of scalar values. Additionally, several approaches focus on fine-tuning parameters introduced during quantization (\textit{e.g.}, quantization indices, group scales, and group zeros), such as OmniQuant~\cite{Shao2023OmniQuantOC}, HQQ~\cite{badri2023hqq}, and PV-Tuning~\cite{malinovskii2024pvtuning}.

One challenge in weight quantization is the presence of outliers in the weights, as LLMs often contain a subset of weights with exceptionally large magnitudes~\cite{dettmers2024spqr,Kim2023SqueezeLLMDQ,Yu2024super_weight}. During quantization, these outliers increase the range of the quantized weights. As a result, most weights are forced into a narrow range to fit the outliers within the quantized representation, which ultimately leads to suboptimal performance. 
One way to address this challenge is through mixed-precision quantization~\cite{dettmers2024spqr,Dettmers2022LLMint88M,Kim2023SqueezeLLMDQ}; however, current hardware lacks efficient support for inference using this technique. 
Recent studies, such as QuIP~\cite{Chee2023QuIP} and QuaRot~\cite{ashkboos2024quarot}, have empirically shown that outliers in weights can be effectively mitigated by applying hadamard orthogonal transformations (hadamard rotations)~\cite{ashkboos2024slicegpt}, which reduce their impact and minimize the need for mixed-precision quantization. Furthermore, DuQuant~\cite{Lin2024DuQuantDO} and SpinQuant~\cite{Liu2024SpinQuant} optimize the rotation matrix through data instead of relying on pre-defined rotations.

Different from previous advancements in layer-wise quantization, \MethodName{} focuses on improving the learning objective of layer-wise quantization methods (detailed in \Cref{ssec:rsq}) by prioritizing important tokens. Our approach also builds on QuaRot by applying rotations to mitigate weight outliers and leveraging GPTQ for quantization. This integration makes \MethodName{} a comprehensive and holistic quantization strategy.

\section{Additional Evaluation Results and Analysis} \label{sec: additional evaluation}

\begin{table}[t]
    \centering
    \caption{Comparison of \MethodName{} and QuaRot on LongEval tasks. The model is quantized to 3-bit. The best-performing method among all quantization approaches is highlighted in bold.}
    \label{tab:longeval results}
    \resizebox{0.95\linewidth}{!}{
    \begin{tabular}{lcccm}
    \toprule
    \multirow{2}[2]{*}{Method} & \multicolumn{4}{c}{LongEval} \\
    \cmidrule{2-5}
     & $L$=300 & $L$=460 & $L$=620 & Avg \\
    \midrule
    Full Model & 100 & 98.80 & 82.00 & 93.60 \\
    \midrule
    \multicolumn{5}{c}{number of samples = 256, sequence length = 4096} \\
    \midrule
    QuaRot & 99.47$_{0.3}$ & 90.87$_{2.6}$ & 52.80$_{6.0}$ & 81.04$_{2.9}$ \\
    \MethodName{} & \textbf{99.60}$_{0.2}$ & \textbf{94.67}$_{3.9}$ & \textbf{58.00}$_{10.3}$ & \textbf{84.09}$_{4.8}$ \\
    \midrule
    \multicolumn{5}{c}{number of samples = 512, sequence length = 2048} \\
    \midrule
    QuaRot & 99.67$_{0.1}$ & 90.87$_{2.7}$ & 54.47$_{3.5}$ & 81.67$_{0.2}$ \\
    \MethodName{} & \textbf{99.80}$_{0.1}$ & \textbf{93.73}$_{1.7}$ & \textbf{55.33}$_{7.1}$ & \textbf{82.95}$_{2.7}$ \\
    \midrule
    \multicolumn{5}{c}{number of samples = 1024, sequence length = 1024} \\
    \midrule
    QuaRot & \textbf{99.80}$_{0.2}$ & 85.53$_{1.8}$ & 49.60$_{2.2}$ & 78.31$_{0.9}$ \\
    \MethodName{} & 99.67$_{0.3}$ & \textbf{88.33}$_{6.3}$ & \textbf{52.00}$_{12.9}$ & \textbf{80.00}$_{6.5}$ \\
    \bottomrule
    \end{tabular}
    }
\end{table}

\subsection{LongEval}
We evaluate the quantized LLaMA3 on the LongEval benchmark (details in \Cref{ssec: extended long-context benchmark details}) and present the results in \Cref{tab:longeval results}. Our findings indicate a clear performance improvement with \MethodName{} compared to QuaRot.
Moreover, in the LongEval evaluation, we observe a larger performance drop as input lengths increase, which aligns with the findings reported by \citet{Li2024EvaluatingQuantizedLLM}. 

\begin{figure}[t]
    \centering
    \includegraphics[width=\linewidth]{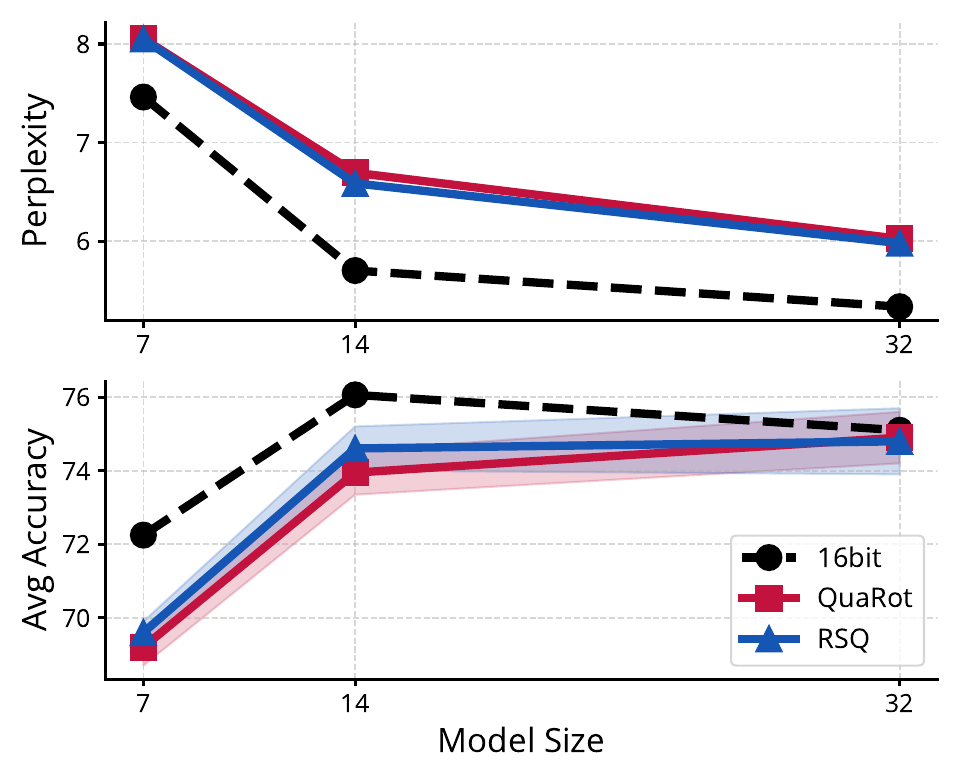}
    \caption{Ablation on model sizes using Qwen2.5.}
    \label{fig:model_scaling_qwen}
\end{figure}

\subsection{Scaling Model Sizes for Qwen2.5}
We also demonstrate \MethodName{}'s and QuaRot performance on different sizes of Qwen2.5 modes (7B, 14B, and 32B) in \Cref{fig:model_scaling_qwen}, showing that \MethodName{} still outperform the baseline for the three models.

\begin{figure}[t]
    \centering
    \includegraphics[width=\linewidth]{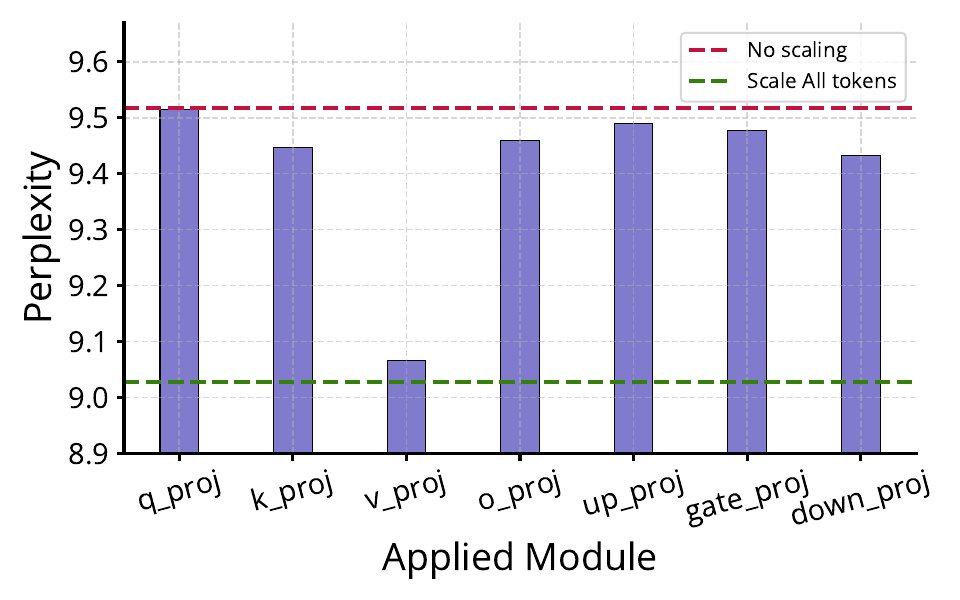}
    \caption{Ablation results on WikiText-2 about quantizing different modules with \MethodName{}.}
    \label{fig:different_modules_ppl}
\end{figure}

\subsection{Applying \MethodName{} on each module independently} \label{ssec: Applying rsq independently}

Generally, we apply \MethodName{} to all transformer modules simultaneously, including query, key, and value projection layers in attention layers and up, gate, and down projection layers in feed-forward networks. In this section, we conduct an ablation study where \MethodName{} is applied independently to each module, while the remaining modules use ``uniform'' token scaling (i.e., no scaling). The results, presented in \Cref{fig:different_modules_ppl}, indicate that while most modules benefit from \MethodName{}, the most significant improvement is observed in \texttt{v\_proj}. We hypothesize that this is because the values (outputs of \texttt{v\_proj}) have the most direct influence on all other tokens compared to other modules. However, we leave a deeper exploration of this phenomenon for future work.

\begin{figure}[t]
    \centering
    \includegraphics[width=\linewidth]{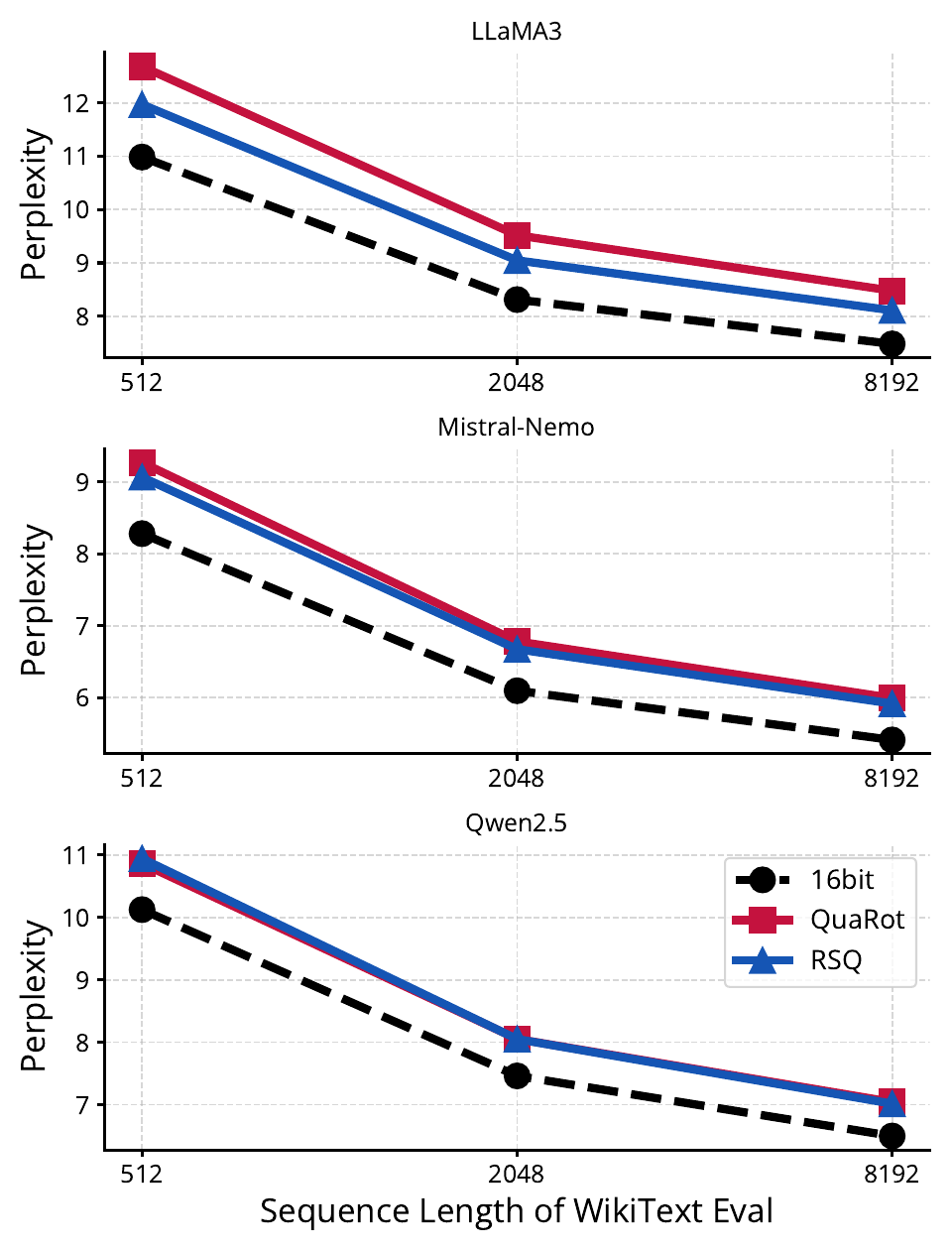}
    \caption{WikiText evaluation with three different context lengths for three models.}
    \label{fig:wiki_length}
\end{figure}

\subsection{Wiki Eval with Different Context Lengths} \label{ssec: wiki eval with different lengths}

Perplexity on WikiText is a widely used metric for evaluating quantization approaches. However, the context length of the evaluation set can significantly impact perplexity values, and previous studies have employed different setups. For instance, for LLaMA models, AQLM~\cite{Egiazarian2024aqlm} reports results with a context length of 4096, while QuaRot~\cite{ashkboos2024quarot} uses 2048.

While we use a context length of 2048 for most of the experiments in this paper, we also report WikiText perplexity at additional context lengths (512 and 8192) in \Cref{fig:wiki_length} for three models. We observe that the performance gap between different approaches remains relatively consistent across context lengths, suggesting that either of the lengths can serve as a reliable metric for comparing methods. However, as expected, longer contexts generally lead to lower perplexity, likely because having more preceding tokens for LLMs to attend to improves prediction accuracy. Therefore, we emphasize that when using perplexity as a comparison metric, it is crucial to maintain consistent settings to ensure fair and meaningful comparisons.

\subsection{The Effect of Scaling without Rotation} \label{ssec: scaling without rotation}

\begin{figure}[t]
    \centering
    \includegraphics[width=\linewidth]{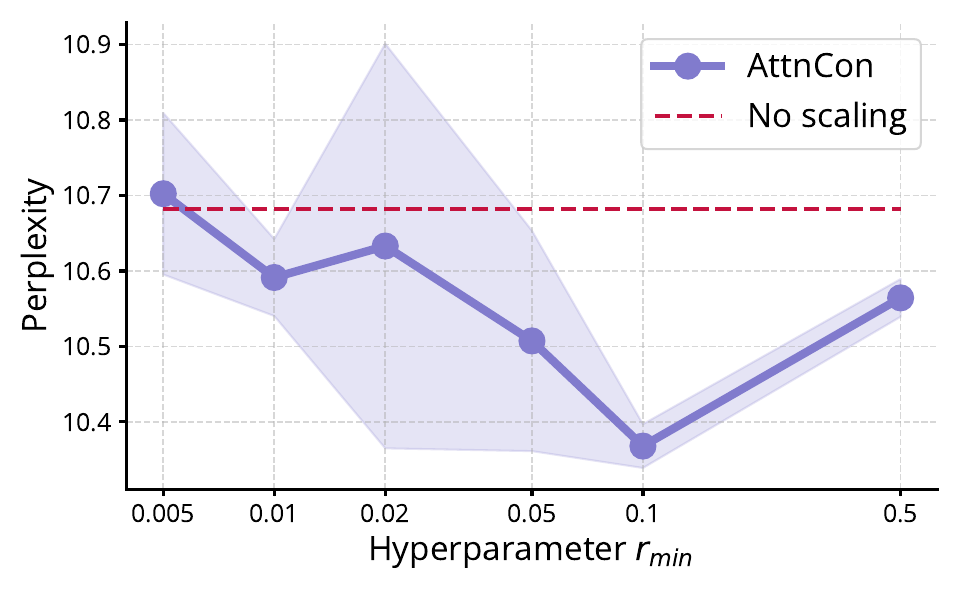}
    \caption{Ablation study of AttnCon on SQ (without rotation).}
    \label{fig:attncon_gptq}
\end{figure}

In previous experiments, we applied the scaling strategy to weights after rotation. In this section, we investigate its effect when the weights remain unrotated. We refer to this approach as SQ (Scale, then Quantize) and present the results of applying AttnCon on SQ in \Cref{fig:attncon_gptq}. Our findings indicate that the best perplexity is achieved at $r_{min} = 0.1$, which is significantly larger than the optimal value observed when weights are rotated ($r_{min} = 0.005$). This suggests that scaling is far more effective when applied to rotated weights. We leave further investigation of this phenomenon for future work.

\section{Visualization}

We visualize the token importance scores assigned by adaptive approaches for three samples and three layers (the 3rd, 11th, and 21st layers), as shown in \Cref{fig:tokenfreq_visualization,fig:actnorm_visualization,fig:actdiff_visualization,fig:tokensim_visualization,fig:attncon_visualization}. Note that we clamp the values into the range [$0.05$\text{-th quantile}, $99.95$\text{-th quantile}] for better visualization.

For TokenFreq (\Cref{fig:tokenfreq_visualization}), the scores are close to one for most tokens, suggesting that many tokens in a sequence appear very less frequently. ActDiff (\Cref{fig:actdiff_visualization}) does not exhibit any clear patterns. In contrast, ActNorm (\Cref{fig:actnorm_visualization}) reveals that the first token tends to have a slightly larger norm, particularly in the 3rd and 11th layers. For TokenSim (\Cref{fig:tokensim_visualization}), we observe that the first token is significantly less similar to others in earlier layers but becomes more similar in deeper layers. Lastly, AttnCon (\Cref{fig:attncon_visualization}) consistently assigns higher scores to the initial and final tokens across all layers.

\input{tables_figures/tokenfreq_visualization}
\input{tables_figures/actnorm_visualization}
\input{tables_figures/actdiff_visualization}
\input{tables_figures/tokensim_visualization}
\input{tables_figures/attncon_visualization}

%% file: tables_figures/tokenfreq_visualization.tex
\begin{figure*}[t]
    \centering
    \begin{subfigure}[b]{0.33\linewidth}
    \centering
        \includegraphics[width=\linewidth]{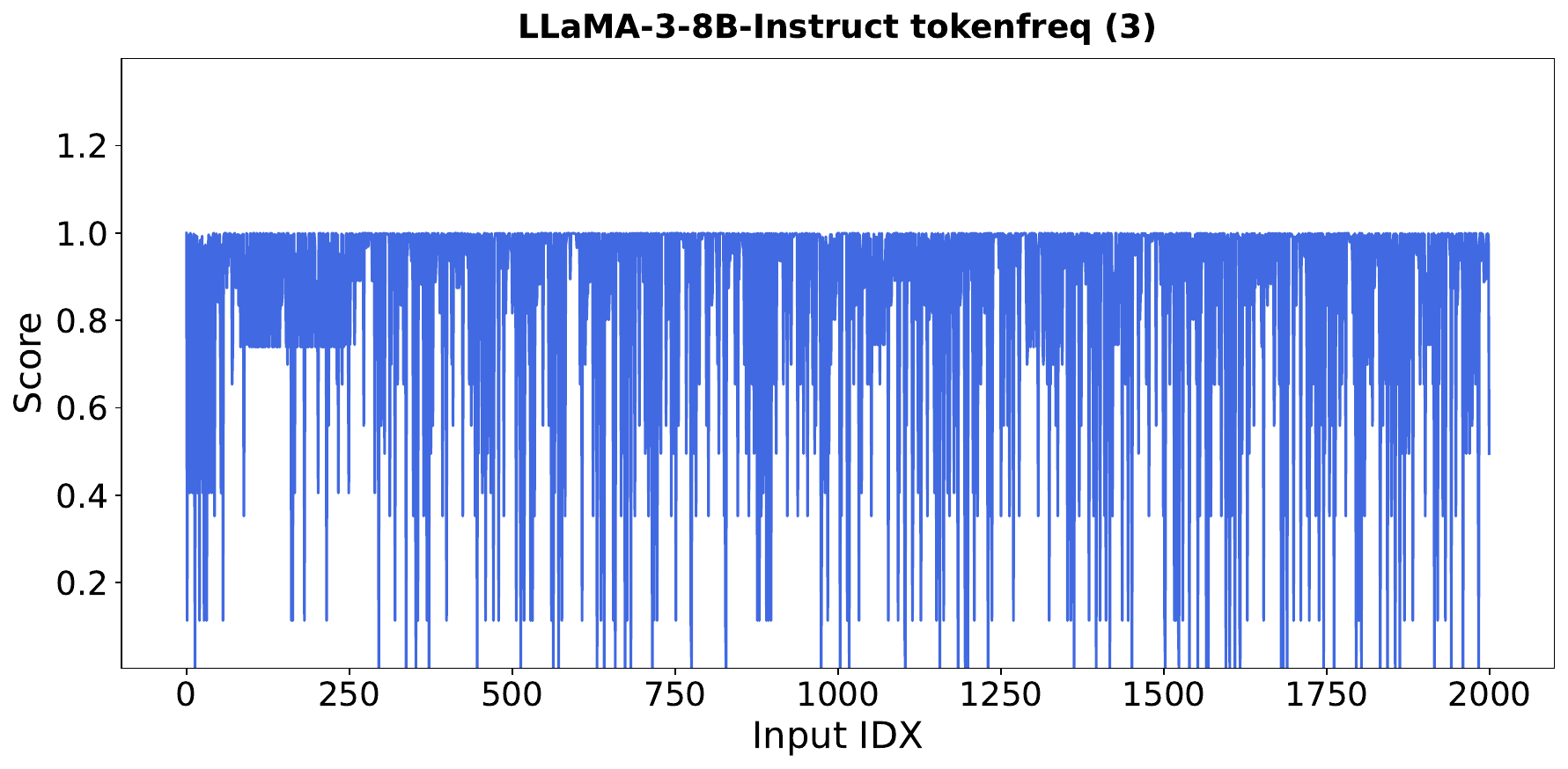}
        \captionsetup{width=0.4\linewidth}
        \caption{Layer 3}
    \end{subfigure}
    \begin{subfigure}[b]{0.33\linewidth}
        \centering
        \includegraphics[width=\linewidth]{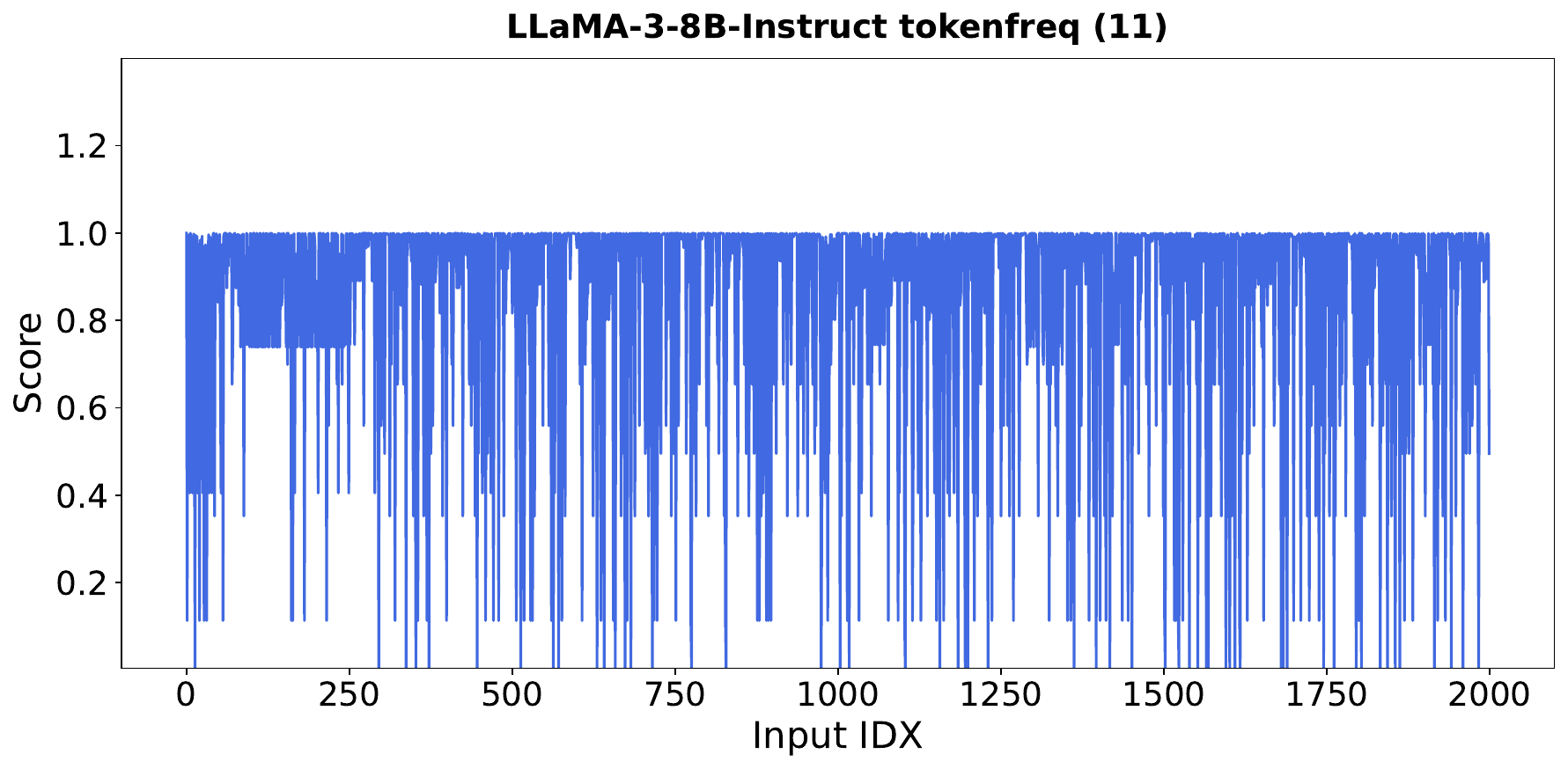}
        \captionsetup{width=0.4\linewidth}
        \caption{Layer 11}
     \end{subfigure}
     \begin{subfigure}[b]{0.33\linewidth}
        \centering
        \includegraphics[width=\linewidth]{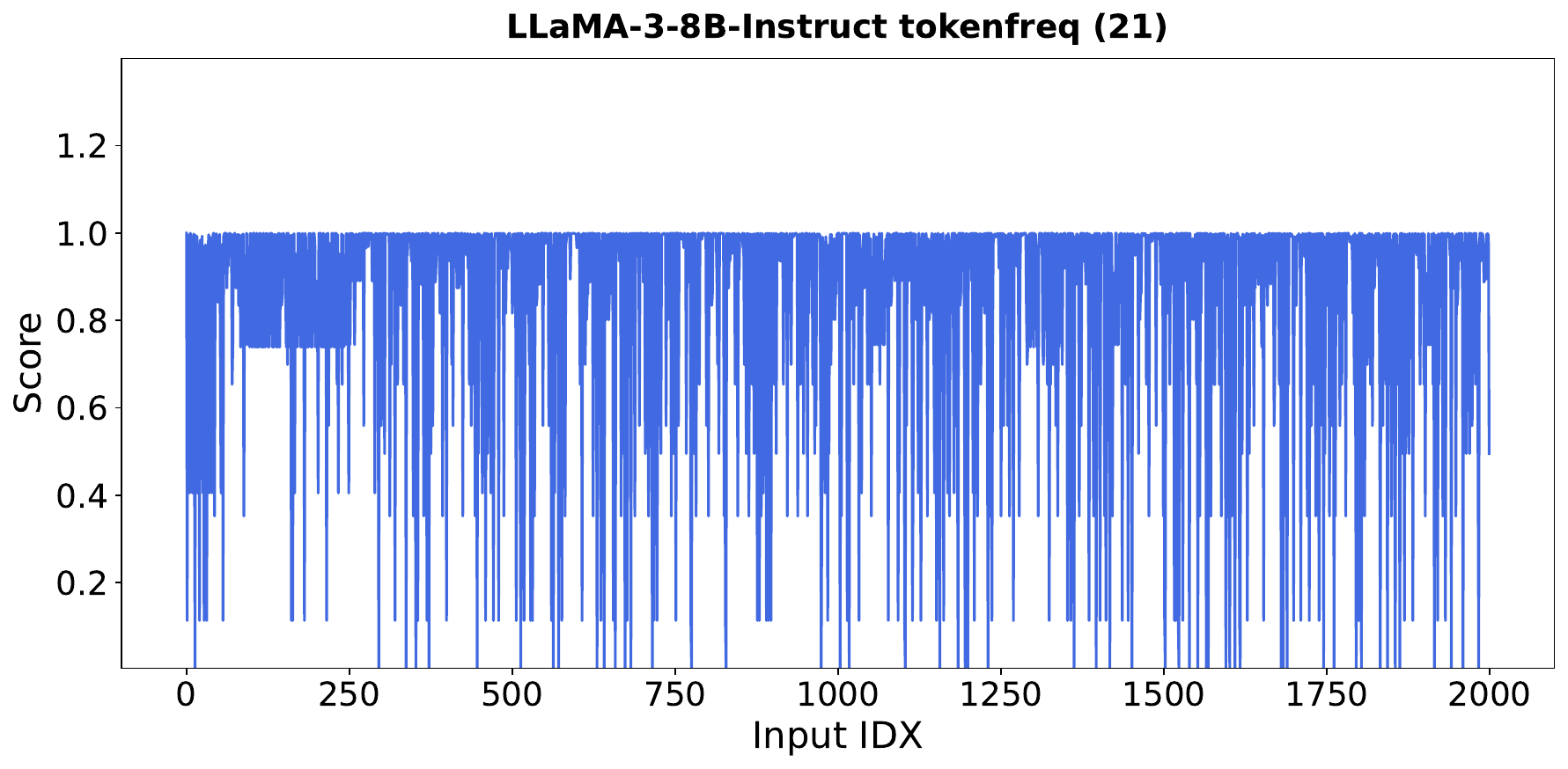}
        \captionsetup{width=0.4\linewidth}
        \caption{Layer 21}
     \end{subfigure}
    
    \vspace{0.3cm} 

    \begin{minipage}{\textwidth}
        \centering
        \textbf{Figures (a) - (c):} first example. 
    \end{minipage}
    \vspace{0.3cm}

    \begin{subfigure}[b]{0.33\linewidth}
    \centering
        \includegraphics[width=\linewidth]{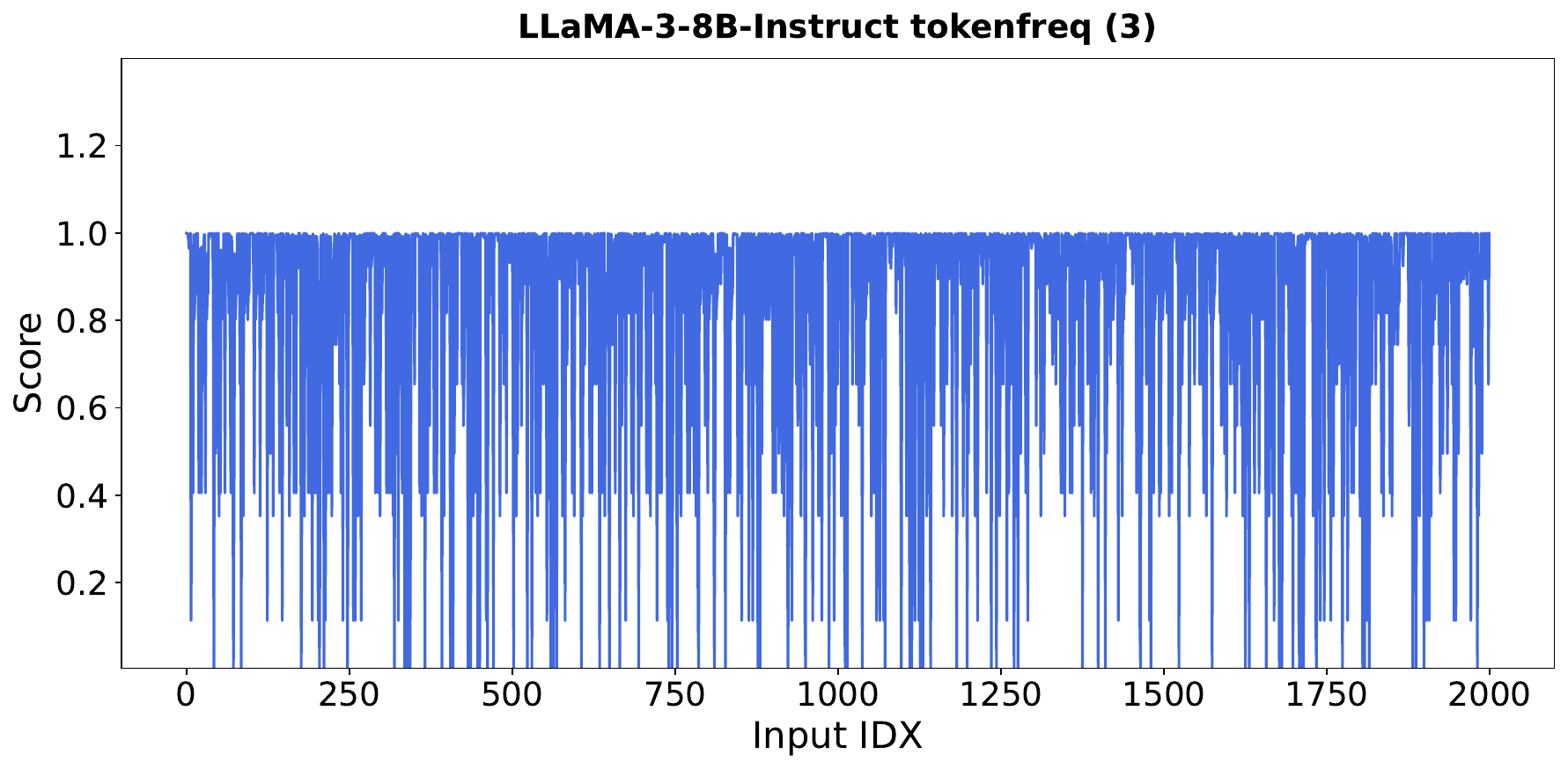}
        \captionsetup{width=0.4\linewidth}
        \caption{Layer 3}
    \end{subfigure}
    \begin{subfigure}[b]{0.33\linewidth}
        \centering
        \includegraphics[width=\linewidth]{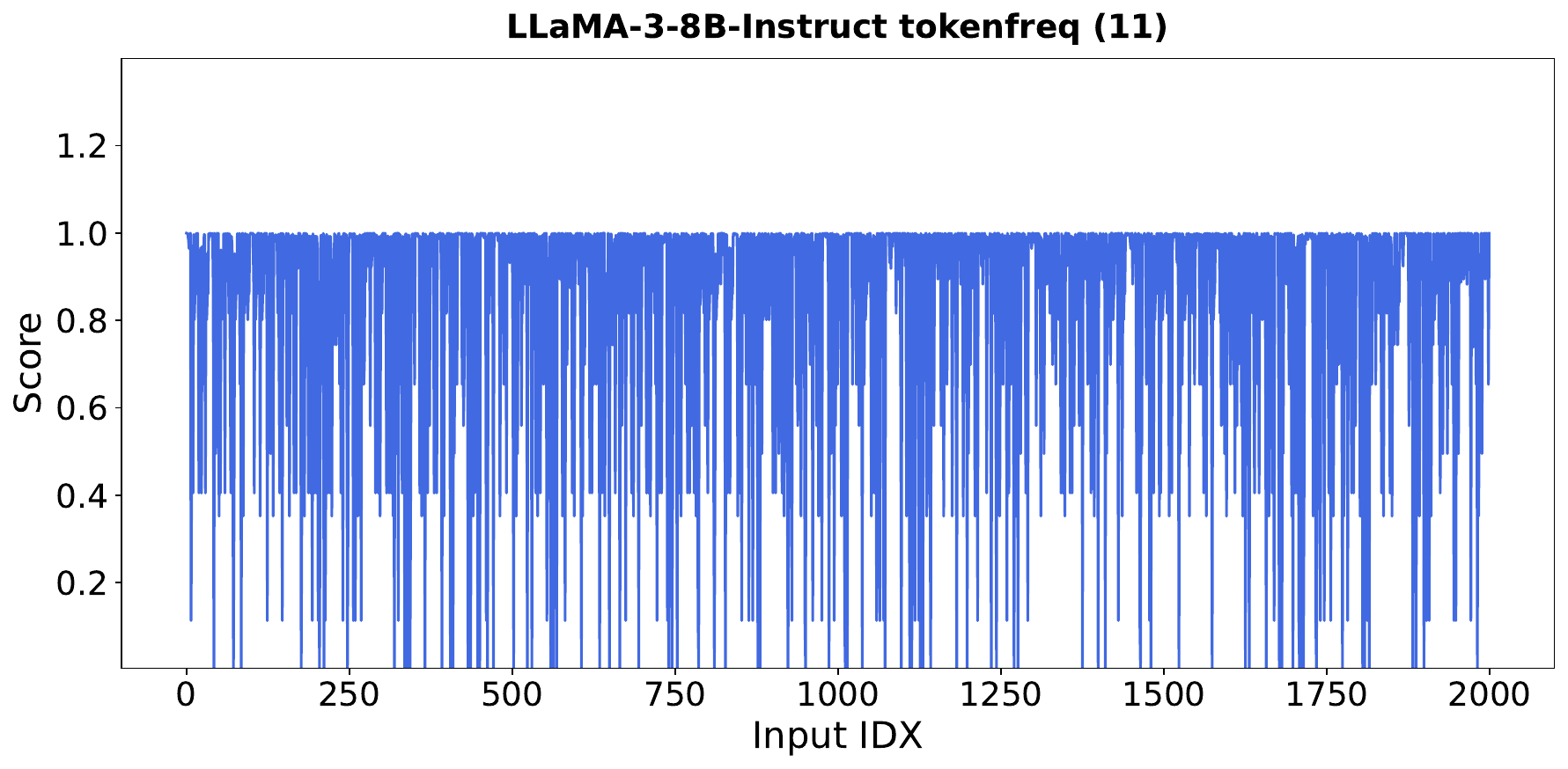}
        \captionsetup{width=0.4\linewidth}
        \caption{Layer 11}
     \end{subfigure}
     \begin{subfigure}[b]{0.33\linewidth}
        \centering
        \includegraphics[width=\linewidth]{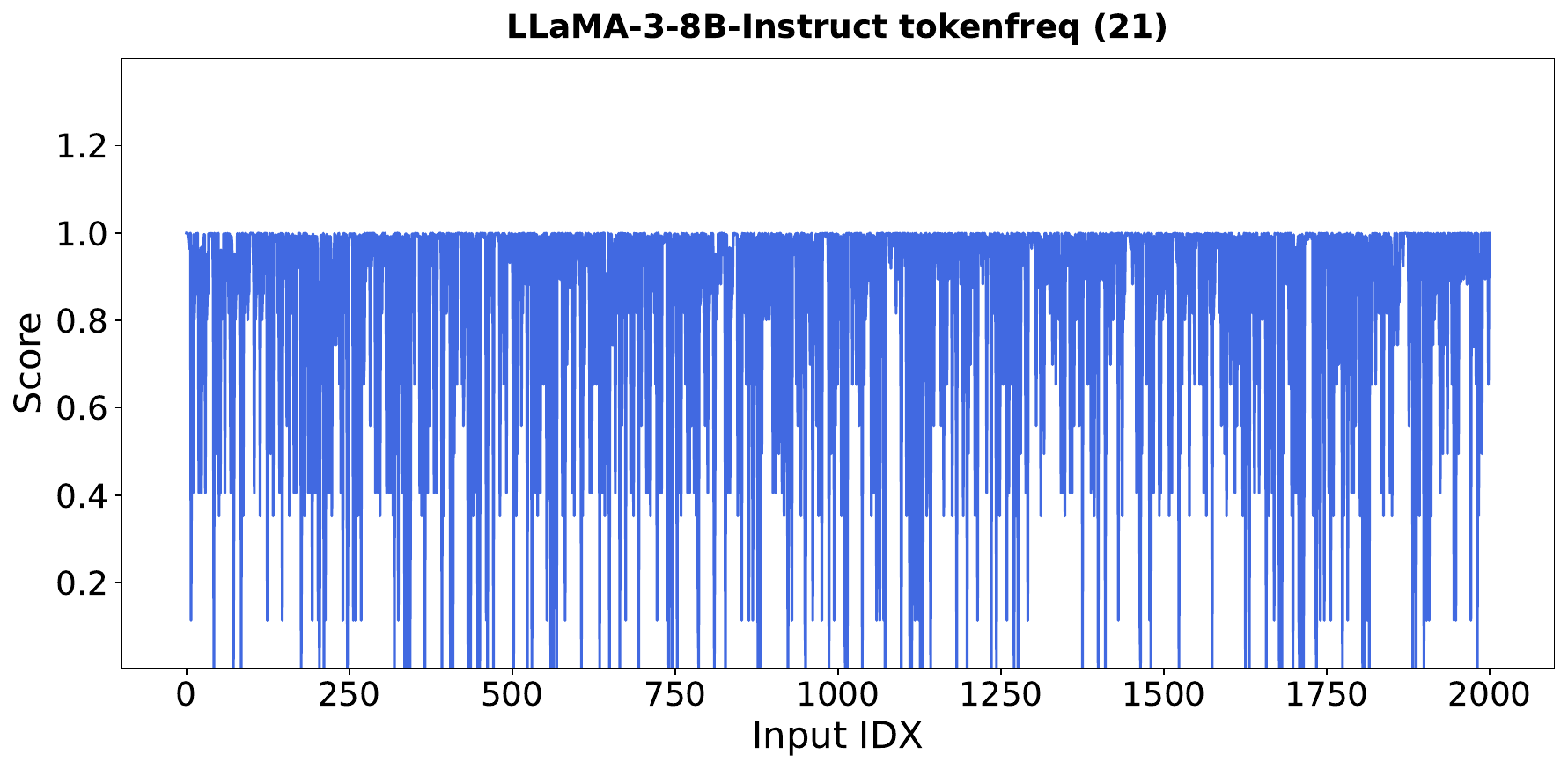}
        \captionsetup{width=0.4\linewidth}
        \caption{Layer 21}
     \end{subfigure}
    
    \vspace{0.3cm} 

     \begin{minipage}{\textwidth}
        \centering
        \textbf{Figures (d) - (f):} second example. 
    \end{minipage}
    \vspace{0.3cm}
    
    \begin{subfigure}[b]{0.33\linewidth}
    \centering
        \includegraphics[width=\linewidth]{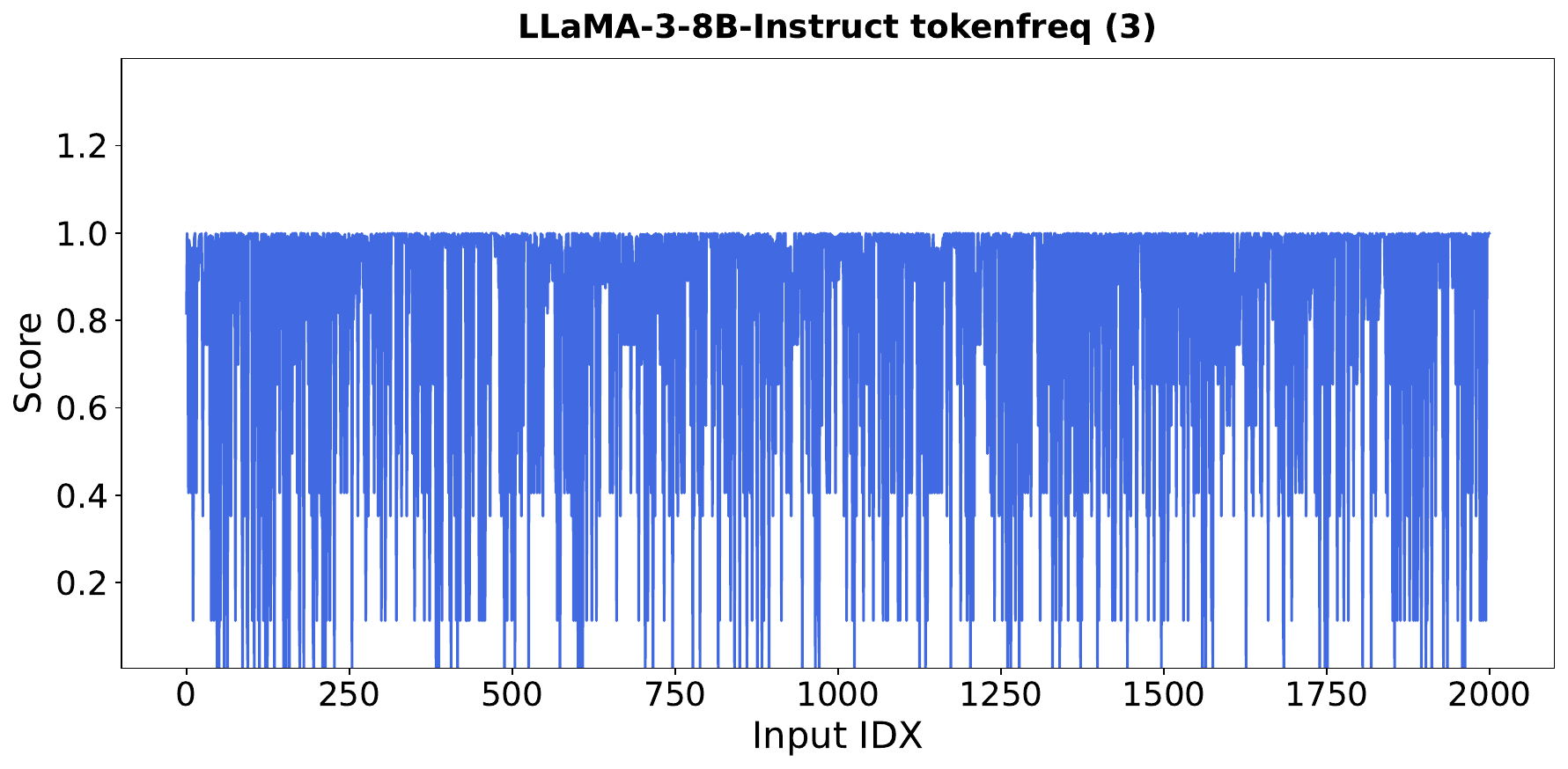}
        \captionsetup{width=0.4\linewidth}
        \caption{Layer 3}
    \end{subfigure}
    \begin{subfigure}[b]{0.33\linewidth}
        \centering
        \includegraphics[width=\linewidth]{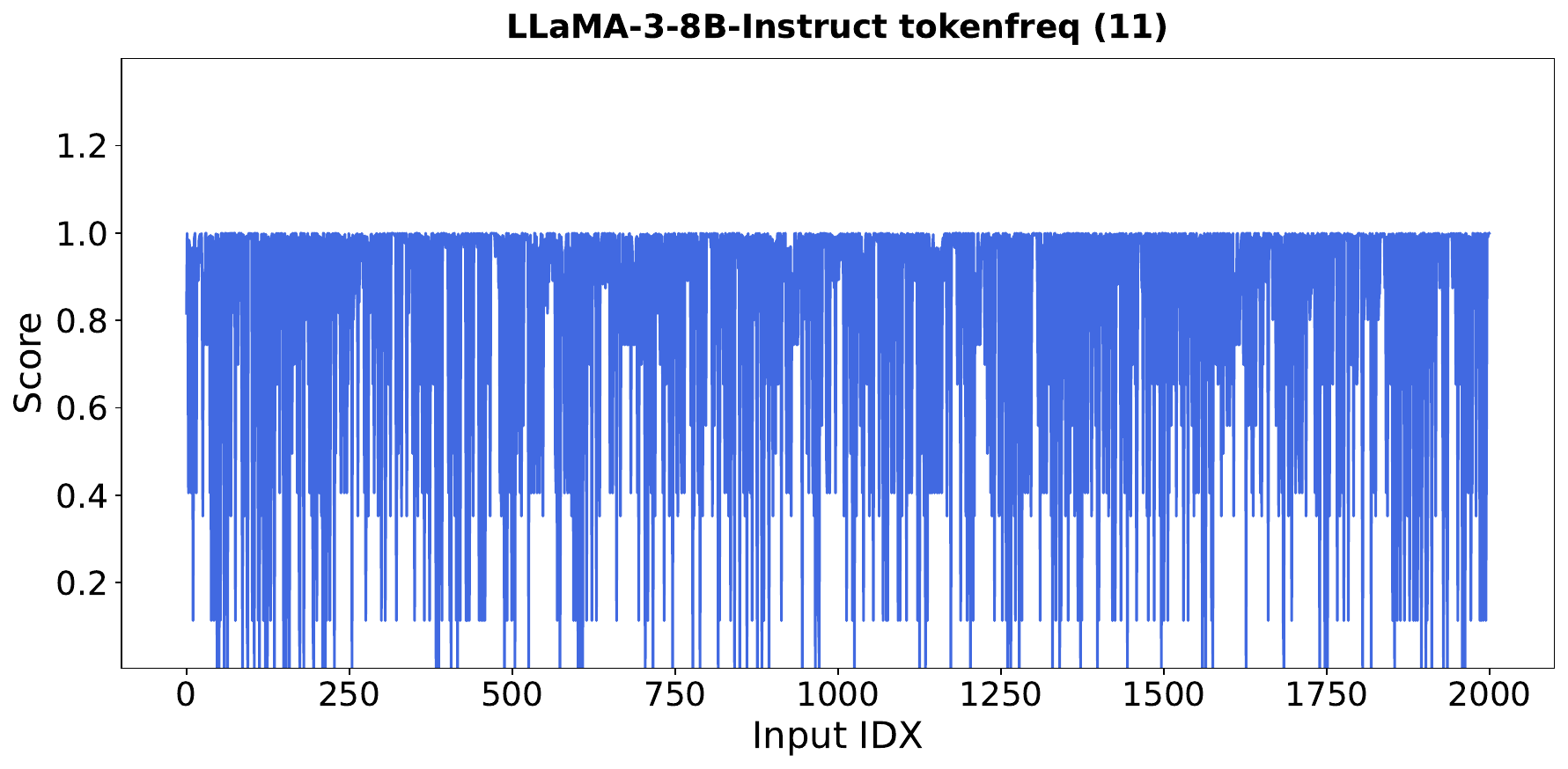}
        \captionsetup{width=0.4\linewidth}
        \caption{Layer 11}
     \end{subfigure}
     \begin{subfigure}[b]{0.33\linewidth}
        \centering
        \includegraphics[width=\linewidth]{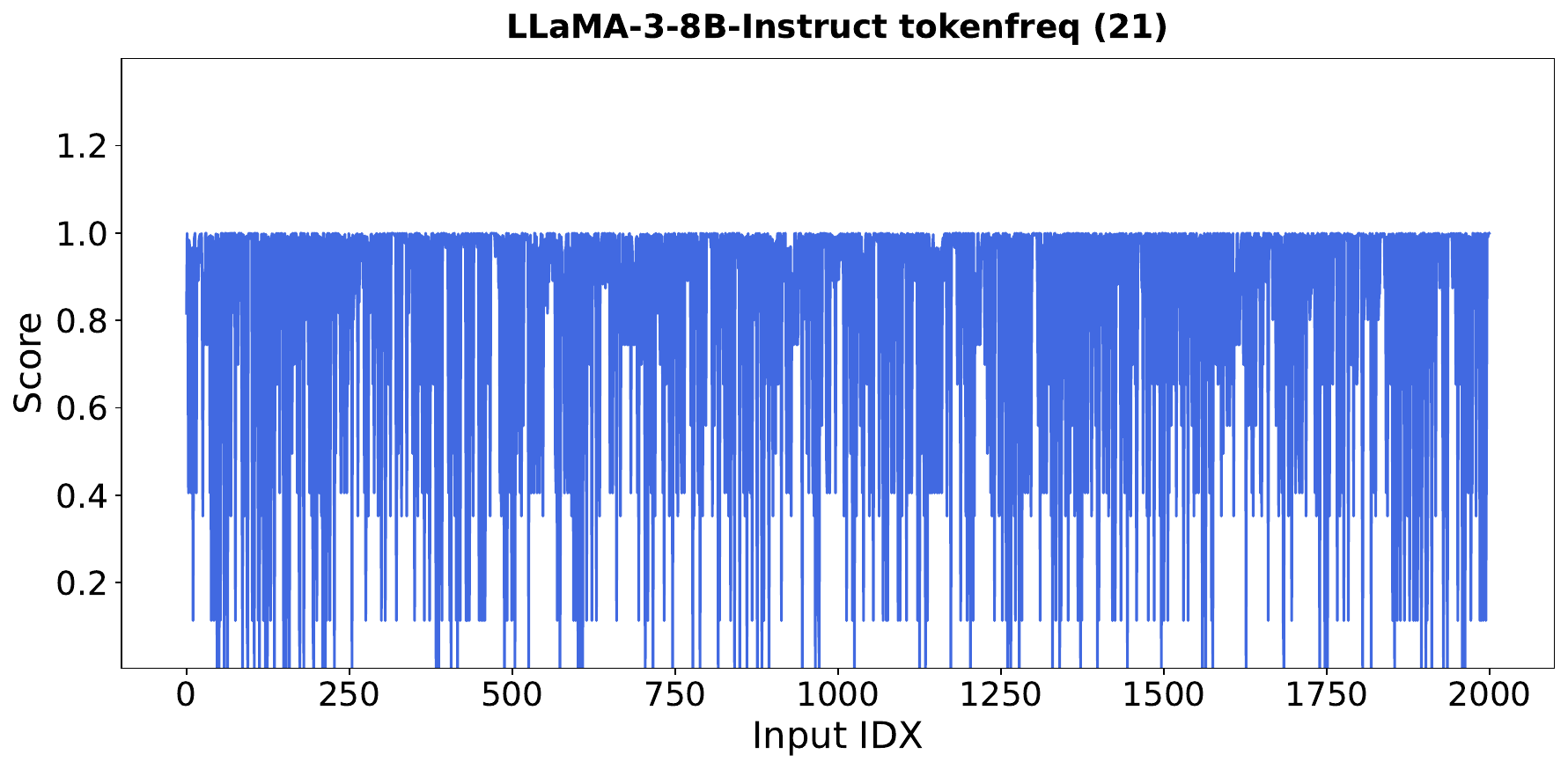}
        \captionsetup{width=0.4\linewidth}
        \caption{Layer 21}
     \end{subfigure}
    
    \vspace{0.3cm} 

     \begin{minipage}{\textwidth}
        \centering
        \textbf{Figures (g) - (i):} third example. 
    \end{minipage}
    \vspace{0.3cm}
    \caption{Visualization of TokenFreq scores across three layers for three different examples.}
    \label{fig:tokenfreq_visualization}
\end{figure*}

%% file: tables_figures/actnorm_visualization.tex
\begin{figure*}[t]
    \centering
    \begin{subfigure}[b]{0.33\linewidth}
    \centering
        \includegraphics[width=\linewidth]{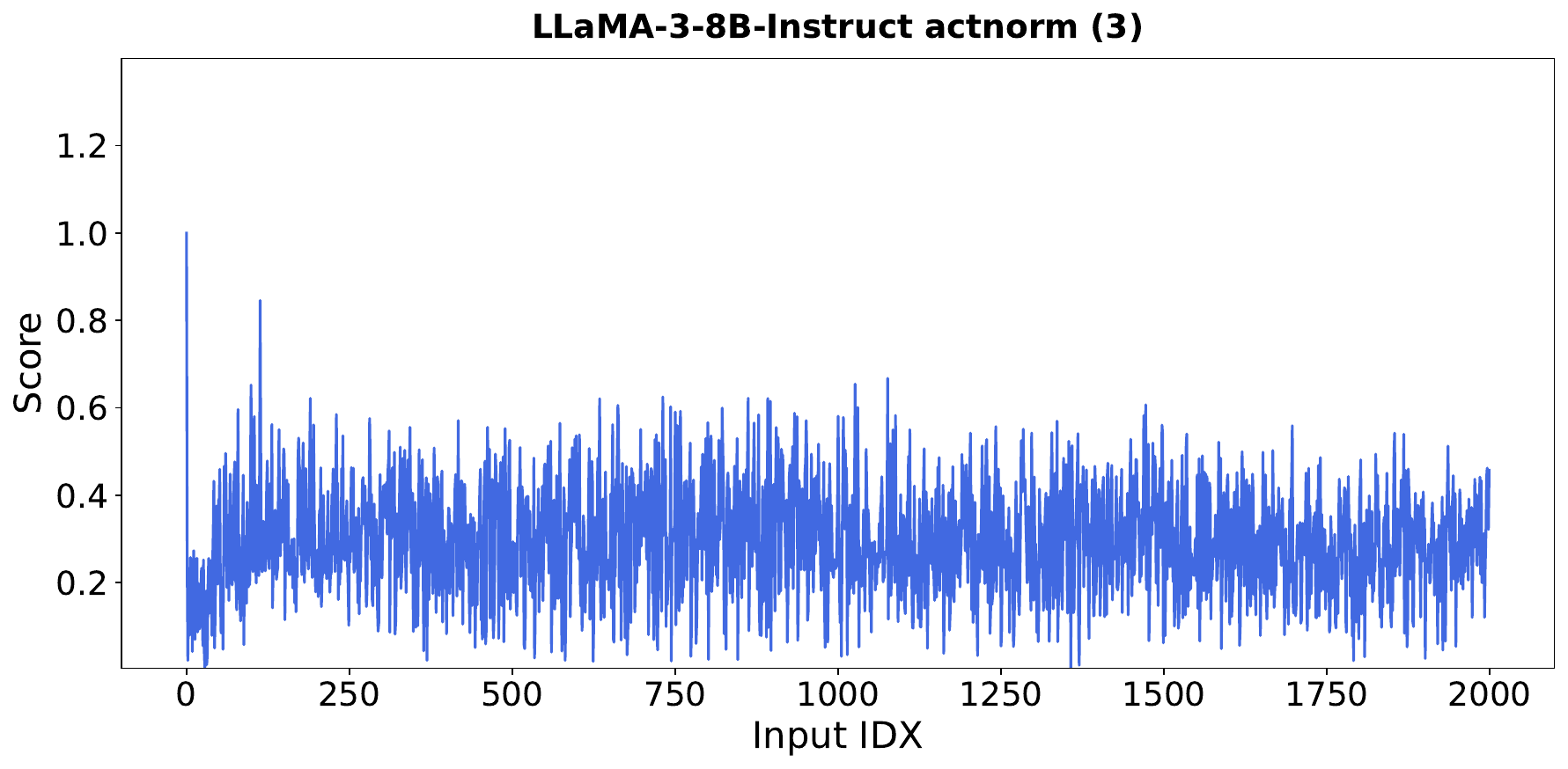}
        \captionsetup{width=0.4\linewidth}
        \caption{Layer 3}
    \end{subfigure}
    \begin{subfigure}[b]{0.33\linewidth}
        \centering
        \includegraphics[width=\linewidth]{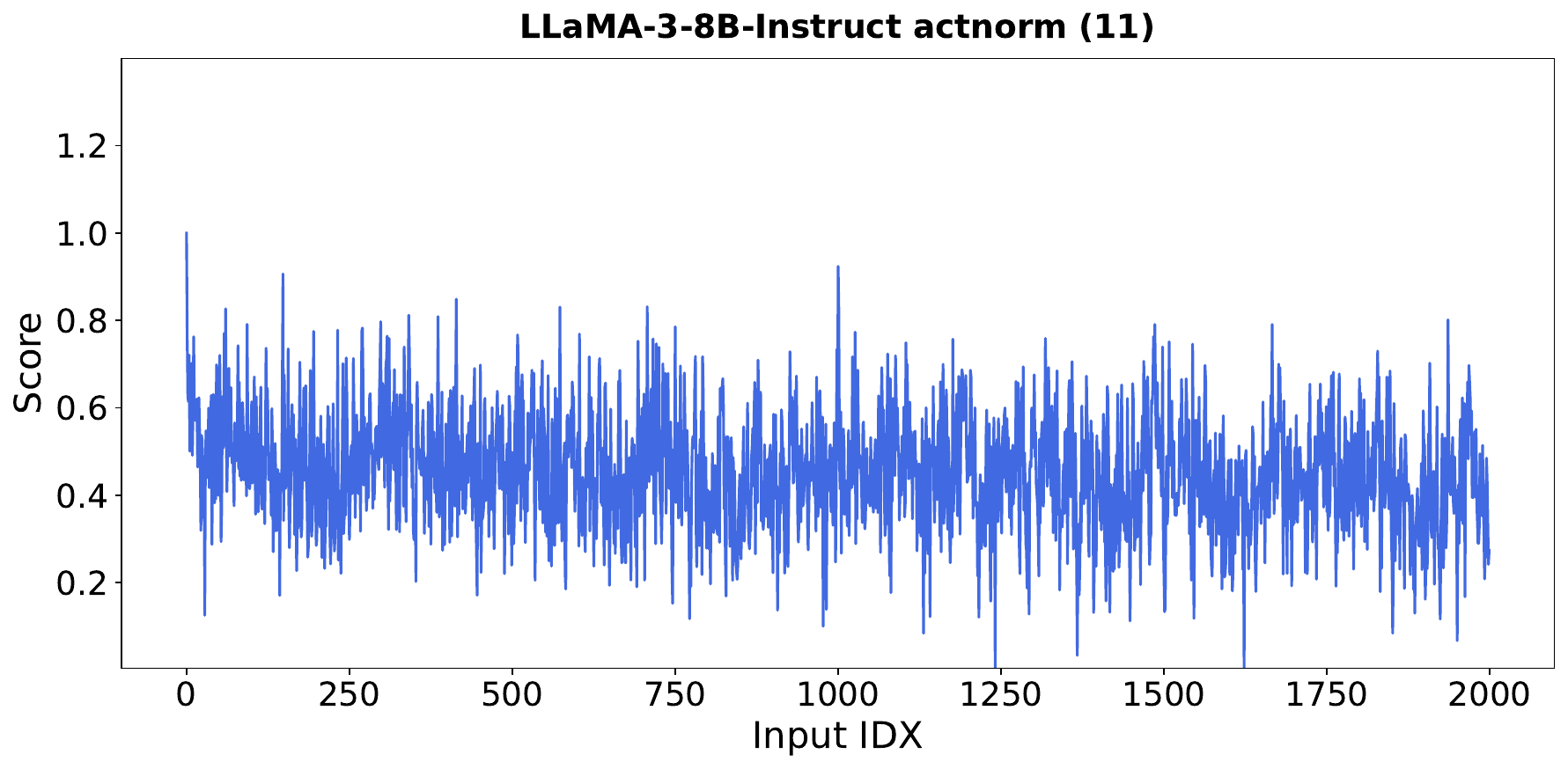}
        \captionsetup{width=0.4\linewidth}
        \caption{Layer 11}
     \end{subfigure}
     \begin{subfigure}[b]{0.33\linewidth}
        \centering
        \includegraphics[width=\linewidth]{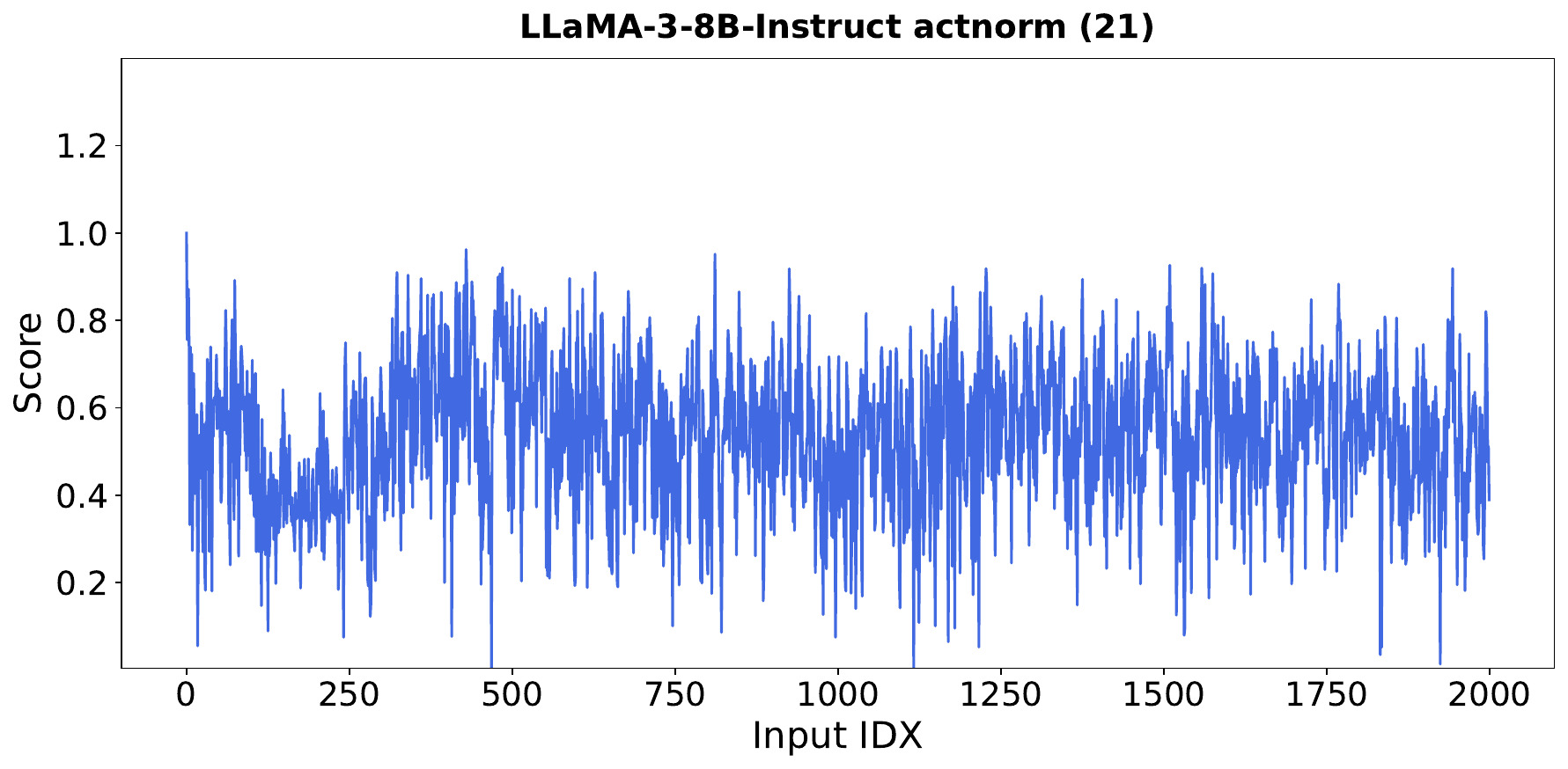}
        \captionsetup{width=0.4\linewidth}
        \caption{Layer 21}
     \end{subfigure}
    
    \vspace{0.3cm} 

    \begin{minipage}{\textwidth}
        \centering
        \textbf{Figures (a) - (c):} first example. 
    \end{minipage}
    \vspace{0.3cm}

    \begin{subfigure}[b]{0.33\linewidth}
    \centering
        \includegraphics[width=\linewidth]{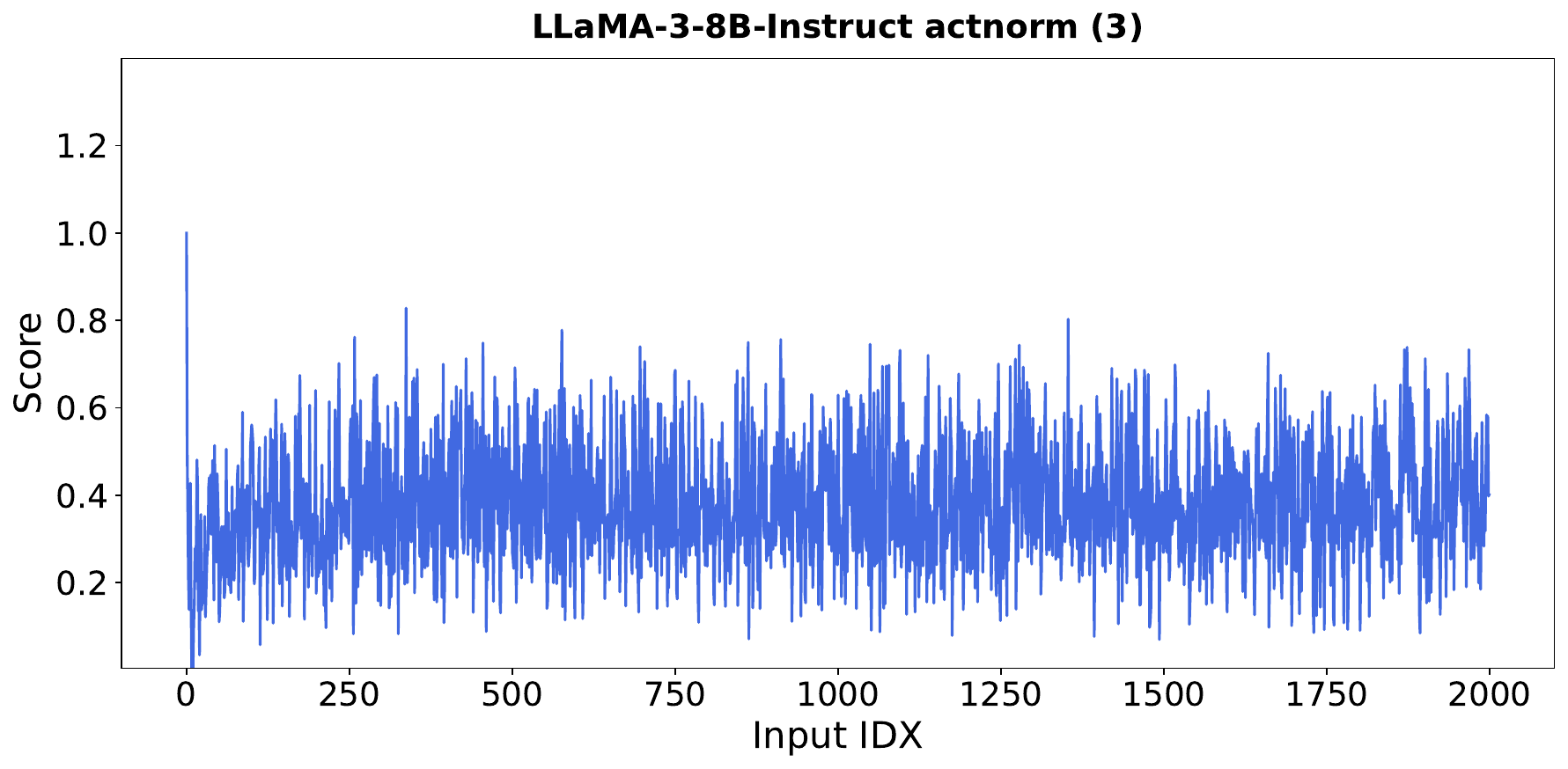}
        \captionsetup{width=0.4\linewidth}
        \caption{Layer 3}
    \end{subfigure}
    \begin{subfigure}[b]{0.33\linewidth}
        \centering
        \includegraphics[width=\linewidth]{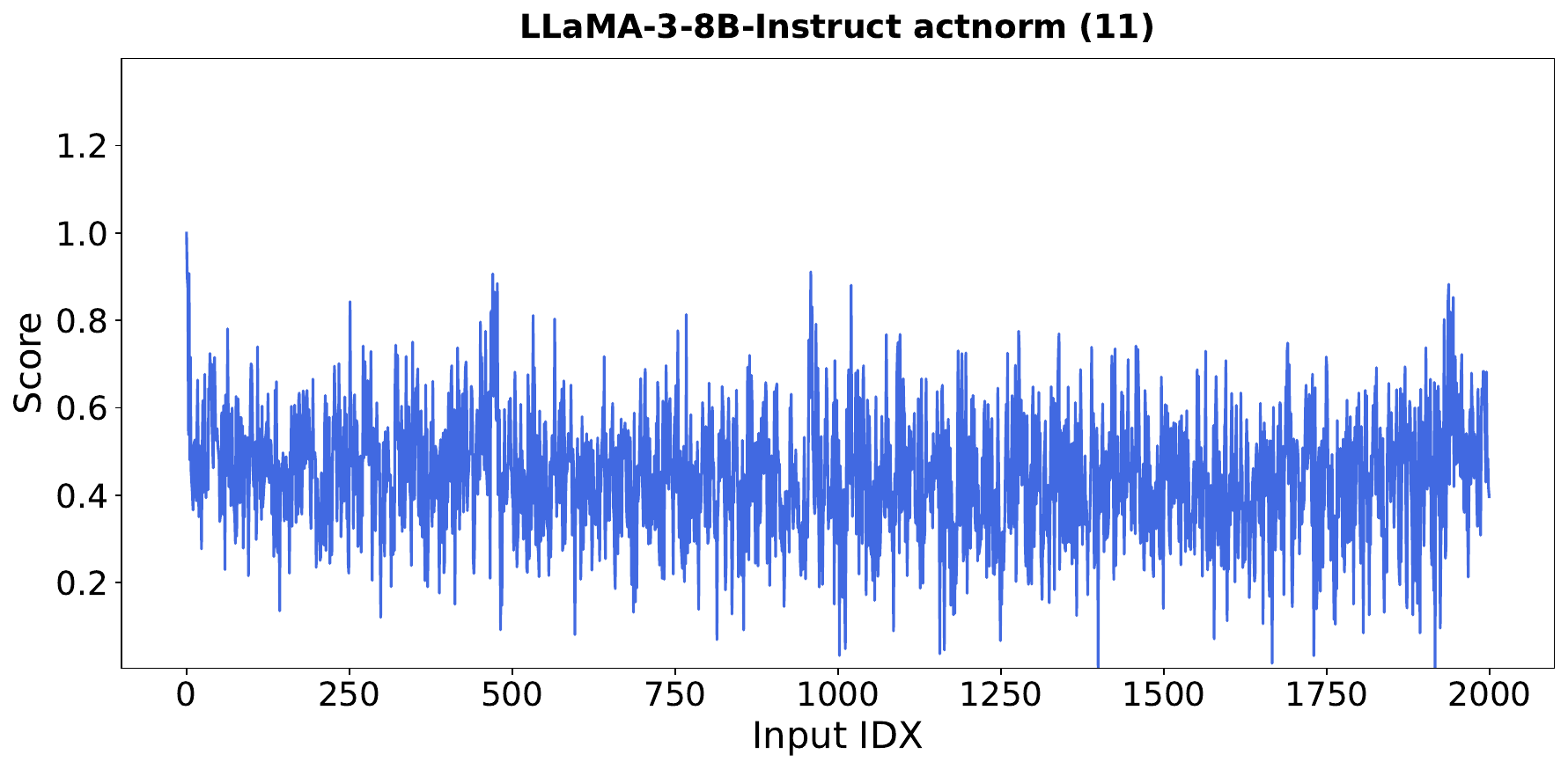}
        \captionsetup{width=0.4\linewidth}
        \caption{Layer 11}
     \end{subfigure}
     \begin{subfigure}[b]{0.33\linewidth}
        \centering
        \includegraphics[width=\linewidth]{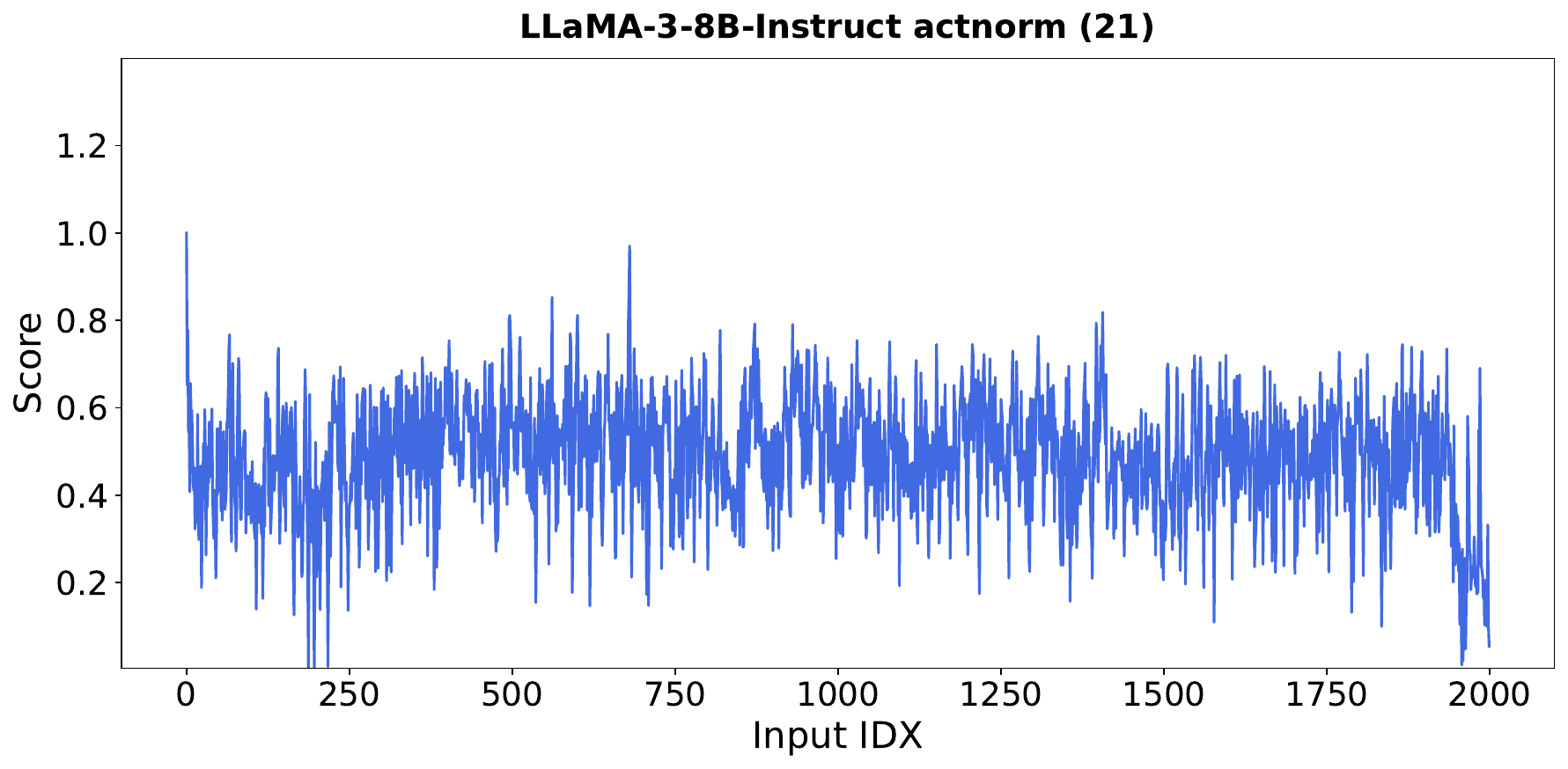}
        \captionsetup{width=0.4\linewidth}
        \caption{Layer 21}
     \end{subfigure}
    
    \vspace{0.3cm} 

     \begin{minipage}{\textwidth}
        \centering
        \textbf{Figures (d) - (f):} second example. 
    \end{minipage}
    \vspace{0.3cm}
    
    \begin{subfigure}[b]{0.33\linewidth}
    \centering
        \includegraphics[width=\linewidth]{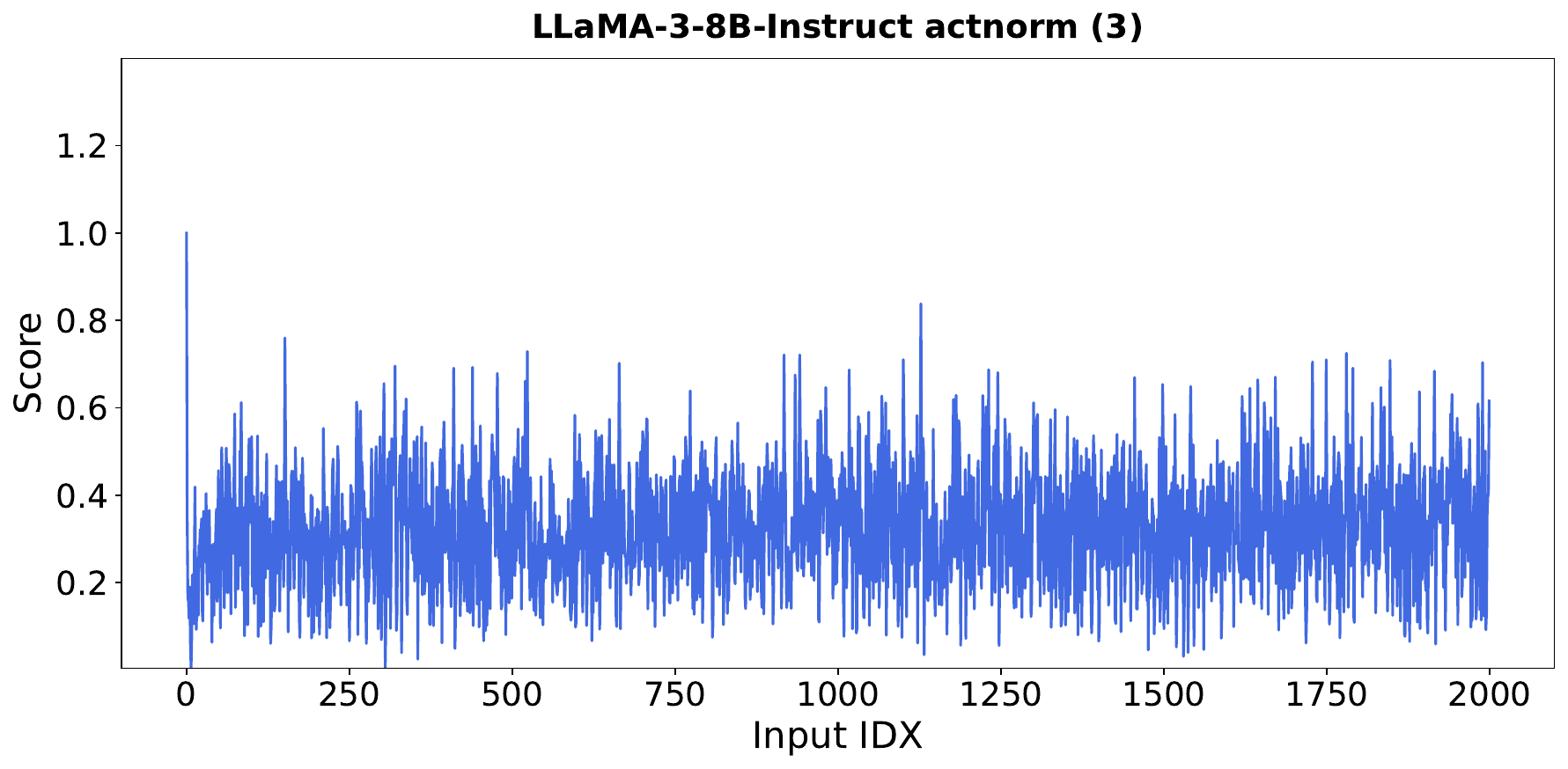}
        \captionsetup{width=0.4\linewidth}
        \caption{Layer 3}
    \end{subfigure}
    \begin{subfigure}[b]{0.33\linewidth}
        \centering
        \includegraphics[width=\linewidth]{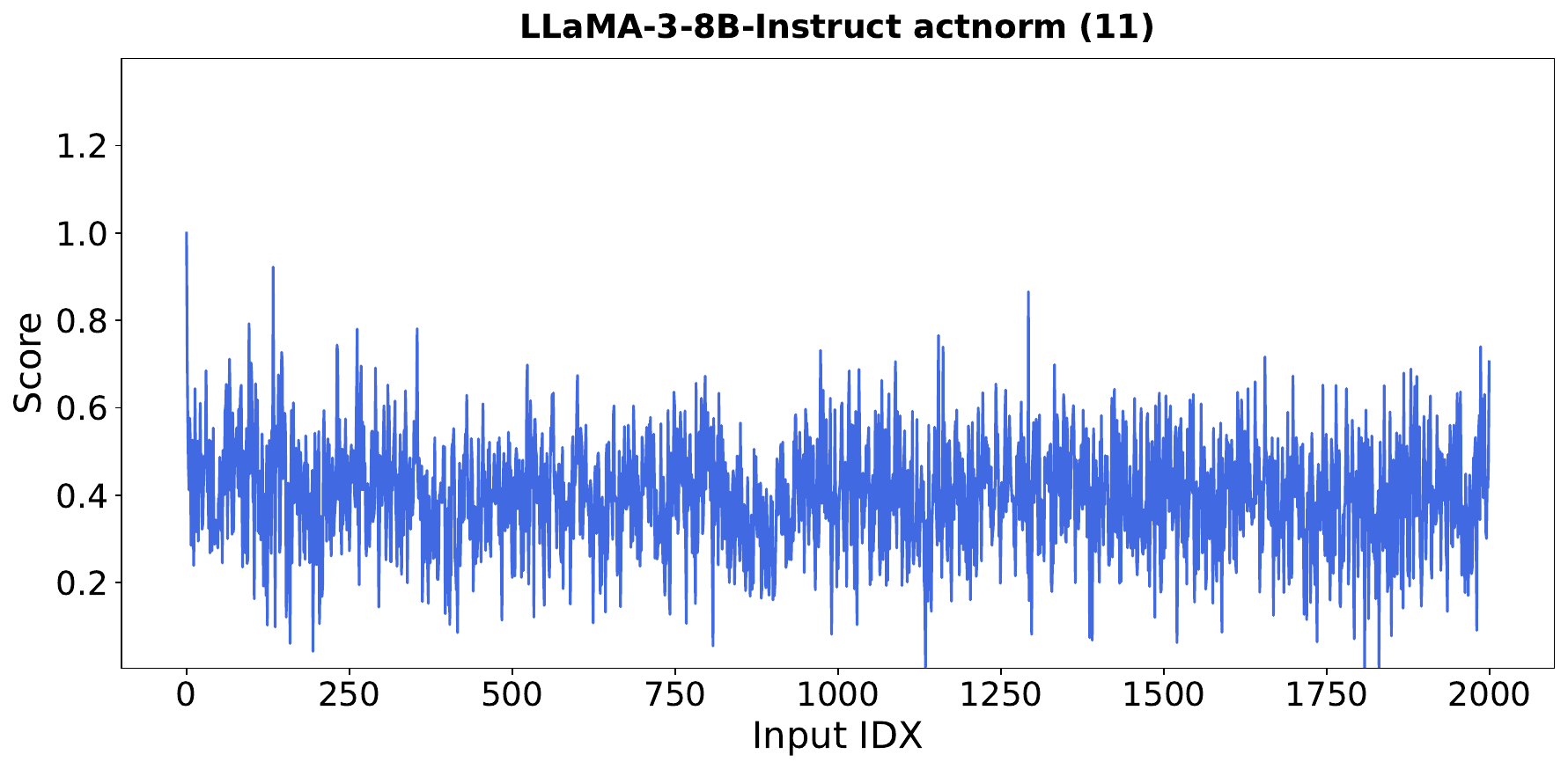}
        \captionsetup{width=0.4\linewidth}
        \caption{Layer 11}
     \end{subfigure}
     \begin{subfigure}[b]{0.33\linewidth}
        \centering
        \includegraphics[width=\linewidth]{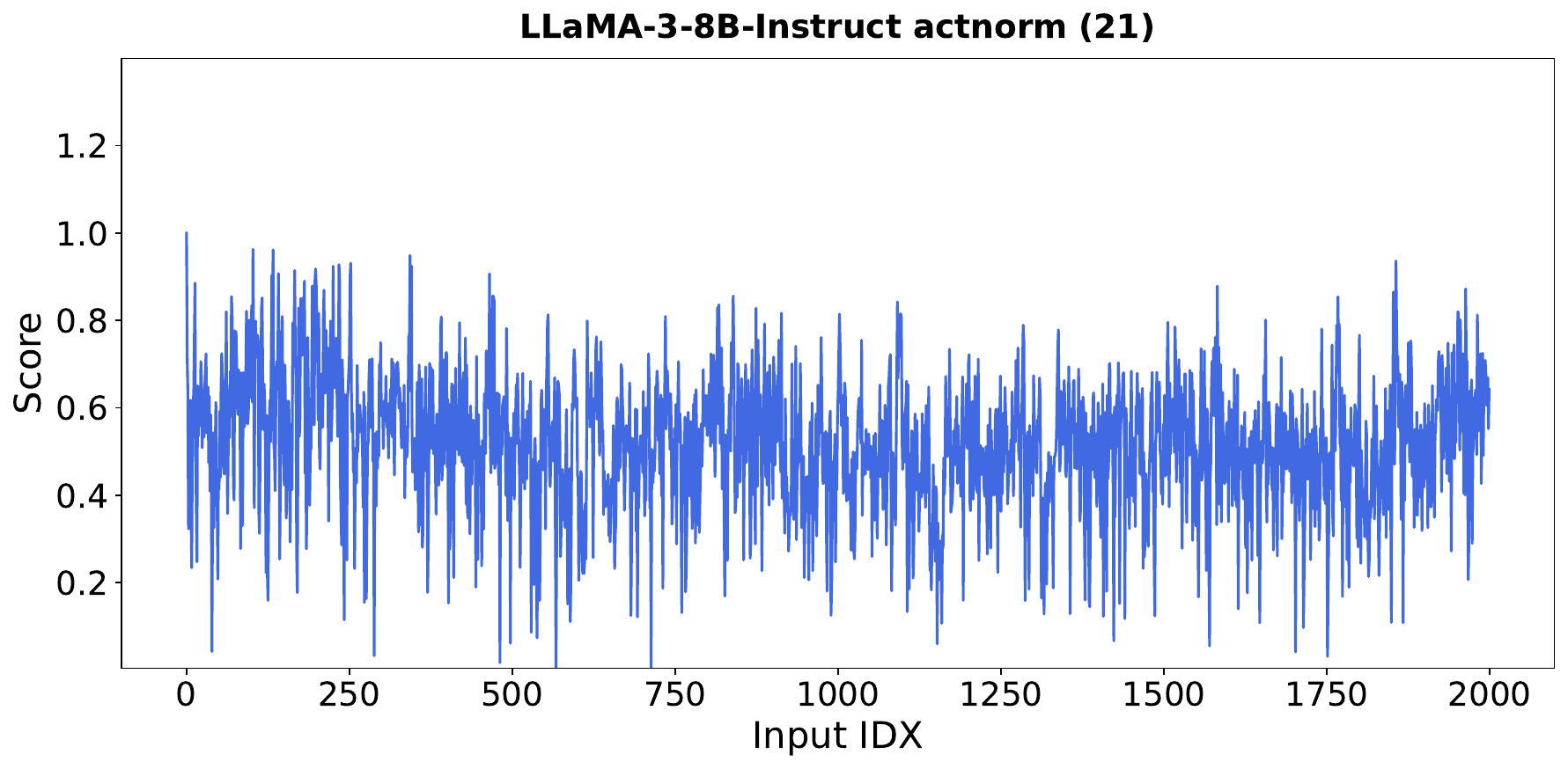}
        \captionsetup{width=0.4\linewidth}
        \caption{Layer 21}
     \end{subfigure}
    
    \vspace{0.3cm} 

     \begin{minipage}{\textwidth}
        \centering
        \textbf{Figures (g) - (i):} third example. 
    \end{minipage}
    \vspace{0.3cm}
    \caption{Visualization of ActNorm scores across three layers for three different examples.}
    \label{fig:actnorm_visualization}
\end{figure*}

%% file: tables_figures/actdiff_visualization.tex
\begin{figure*}[t]
    \centering
    \begin{subfigure}[b]{0.33\linewidth}
    \centering
        \includegraphics[width=\linewidth]{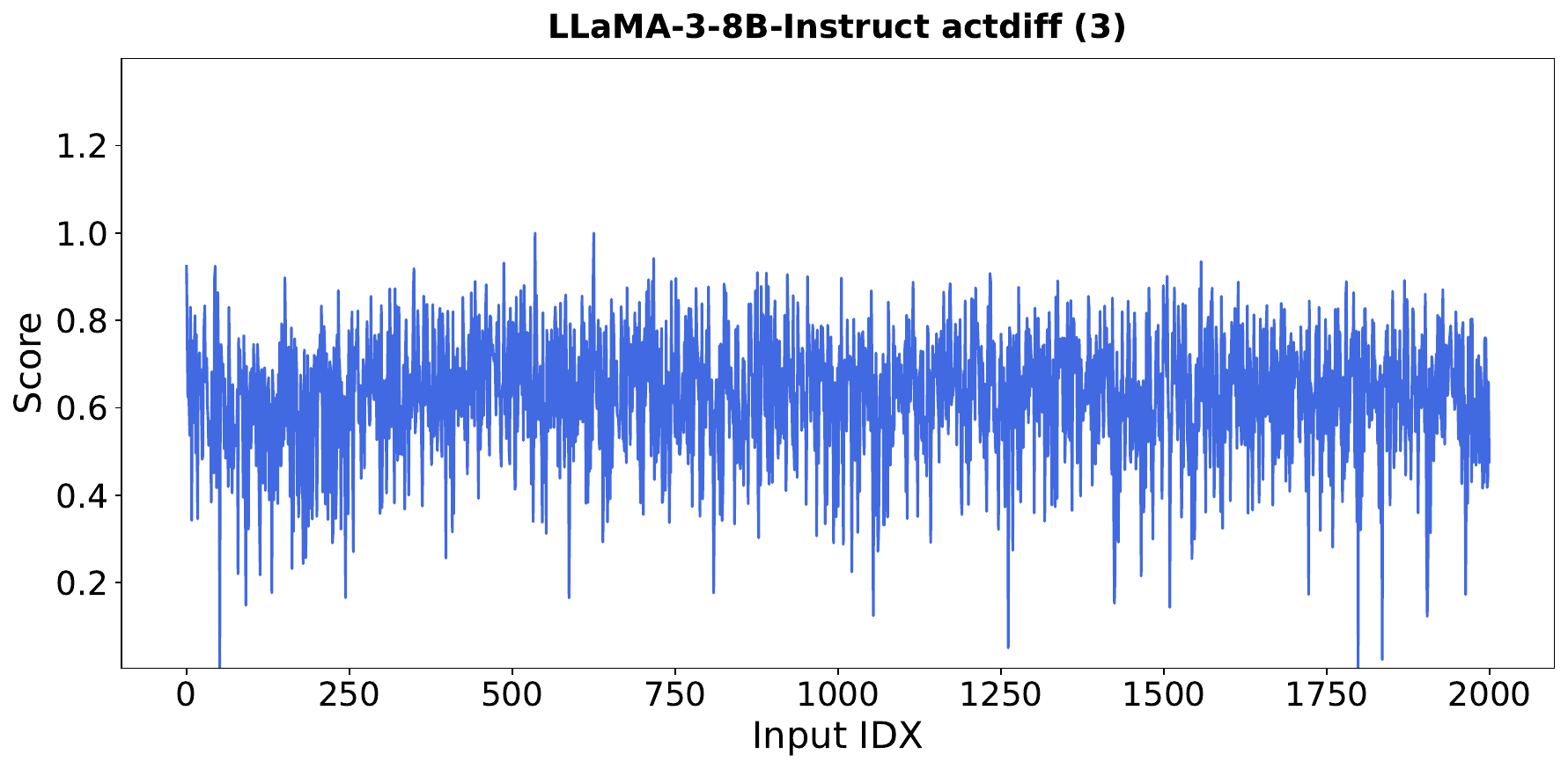}
        \captionsetup{width=0.4\linewidth}
        \caption{Layer 3}
    \end{subfigure}
    \begin{subfigure}[b]{0.33\linewidth}
        \centering
        \includegraphics[width=\linewidth]{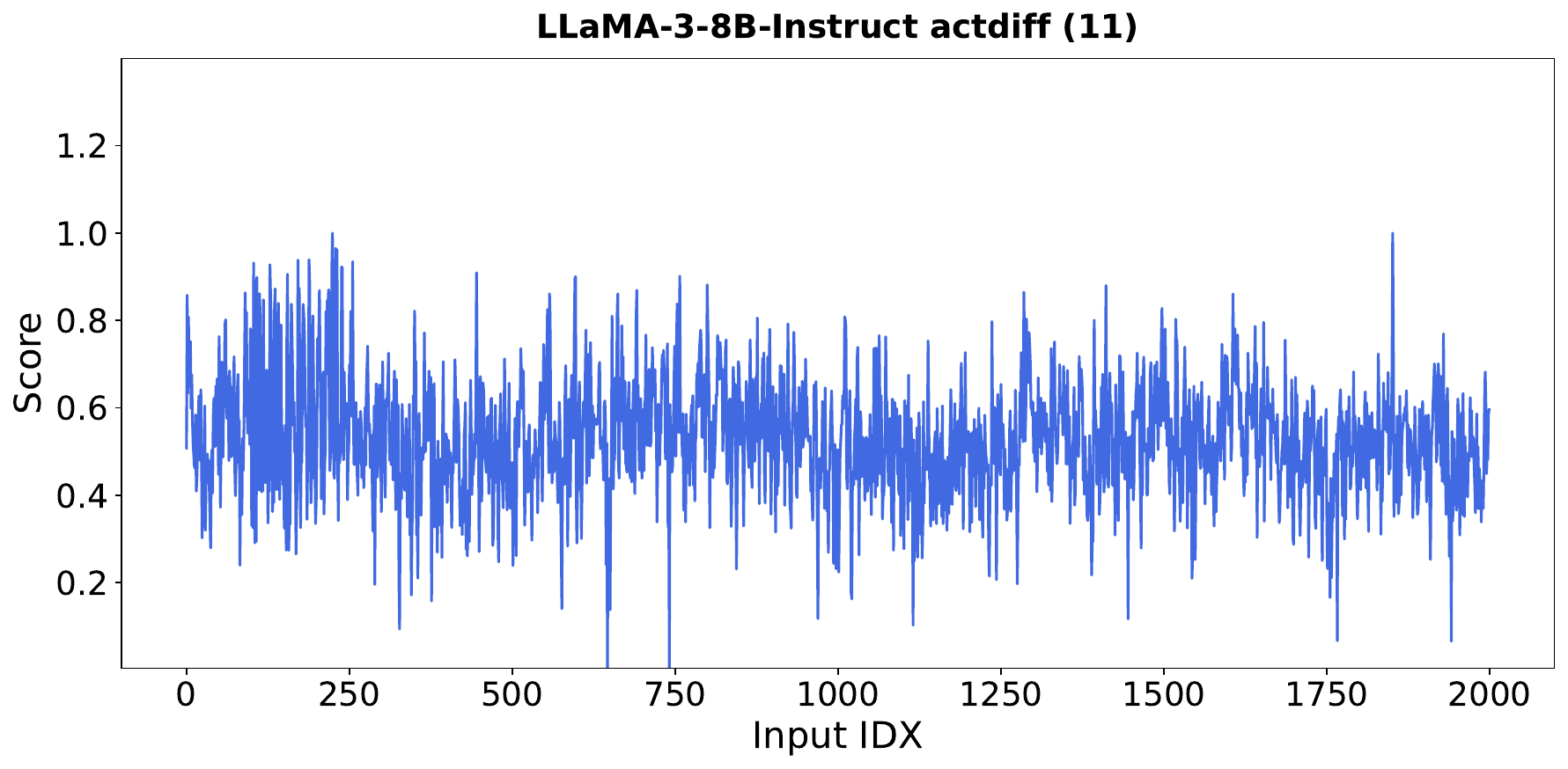}
        \captionsetup{width=0.4\linewidth}
        \caption{Layer 11}
     \end{subfigure}
     \begin{subfigure}[b]{0.33\linewidth}
        \centering
        \includegraphics[width=\linewidth]{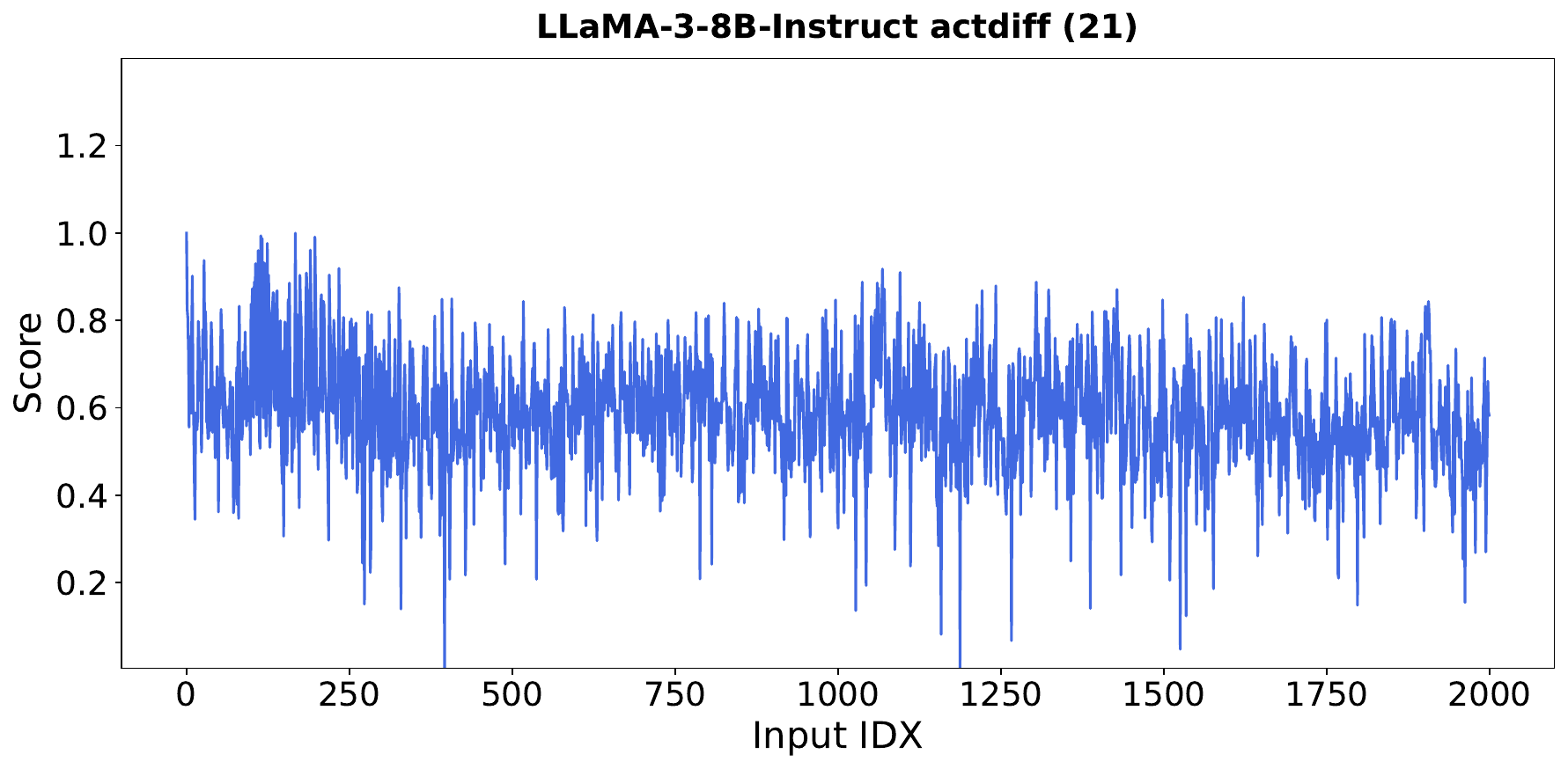}
        \captionsetup{width=0.4\linewidth}
        \caption{Layer 21}
     \end{subfigure}
    
    \vspace{0.3cm} 

    \begin{minipage}{\textwidth}
        \centering
        \textbf{Figures (a) - (c):} first example. 
    \end{minipage}
    \vspace{0.3cm}

    \begin{subfigure}[b]{0.33\linewidth}
    \centering
        \includegraphics[width=\linewidth]{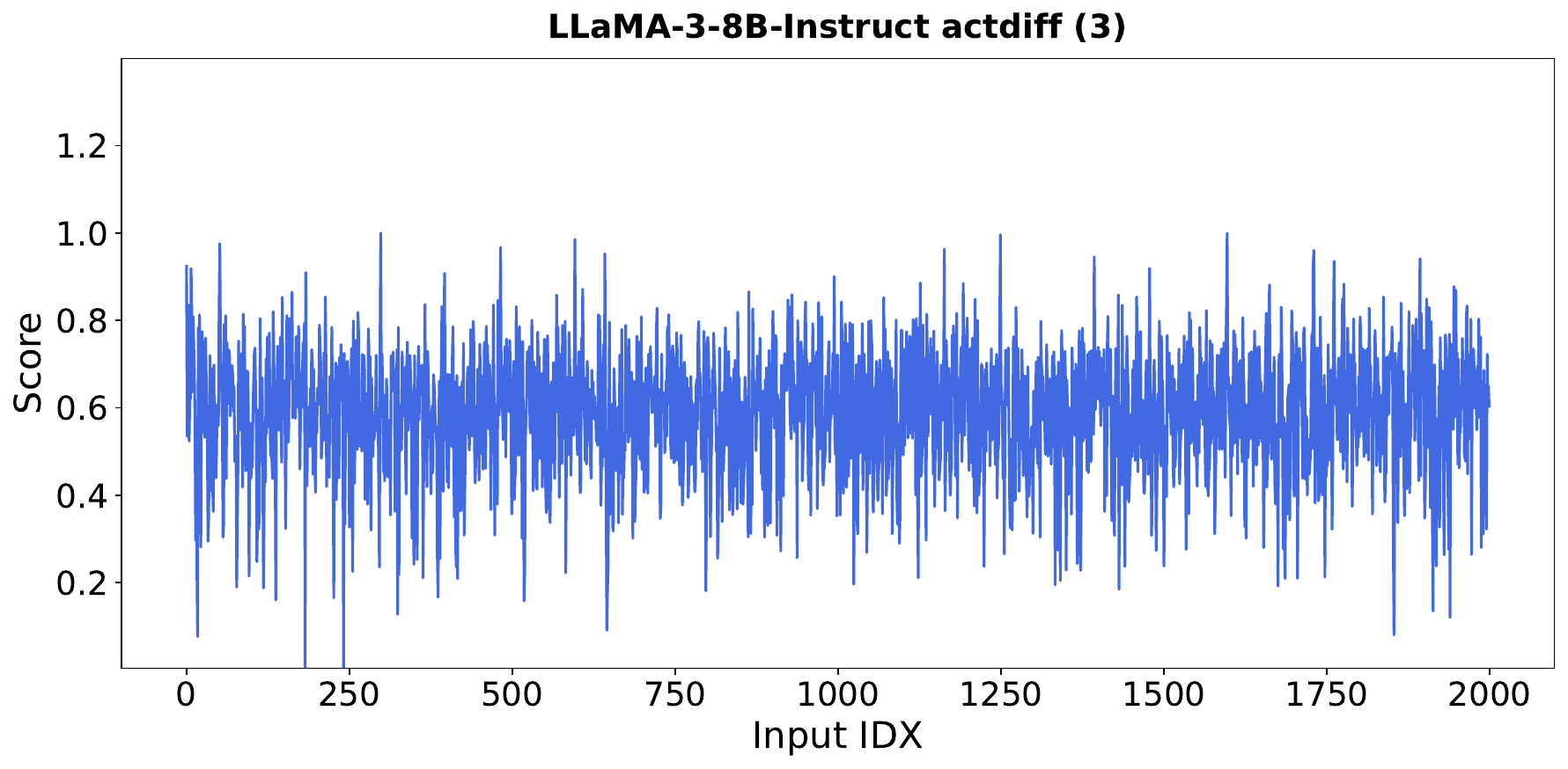}
        \captionsetup{width=0.4\linewidth}
        \caption{Layer 3}
    \end{subfigure}
    \begin{subfigure}[b]{0.33\linewidth}
        \centering
        \includegraphics[width=\linewidth]{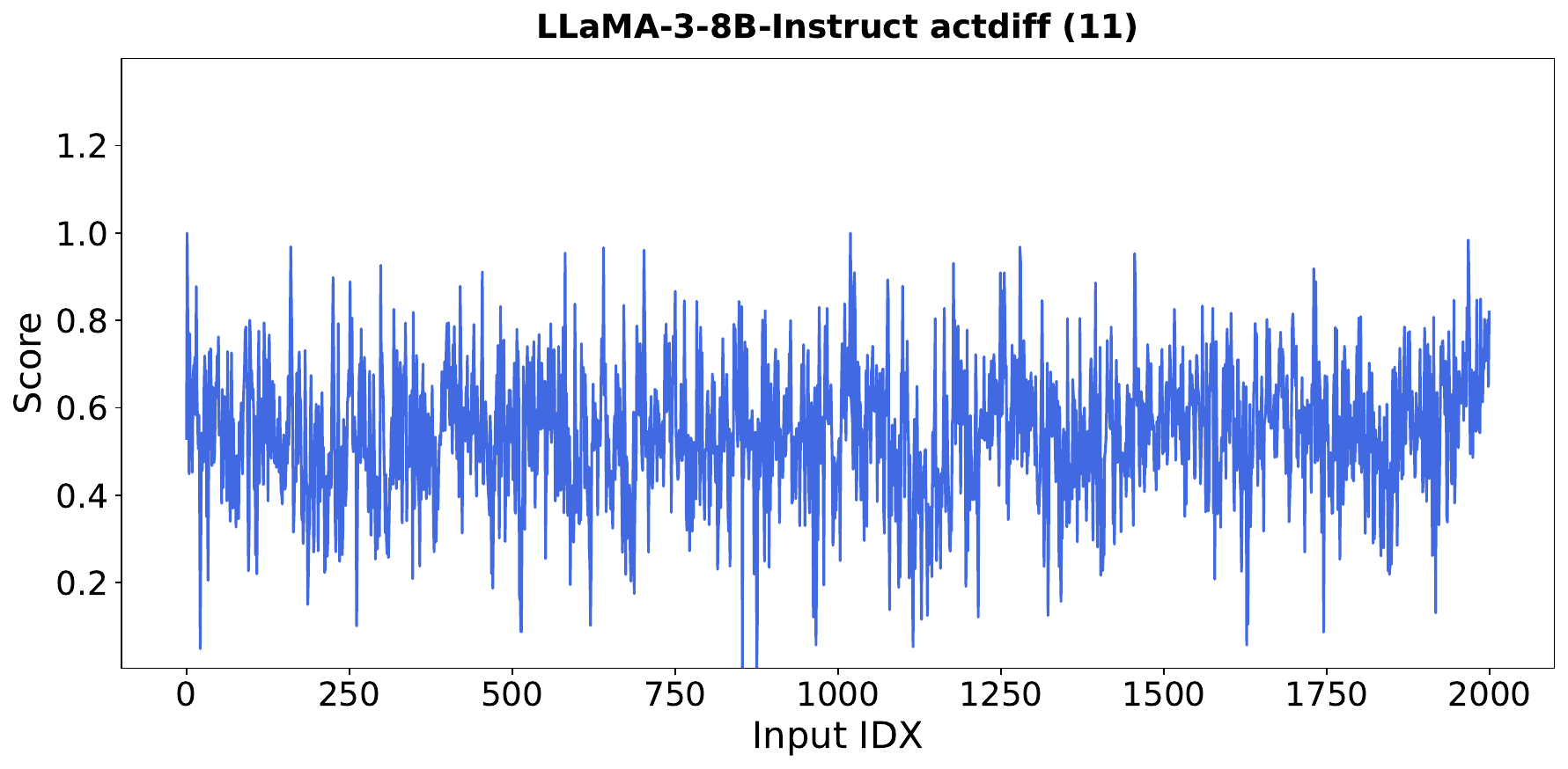}
        \captionsetup{width=0.4\linewidth}
        \caption{Layer 11}
     \end{subfigure}
     \begin{subfigure}[b]{0.33\linewidth}
        \centering
        \includegraphics[width=\linewidth]{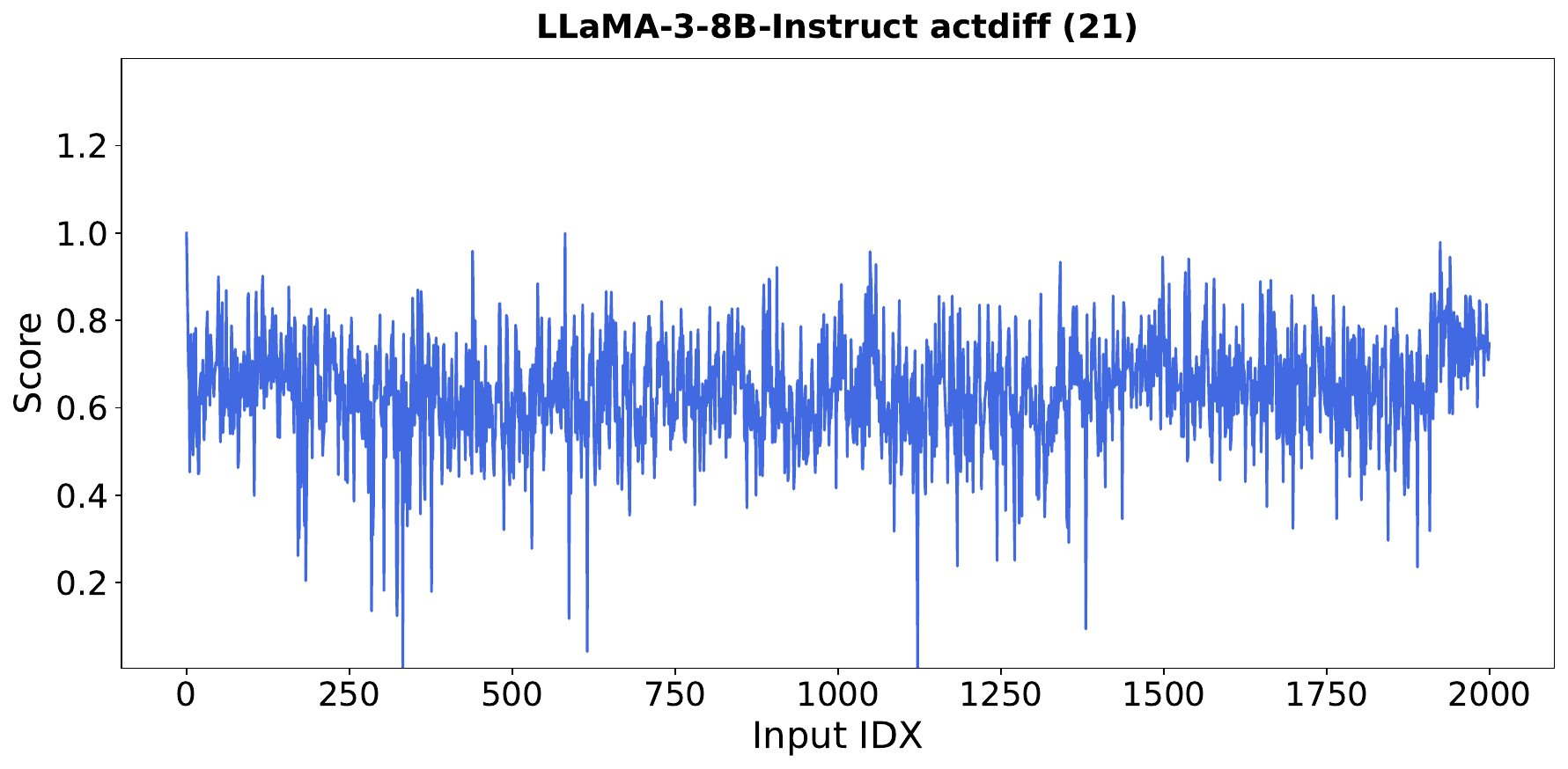}
        \captionsetup{width=0.4\linewidth}
        \caption{Layer 21}
     \end{subfigure}
    
    \vspace{0.3cm} 

     \begin{minipage}{\textwidth}
        \centering
        \textbf{Figures (d) - (f):} second example. 
    \end{minipage}
    \vspace{0.3cm}
    
    \begin{subfigure}[b]{0.33\linewidth}
    \centering
        \includegraphics[width=\linewidth]{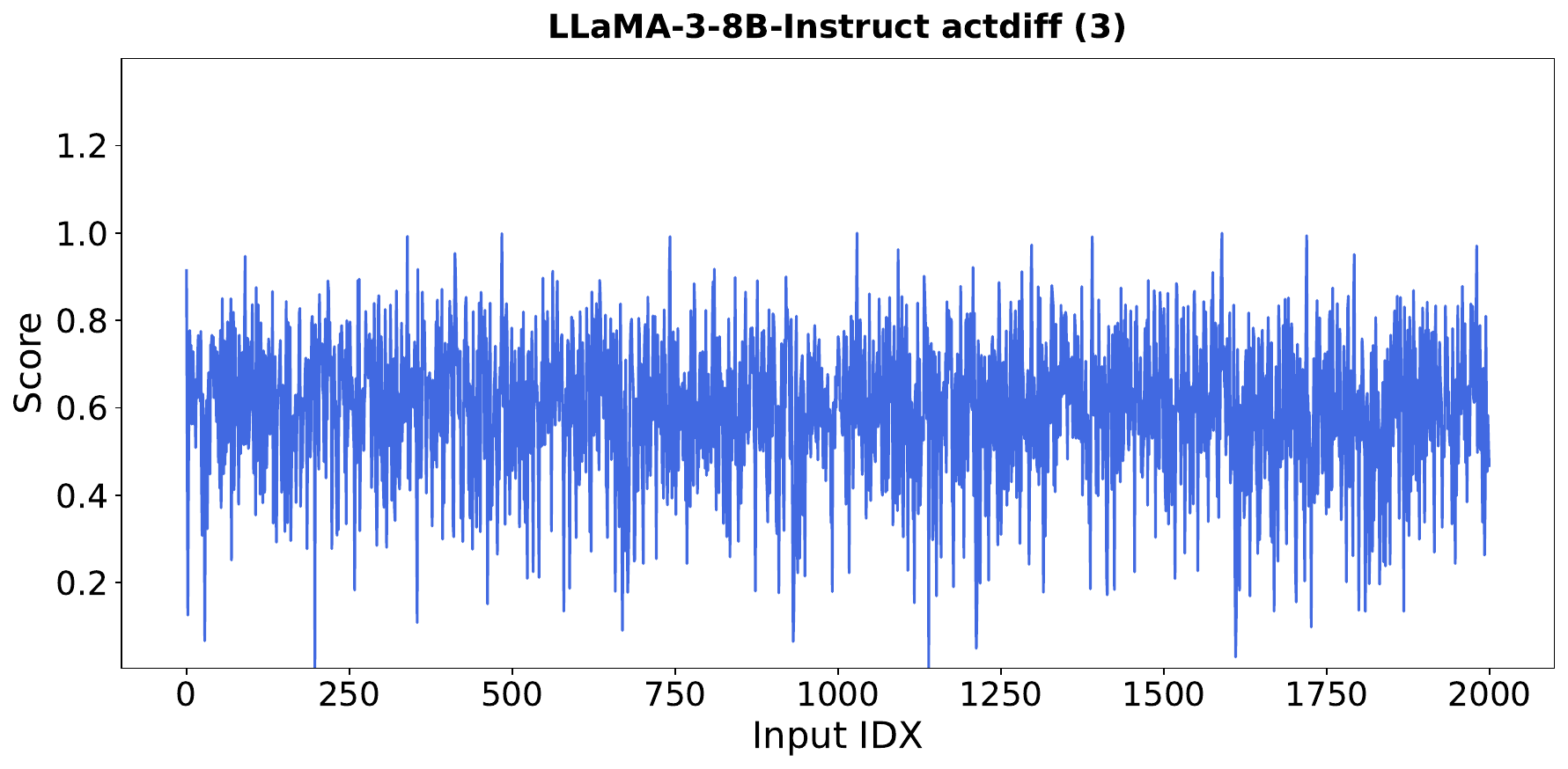}
        \captionsetup{width=0.4\linewidth}
        \caption{Layer 3}
    \end{subfigure}
    \begin{subfigure}[b]{0.33\linewidth}
        \centering
        \includegraphics[width=\linewidth]{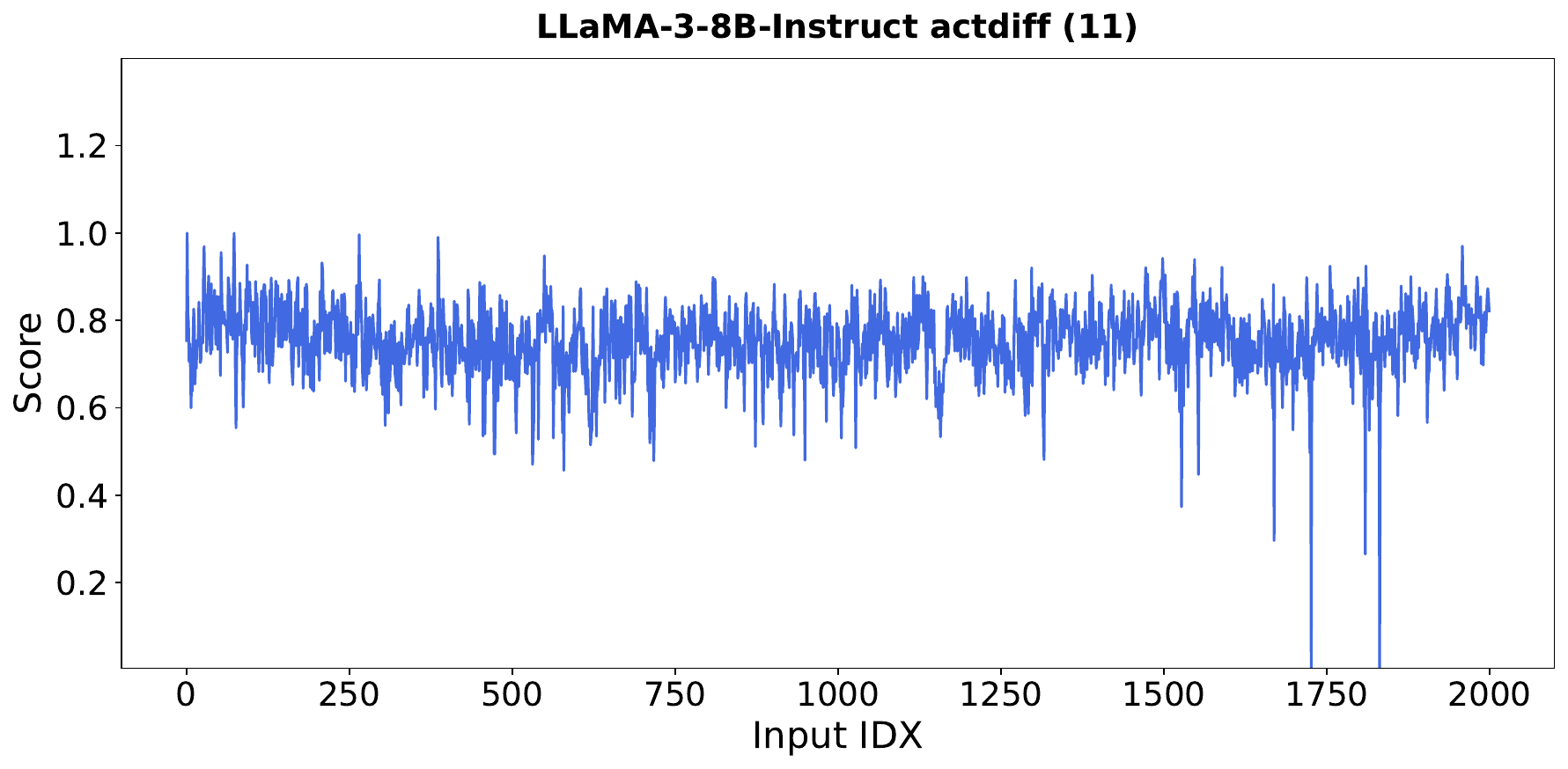}
        \captionsetup{width=0.4\linewidth}
        \caption{Layer 11}
     \end{subfigure}
     \begin{subfigure}[b]{0.33\linewidth}
        \centering
        \includegraphics[width=\linewidth]{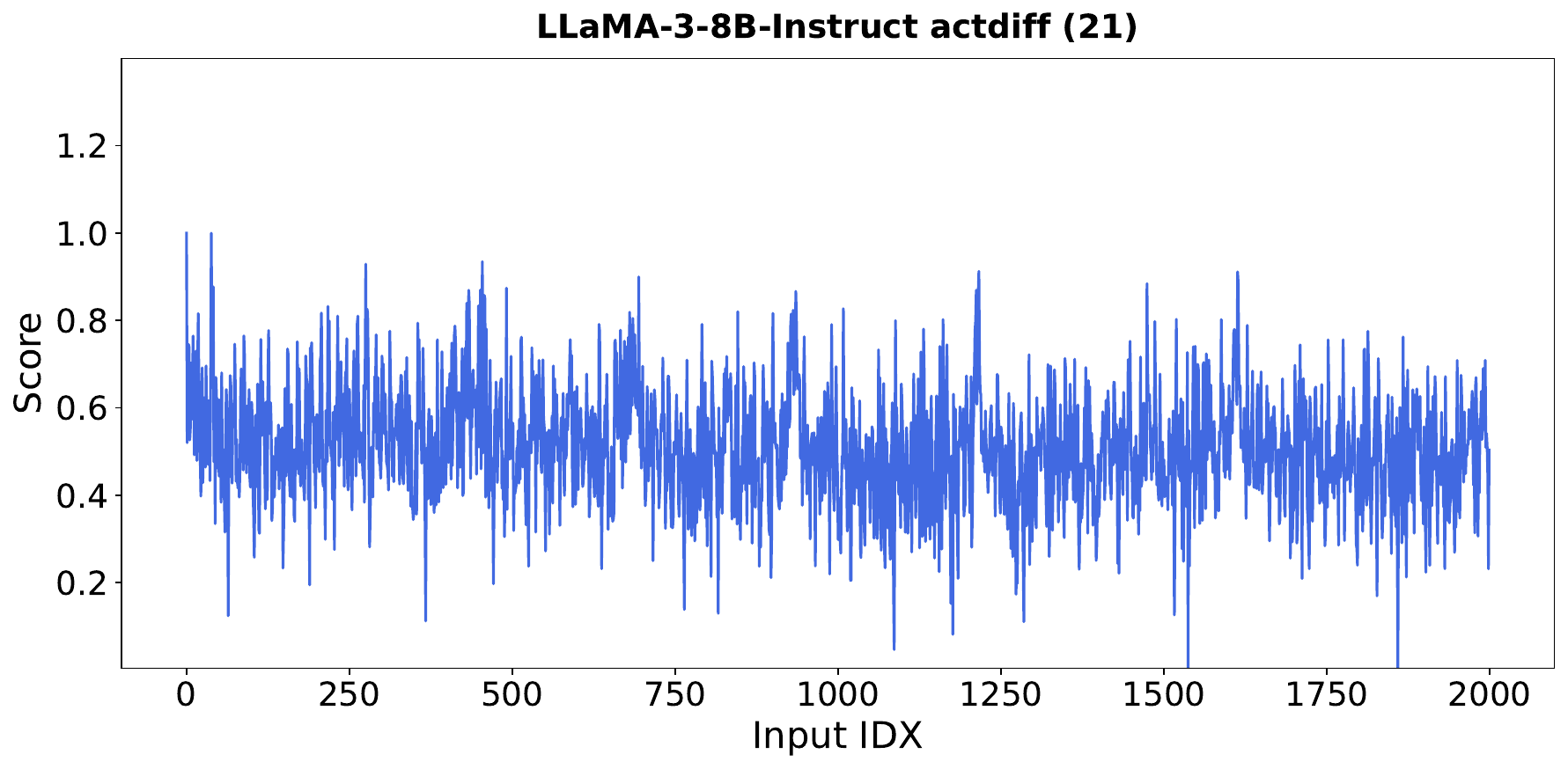}
        \captionsetup{width=0.4\linewidth}
        \caption{Layer 21}
     \end{subfigure}
    
    \vspace{0.3cm} 

     \begin{minipage}{\textwidth}
        \centering
        \textbf{Figures (g) - (i):} third example. 
    \end{minipage}
    \vspace{0.3cm}
    \caption{Visualization of ActDiff scores across three layers for three different examples.}
    \label{fig:actdiff_visualization}
\end{figure*}

%% file: tables_figures/tokensim_visualization.tex
\begin{figure*}[t]
    \centering
    \begin{subfigure}[b]{0.33\linewidth}
    \centering
        \includegraphics[width=\linewidth]{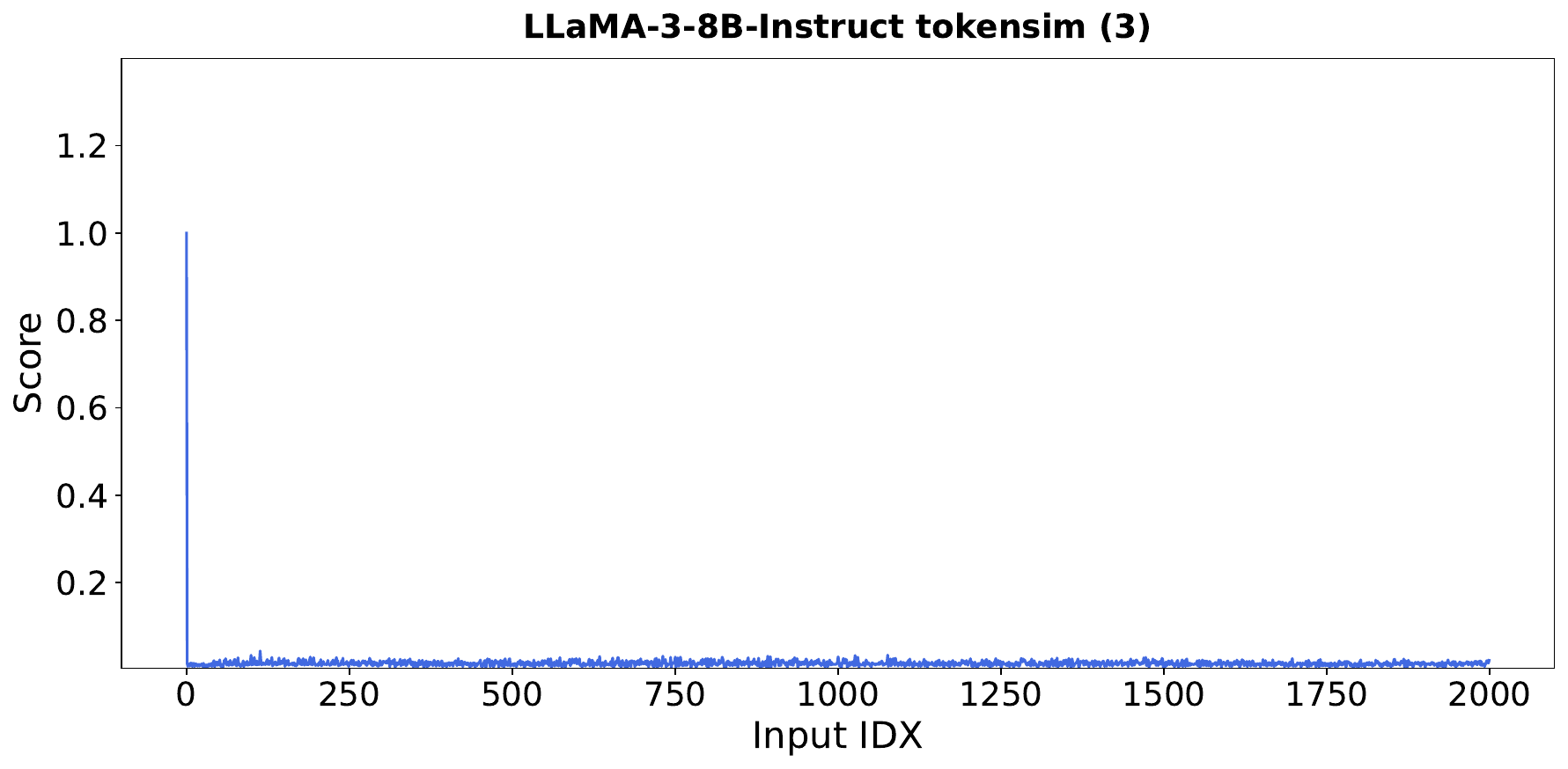}
        \captionsetup{width=0.4\linewidth}
        \caption{Layer 3}
    \end{subfigure}
    \begin{subfigure}[b]{0.33\linewidth}
        \centering
        \includegraphics[width=\linewidth]{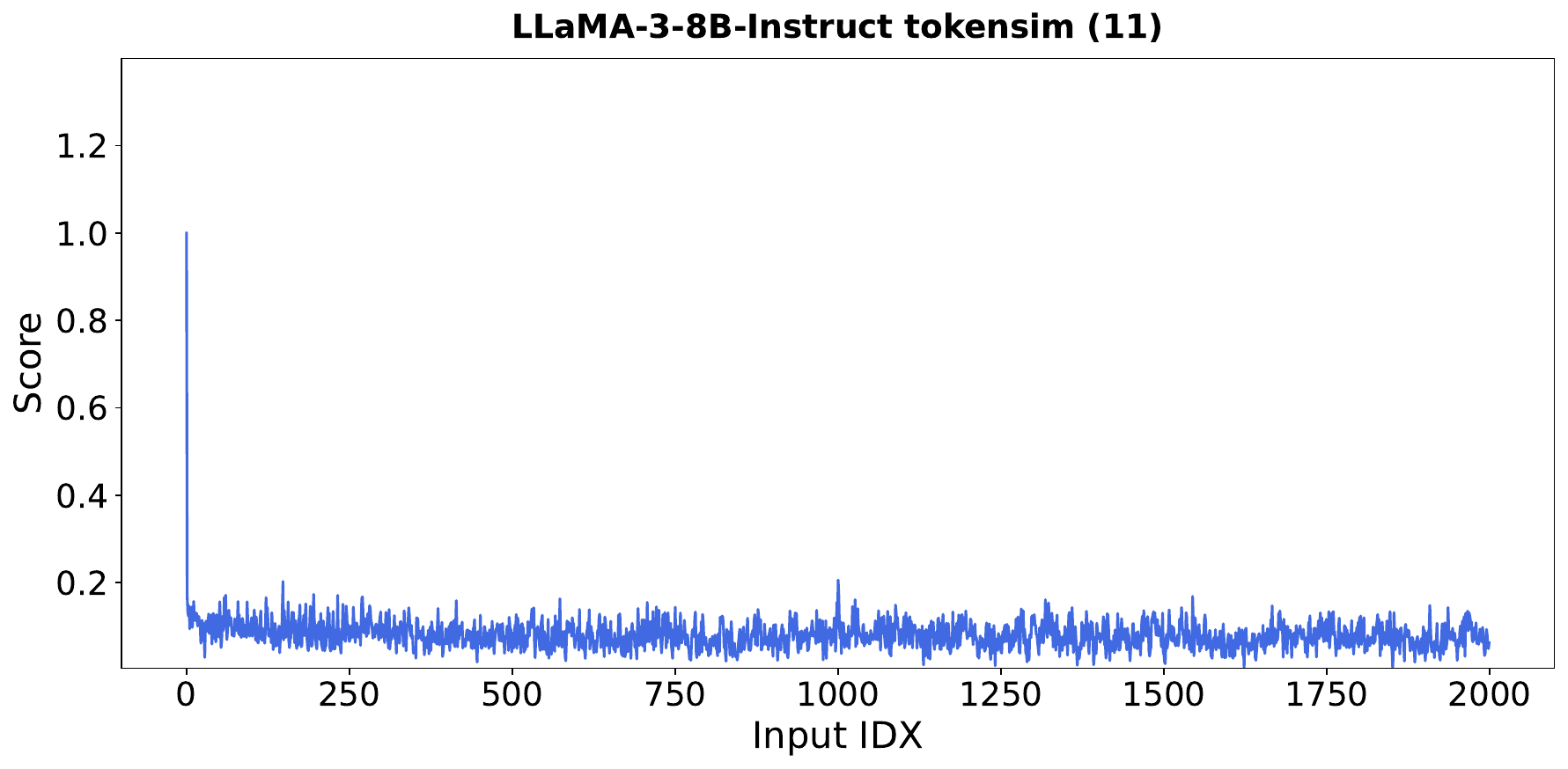}
        \captionsetup{width=0.4\linewidth}
        \caption{Layer 11}
     \end{subfigure}
     \begin{subfigure}[b]{0.33\linewidth}
        \centering
        \includegraphics[width=\linewidth]{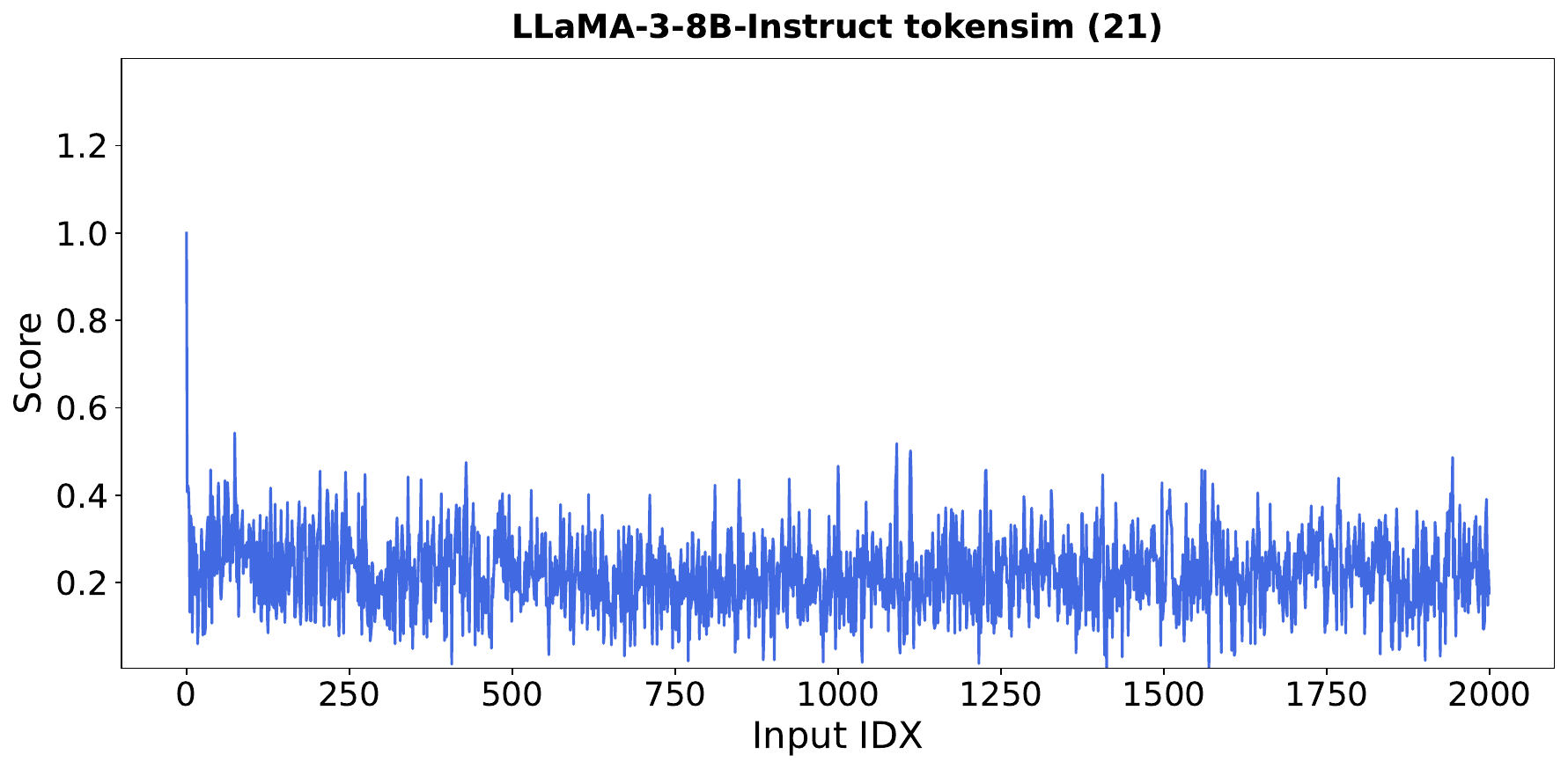}
        \captionsetup{width=0.4\linewidth}
        \caption{Layer 21}
     \end{subfigure}
    
    \vspace{0.3cm} 

    \begin{minipage}{\textwidth}
        \centering
        \textbf{Figures (a) - (c):} first example. 
    \end{minipage}
    \vspace{0.3cm}

    \begin{subfigure}[b]{0.33\linewidth}
    \centering
        \includegraphics[width=\linewidth]{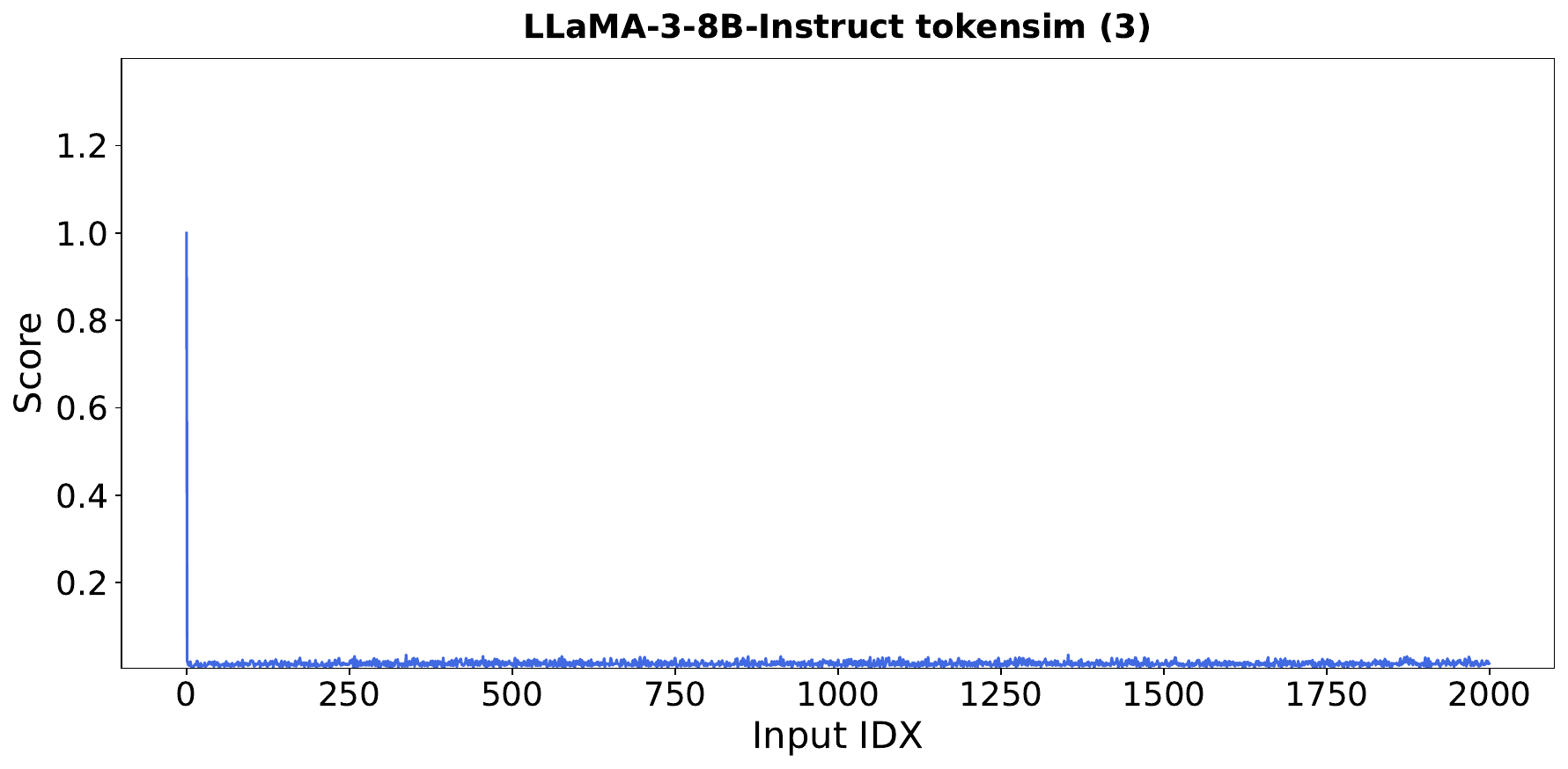}
        \captionsetup{width=0.4\linewidth}
        \caption{Layer 3}
    \end{subfigure}
    \begin{subfigure}[b]{0.33\linewidth}
        \centering
        \includegraphics[width=\linewidth]{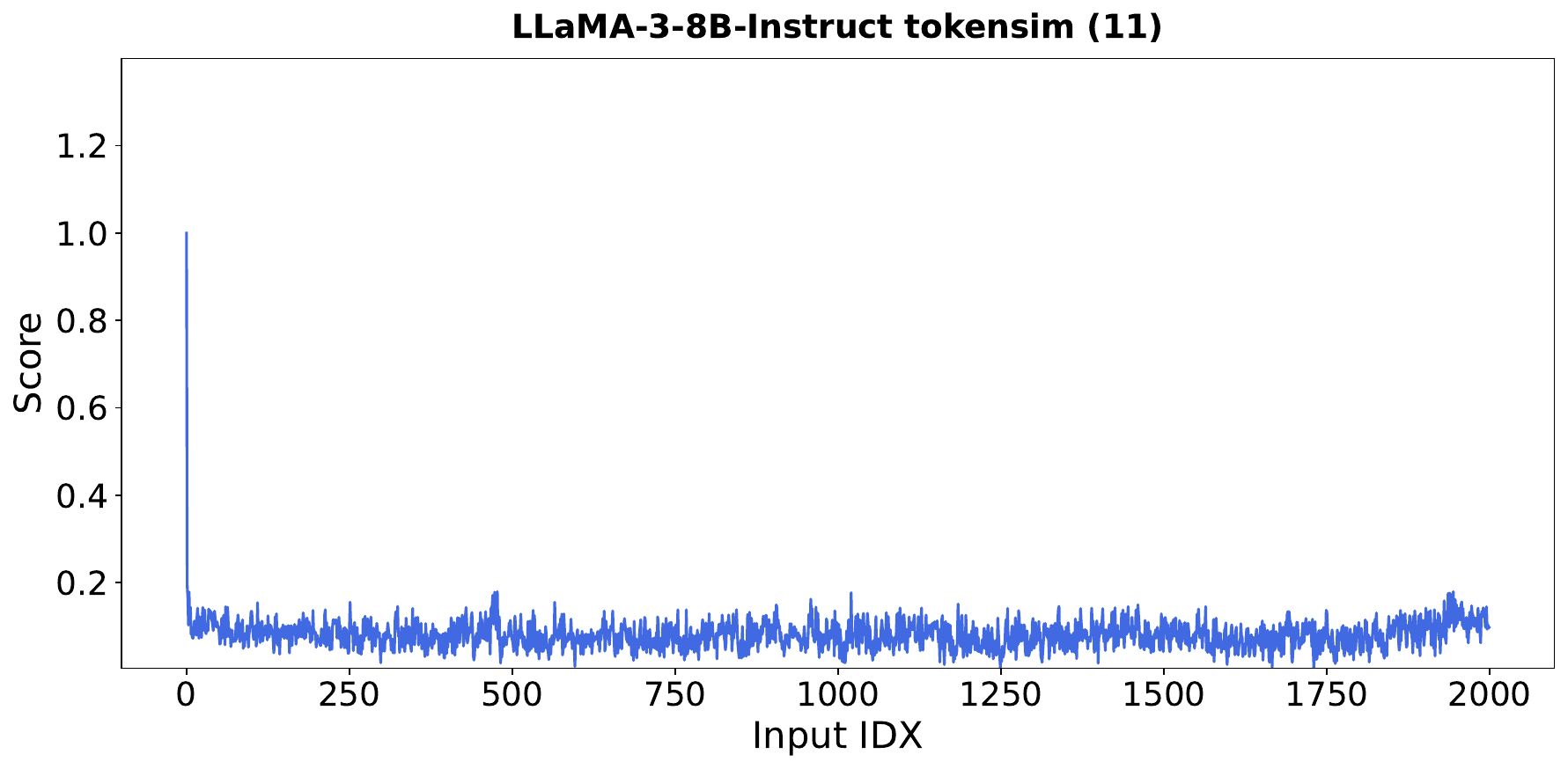}
        \captionsetup{width=0.4\linewidth}
        \caption{Layer 11}
     \end{subfigure}
     \begin{subfigure}[b]{0.33\linewidth}
        \centering
        \includegraphics[width=\linewidth]{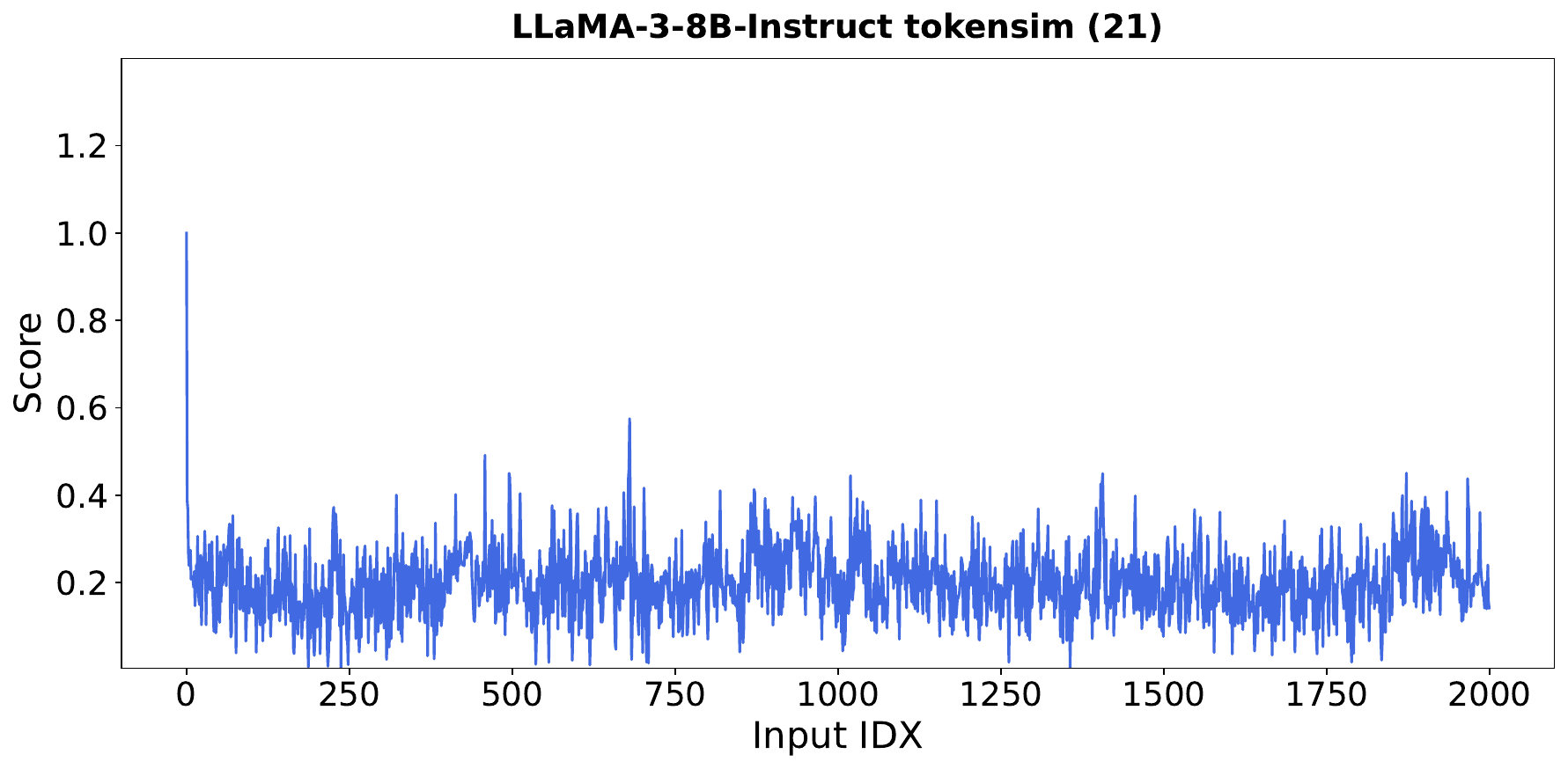}
        \captionsetup{width=0.4\linewidth}
        \caption{Layer 21}
     \end{subfigure}
    
    \vspace{0.3cm} 

     \begin{minipage}{\textwidth}
        \centering
        \textbf{Figures (d) - (f):} second example. 
    \end{minipage}
    \vspace{0.3cm}
    
    \begin{subfigure}[b]{0.33\linewidth}
    \centering
        \includegraphics[width=\linewidth]{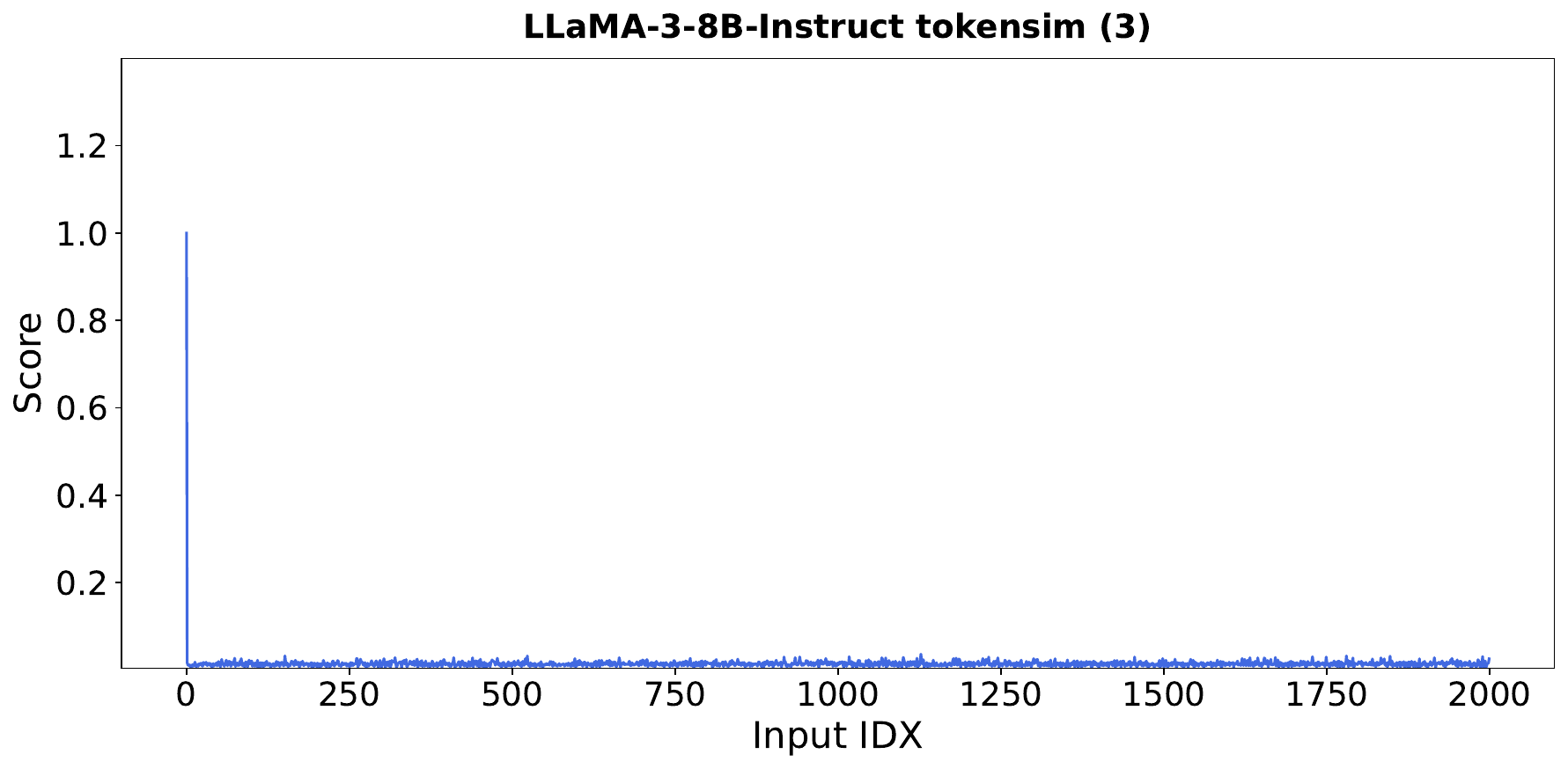}
        \captionsetup{width=0.4\linewidth}
        \caption{Layer 3}
    \end{subfigure}
    \begin{subfigure}[b]{0.33\linewidth}
        \centering
        \includegraphics[width=\linewidth]{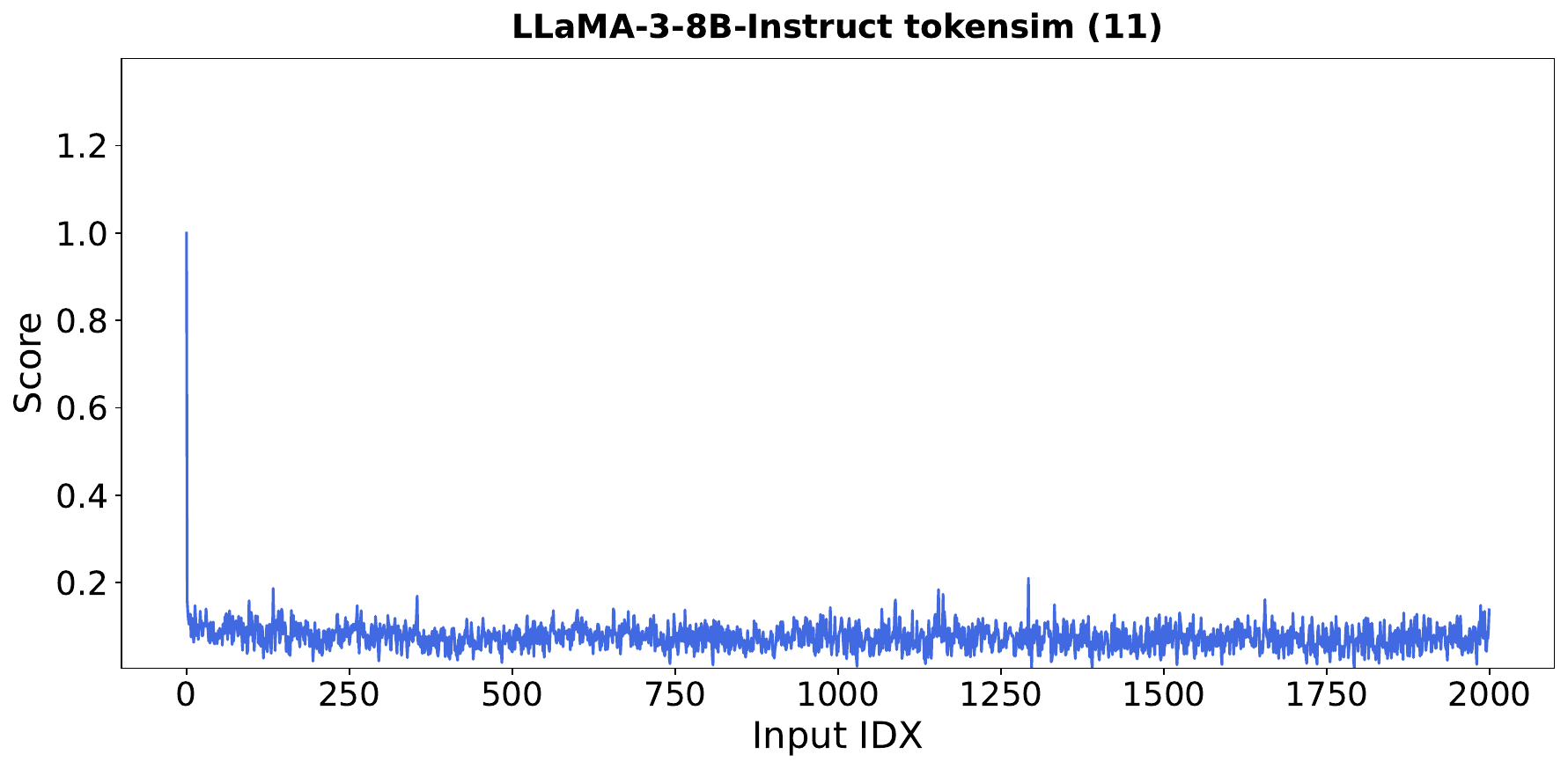}
        \captionsetup{width=0.4\linewidth}
        \caption{Layer 11}
     \end{subfigure}
     \begin{subfigure}[b]{0.33\linewidth}
        \centering
        \includegraphics[width=\linewidth]{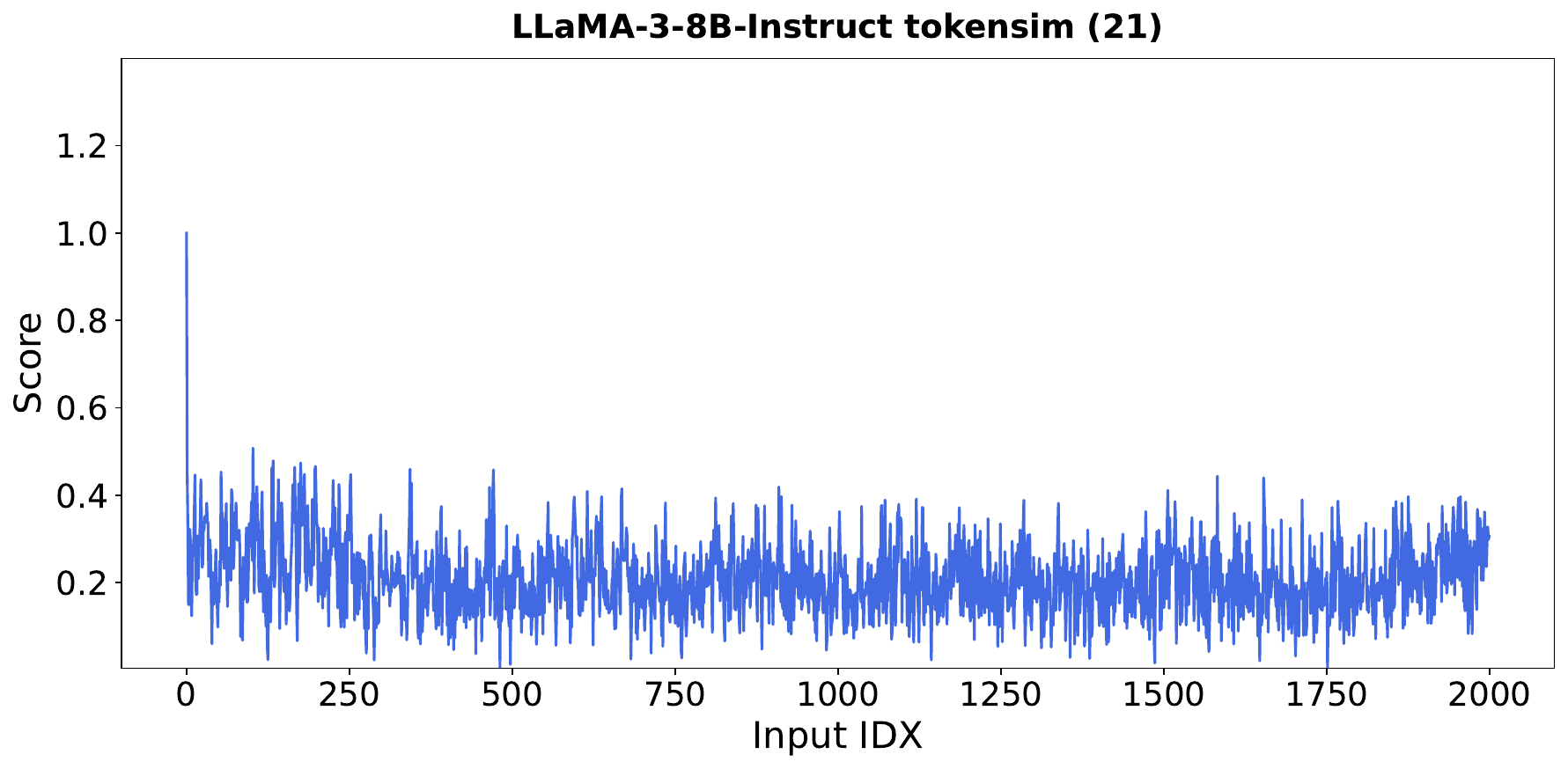}
        \captionsetup{width=0.4\linewidth}
        \caption{Layer 21}
     \end{subfigure}
    
    \vspace{0.3cm} 

     \begin{minipage}{\textwidth}
        \centering
        \textbf{Figures (g) - (i):} third example. 
    \end{minipage}
    \vspace{0.3cm}
    \caption{Visualization of TokenSim scores across three layers for three different examples.}
    \label{fig:tokensim_visualization}
\end{figure*}

%% file: tables_figures/attncon_visualization.tex
\begin{figure*}[t]
    \centering
    \begin{subfigure}[b]{0.33\linewidth}
    \centering
        \includegraphics[width=\linewidth]{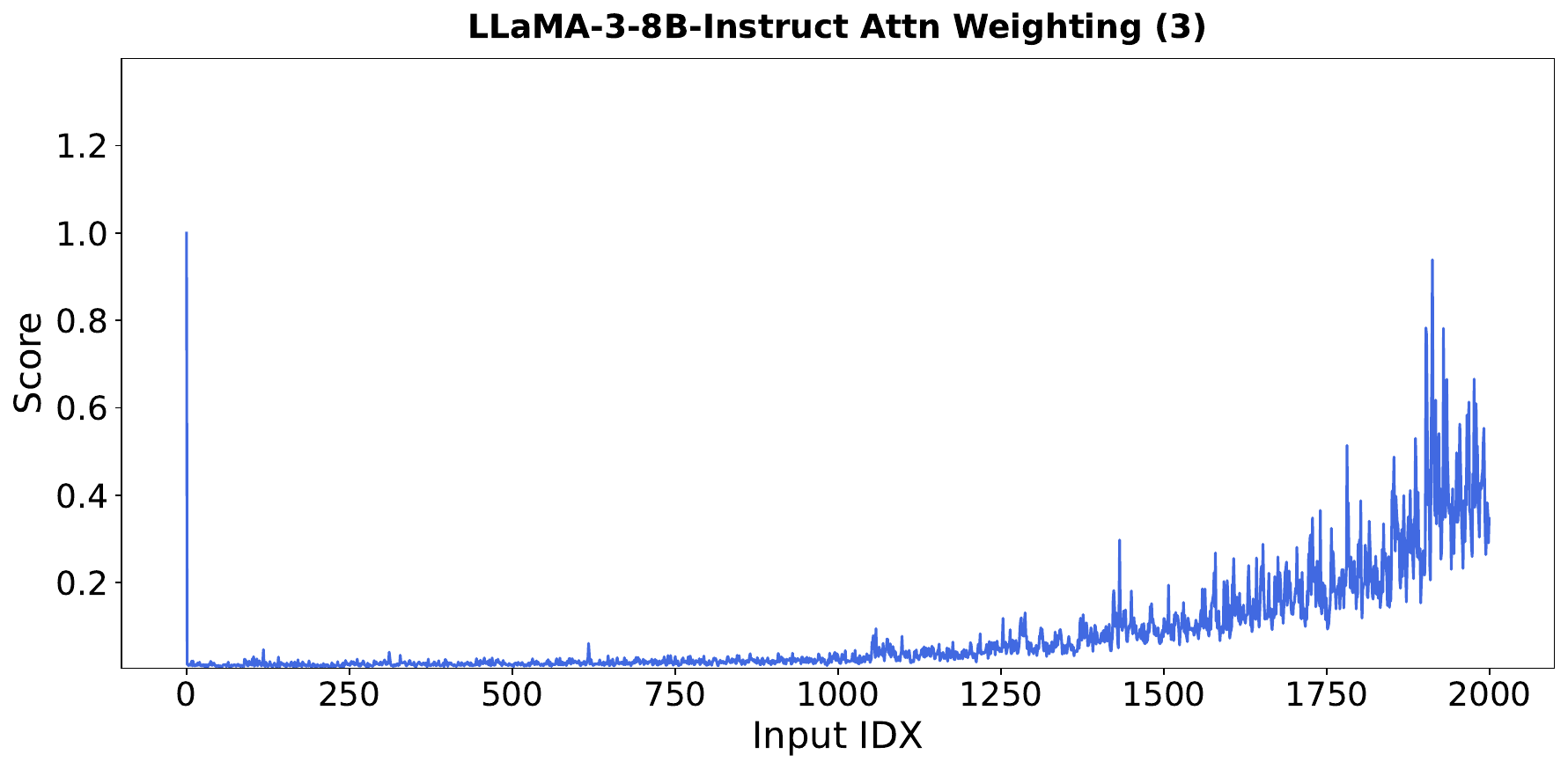}
        \captionsetup{width=0.4\linewidth}
        \caption{Layer 3}
    \end{subfigure}
    \begin{subfigure}[b]{0.33\linewidth}
        \centering
        \includegraphics[width=\linewidth]{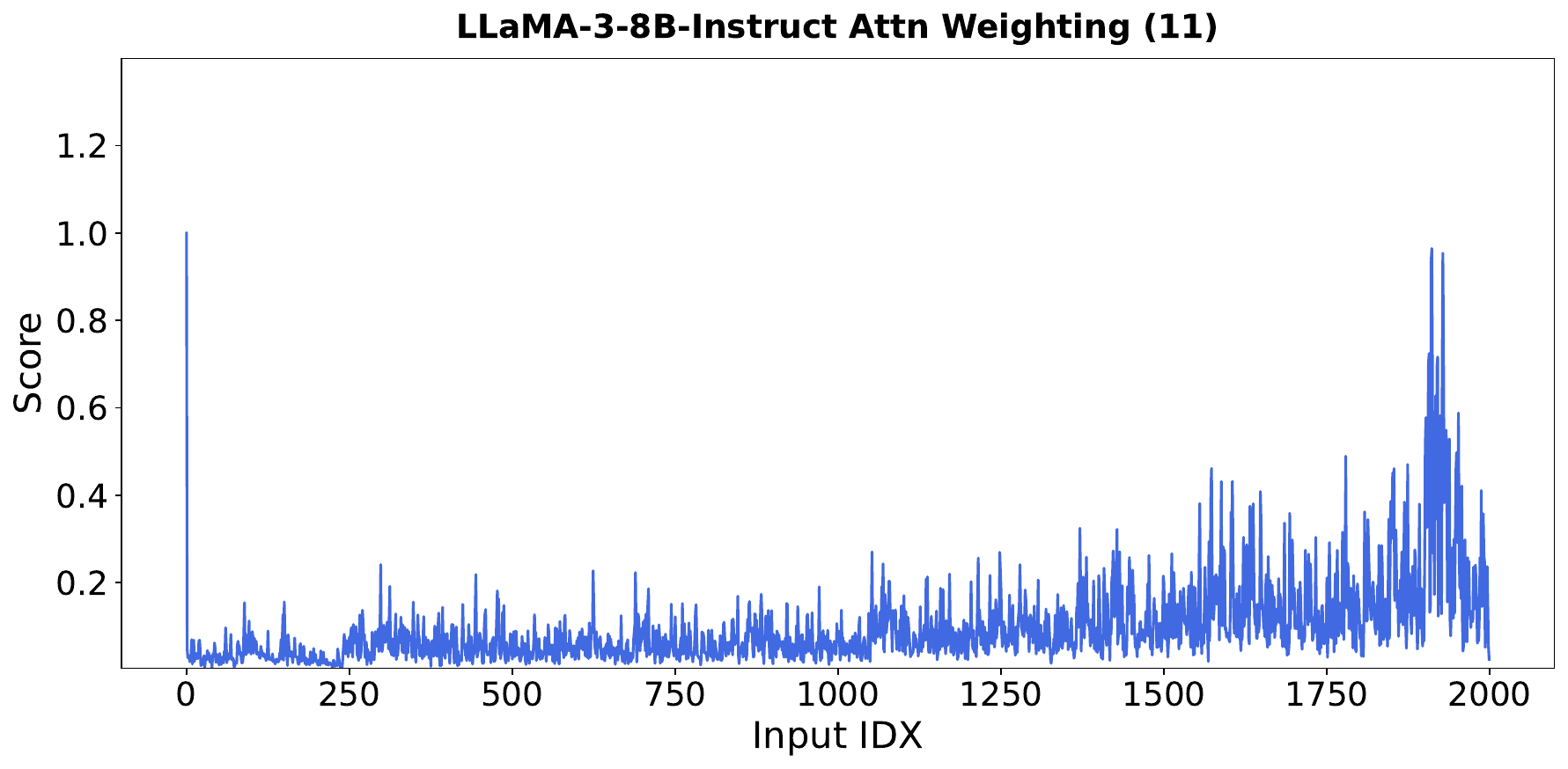}
        \captionsetup{width=0.4\linewidth}
        \caption{Layer 11}
     \end{subfigure}
     \begin{subfigure}[b]{0.33\linewidth}
        \centering
        \includegraphics[width=\linewidth]{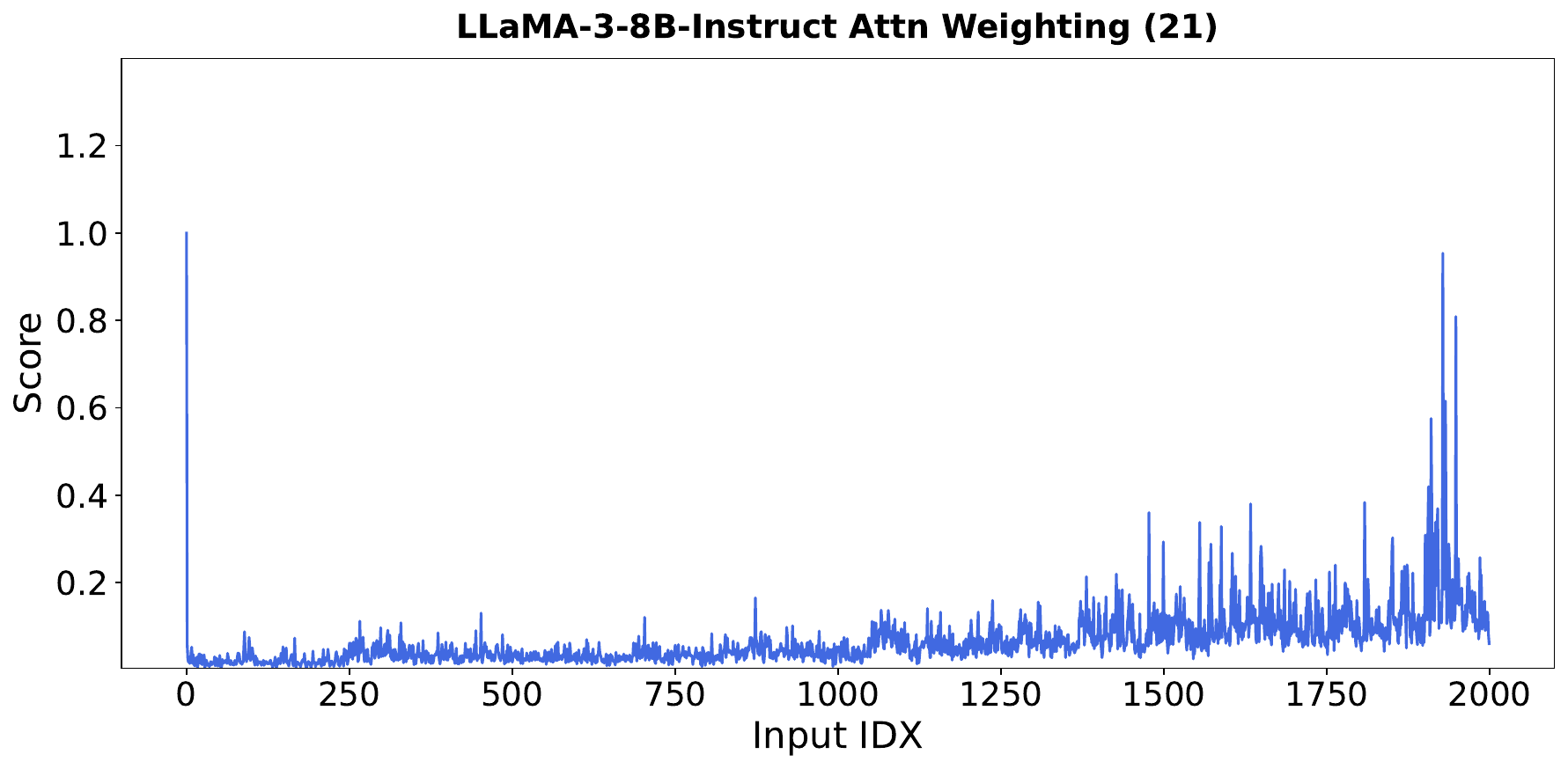}
        \captionsetup{width=0.4\linewidth}
        \caption{Layer 21}
     \end{subfigure}
    
    \vspace{0.3cm} 

    \begin{minipage}{\textwidth}
        \centering
        \textbf{Figures (a) - (c):} first example. 
    \end{minipage}
    \vspace{0.3cm}

    \begin{subfigure}[b]{0.33\linewidth}
    \centering
        \includegraphics[width=\linewidth]{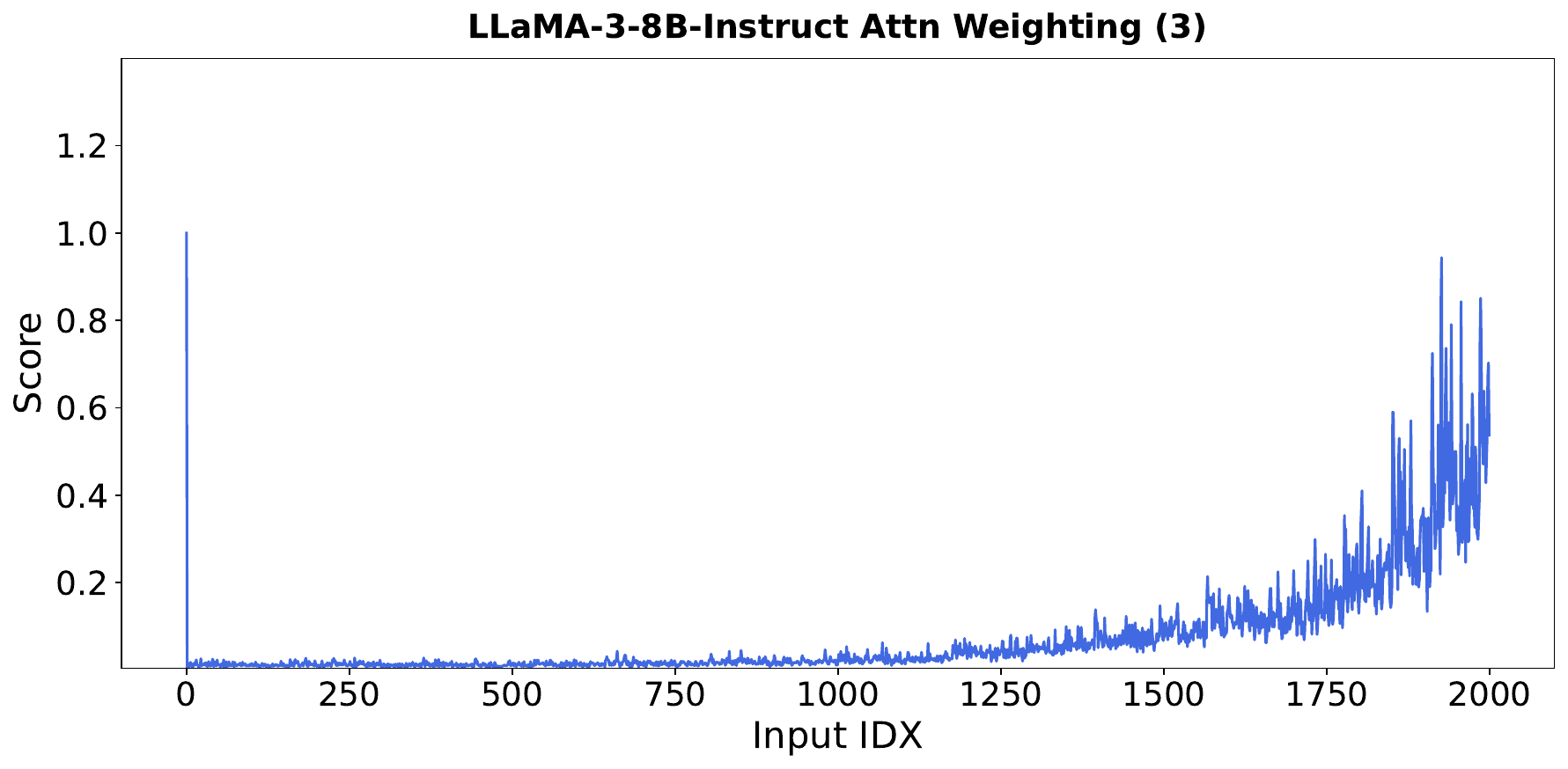}
        \captionsetup{width=0.4\linewidth}
        \caption{Layer 3}
    \end{subfigure}
    \begin{subfigure}[b]{0.33\linewidth}
        \centering
        \includegraphics[width=\linewidth]{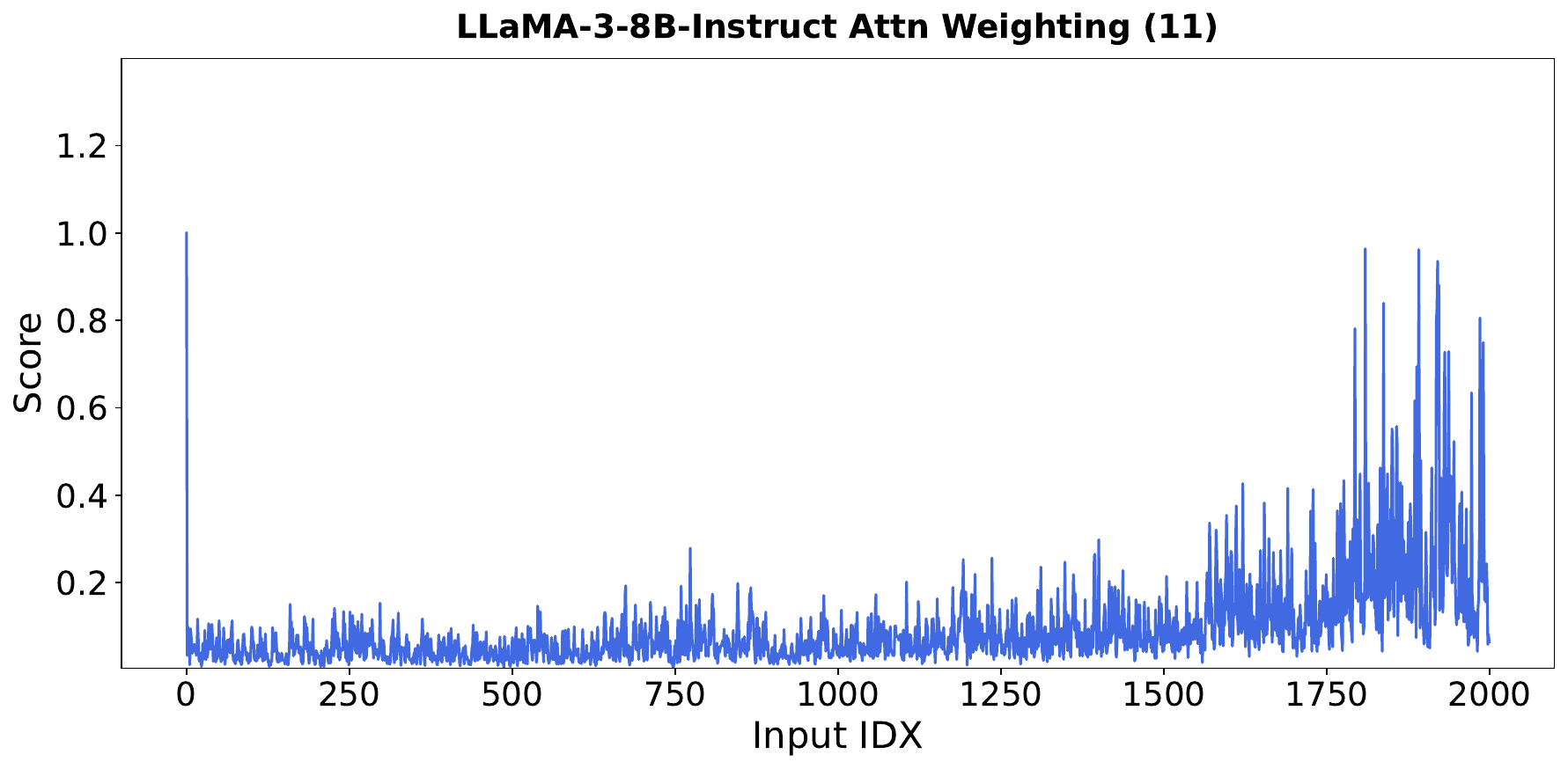}
        \captionsetup{width=0.4\linewidth}
        \caption{Layer 11}
     \end{subfigure}
     \begin{subfigure}[b]{0.33\linewidth}
        \centering
        \includegraphics[width=\linewidth]{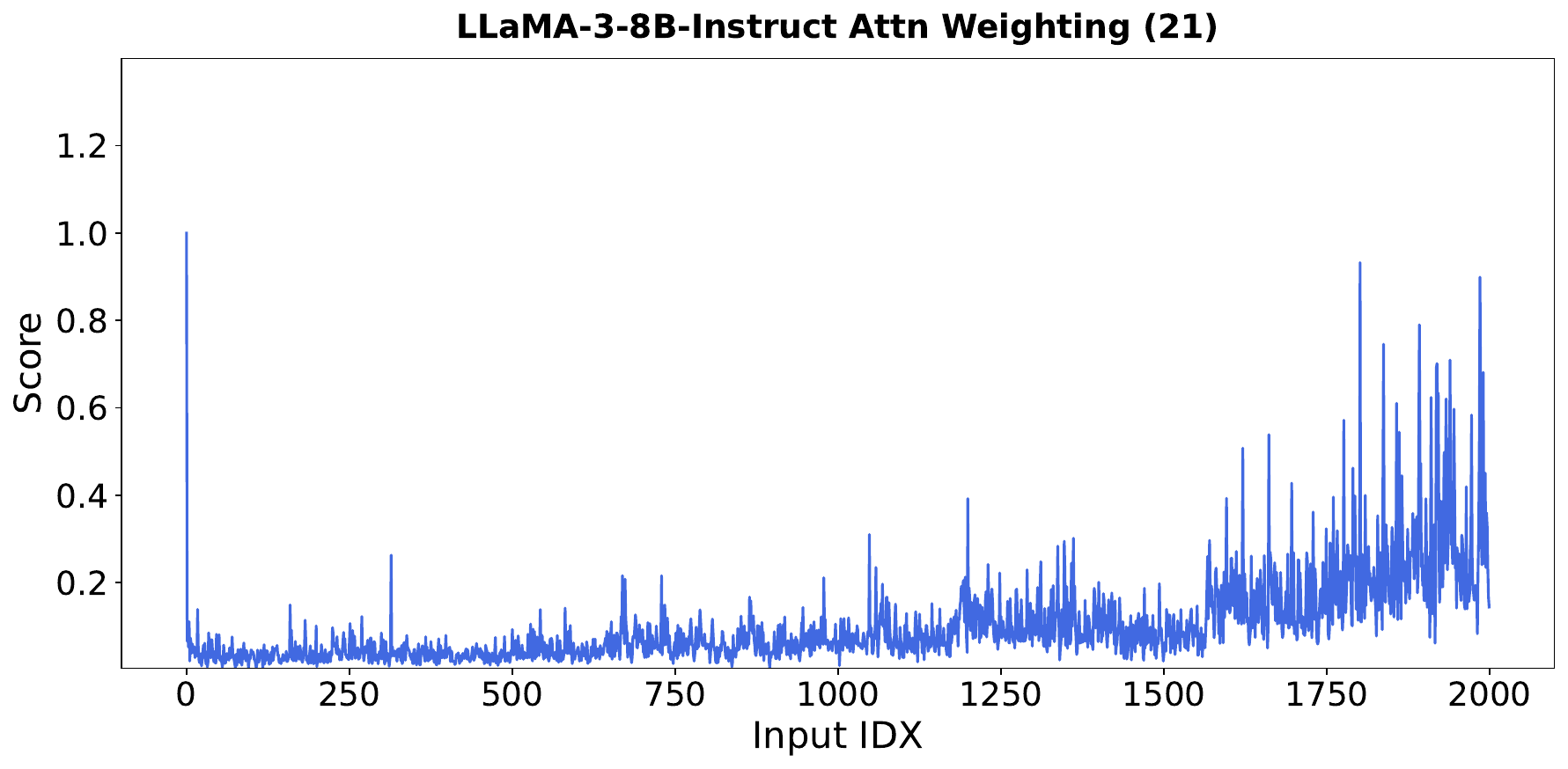}
        \captionsetup{width=0.4\linewidth}
        \caption{Layer 21}
     \end{subfigure}
    
    \vspace{0.3cm} 

     \begin{minipage}{\textwidth}
        \centering
        \textbf{Figures (d) - (f):} second example. 
    \end{minipage}
    \vspace{0.3cm}
    
    \begin{subfigure}[b]{0.33\linewidth}
    \centering
        \includegraphics[width=\linewidth]{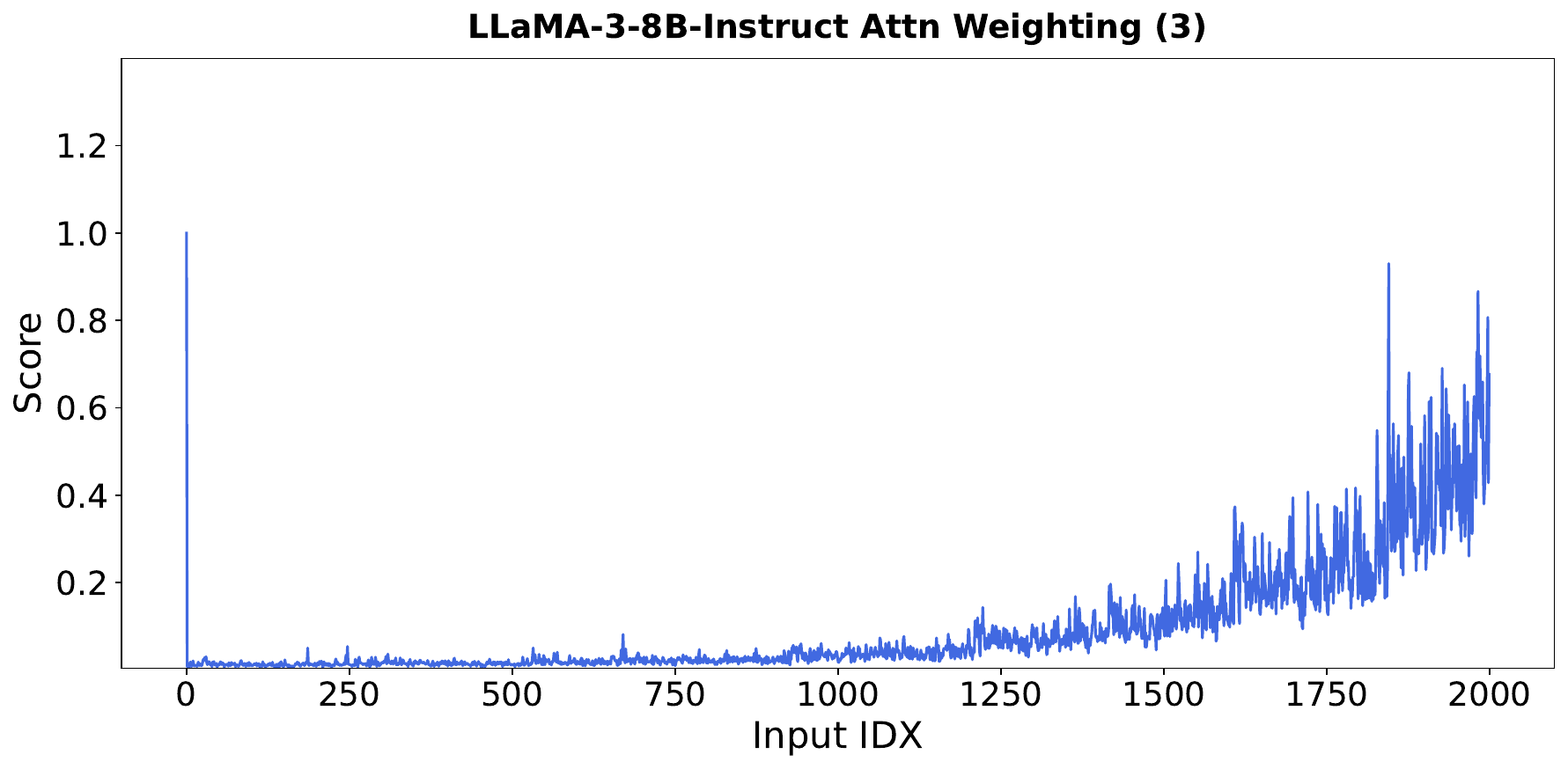}
        \captionsetup{width=0.4\linewidth}
        \caption{Layer 3}
    \end{subfigure}
    \begin{subfigure}[b]{0.33\linewidth}
        \centering
        \includegraphics[width=\linewidth]{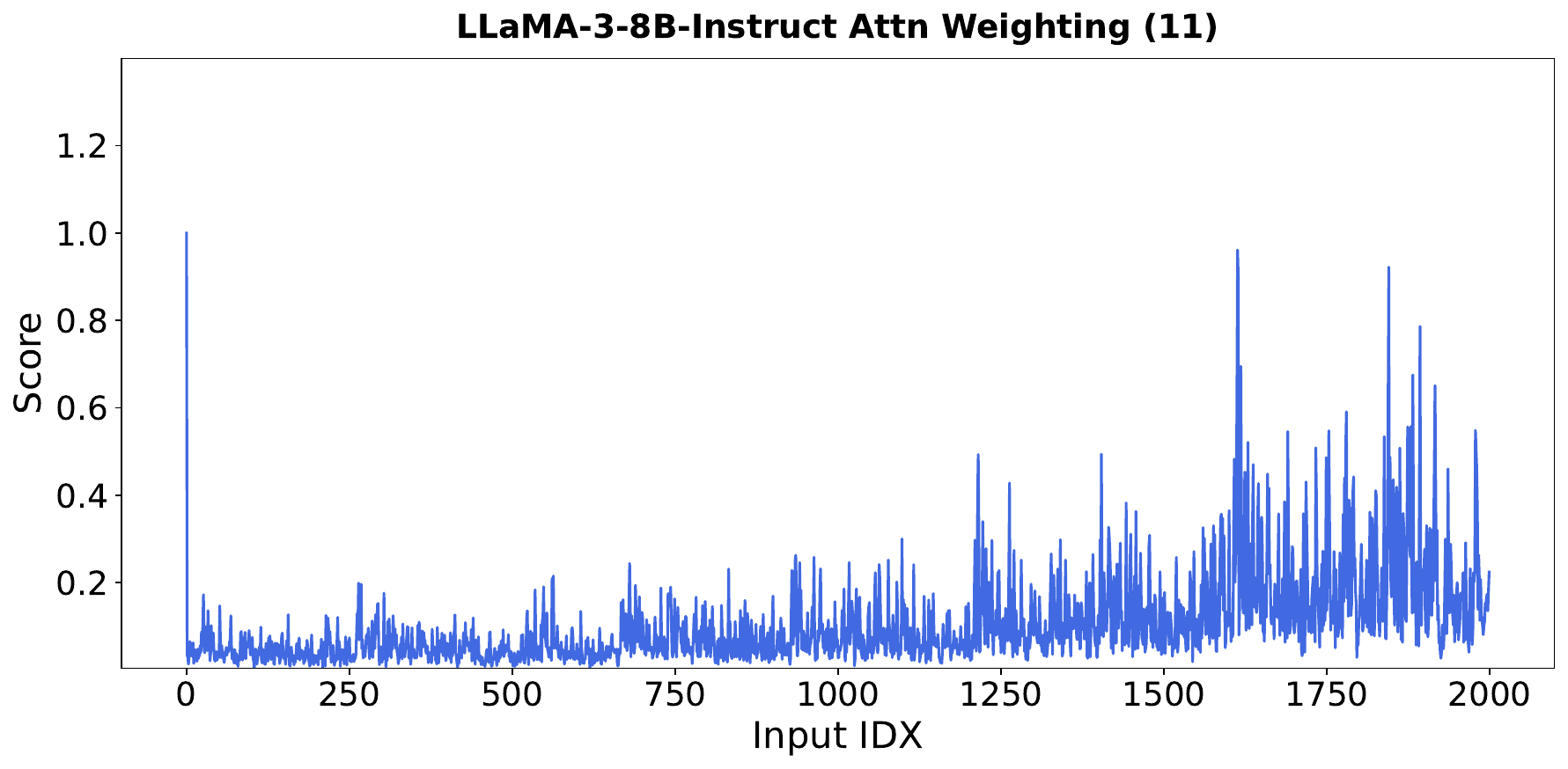}
        \captionsetup{width=0.4\linewidth}
        \caption{Layer 11}
     \end{subfigure}
     \begin{subfigure}[b]{0.33\linewidth}
        \centering
        \includegraphics[width=\linewidth]{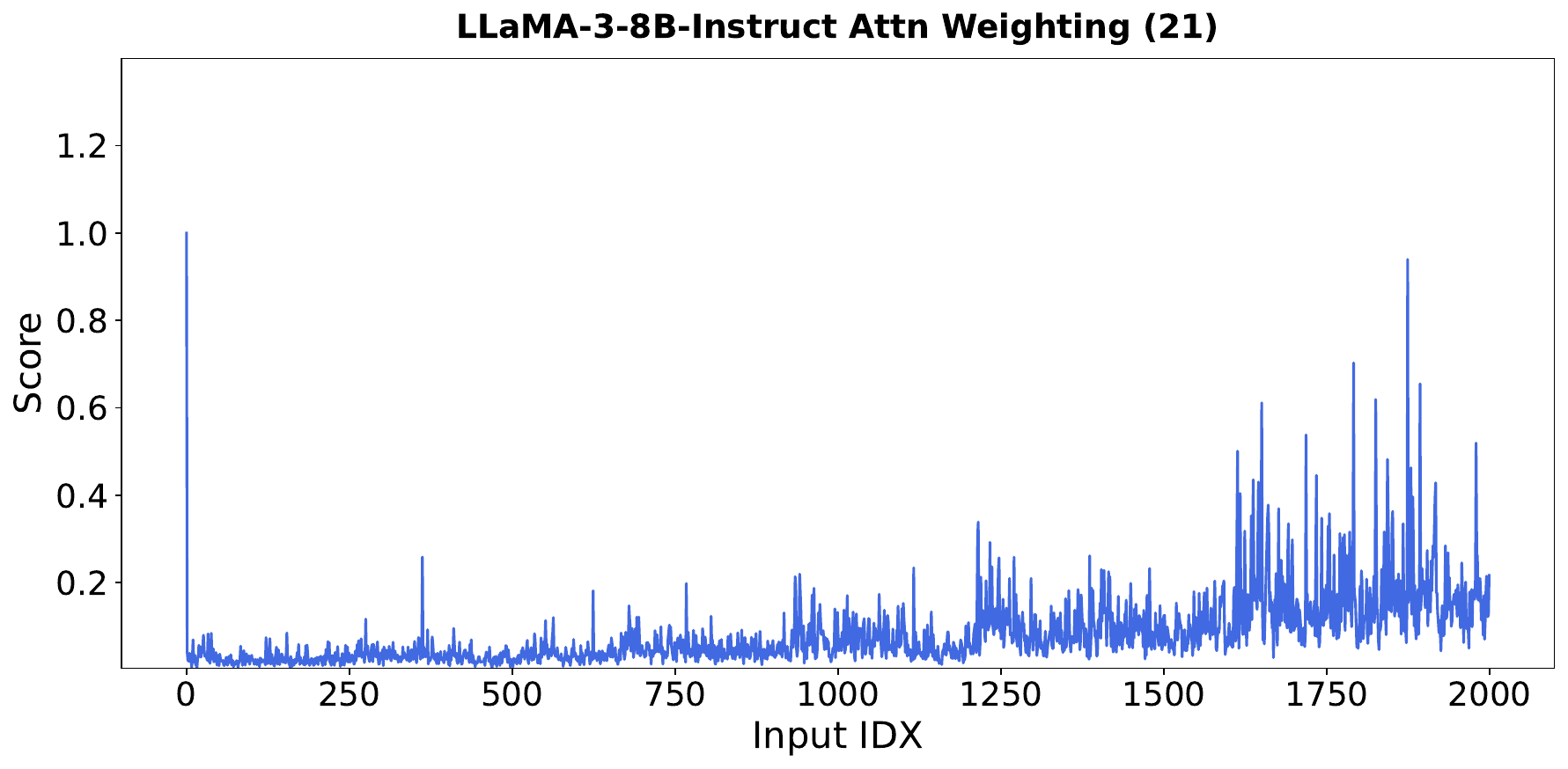}
        \captionsetup{width=0.4\linewidth}
        \caption{Layer 21}
     \end{subfigure}
    
    \vspace{0.3cm} 

     \begin{minipage}{\textwidth}
        \centering
        \textbf{Figures (g) - (i):} third example. 
    \end{minipage}
    \vspace{0.3cm}
    \caption{Visualization of AttnCon scores across three layers for three different examples.}
    \label{fig:attncon_visualization}
\end{figure*}